\documentclass{article}
\usepackage{arxiv}

\usepackage[utf8]{inputenc} 
\usepackage[T1]{fontenc}    
\usepackage{hyperref}       
\usepackage{url}            
\usepackage{booktabs}       
\usepackage{amsfonts}       
\usepackage{nicefrac}       
\usepackage{microtype}      
\usepackage{graphicx}
\usepackage{doi}
\usepackage[numbers]{natbib}

\usepackage{cite}
\usepackage{amsmath,amssymb}
\usepackage{algorithmic}
\usepackage{textcomp}
\usepackage[nolist,printonlyused]{acronym}

\newacro{US}[U.S.]{United States}
\newacro{FBI}[FBI]{Federal Bureau of Investigations}
\newacro{CIA}[CIA]{Central Intelligence Agency}

\begin{acronym}[GDPR] 

\acrodefplural{BNN}[BNNs]{Bayesian Neural Networks}
\acro{DoD}{Department of Defense}
\acro{NPS}{Naval Postgraduate School}
\acro{ML}{machine learning}
\acro{FL}{federated learning}
\acro{HFL}{hierarchical \ac{FL}}
\acro{BDL}{Bayesian deep learning}
\acro{DL}{deep learning}
\acro{CL}{continual learning}
\acro{AL}{active learning}
\acro{CNN}{convolutional neural network}
\acro{GDPR}{General Data Protection Regulation}
\acro{BFL}{Bayesian federated learning}
\acro{IID}{independent and identically distributed}
\acro{VI}{variational inference}
\acro{NWA}{Naive Weighted Averaging}
\acro{WS}{Weighted Sum of Normal Distributions}
\acro{LP}{Linear Pooling}
\acro{IV}{Inverse Variance Weighting}
\acro{DWC}{Distributed Weight Consolidation}
\acro{CLP}{Centered Linear Pool}
\acro{LLP}{Log-Linear Pooling}
\acro{WC}{Weighted Conflation}
\acro{BNN}{Bayesian neural network}
\acro{MC}{Monte Carlo}
\acro{KL}{Kullback-Leibler}
\acro{NLL}{negative log likelihood}
\acro{SGD}{Stochastic Gradient Descent}
\acro{ELBO}{Evidence Lower Bound}
\acro{TL}{transfer learning}
\acro{ResNet}{residual network}
\acro{mAP}{mean average precision}
\acro{AUC}{area under the curve}
\acro{VI-PANN}{variational inference pre-trained audio neural network}
\acro{PANN}{pre-trained audio neural network}
\acro{MCMC}{Markov Chain Monte Carlo}
\acro{CI}{confidence interval}
\acro{OOD}{out-of-distribution}
\acro{SOTA}{state-of-the-art}
\end{acronym}

\usepackage[caption=false]{subfig}
\usepackage{multirow}
\usepackage{hhline}

\title{VI-PANN: Harnessing Transfer Learning and Uncertainty-Aware Variational Inference for Improved Generalization in Audio Pattern Recognition}

\date{February 29, 2024}	

\author{ \href{https://orcid.org/0000-0002-2565-6706}{John~Fischer} \\
	Department of Computer Science\\
	Naval Postgraduate School\\
	Monterey, CA 93954 \\
	\texttt{john.fischer@nps.edu} \\
	\And
	\href{https://orcid.org/0000-0003-3305-8412}{Marko~Orescanin} \\
	Department of Computer Science\\
	Naval Postgraduate School\\
	Monterey, CA 93954 \\
	\texttt{marko.orescanin@nps.edu} \\
	\And
	\href{https://orcid.org/0009-0006-2932-8458}{Eric Eckstrand} \\
	Data Science and Analytics Group\\
	Naval Postgraduate School\\
	Monterey, CA 93954 \\
	\texttt{eric.eckstrand@nps.edu} \\
}

\renewcommand{\shorttitle}{VI-PANN: Harnessing Transfer Learning and Uncertainty-Aware Variational Inference}

\hypersetup{
pdftitle={VI-PANN: Harnessing Transfer Learning and Uncertainty-Aware Variational Inference for Improved Generalization in Audio Pattern Recognition},
pdfauthor={John~Fischer,Marko~Orescanin,Eric~Eckstrand},
pdfkeywords={audio event detection, AudioSet, Bayesian deep learning, transfer learning, uncertainty decomposition, uncertainty quantification},
}

\begin{document}
\maketitle

\begin{abstract}
	Transfer learning (TL) is an increasingly popular approach to training deep learning (DL) models that leverages the knowledge gained by training a foundation model on diverse, large-scale datasets for use on downstream tasks where less domain- or task-specific data is available.  The literature is rich with TL techniques and applications; however, the bulk of the research makes use of deterministic DL models which are often uncalibrated and lack the ability to communicate a measure of epistemic (model) uncertainty in prediction.  Unlike their deterministic counterparts, Bayesian DL (BDL) models are often well-calibrated, provide access to epistemic uncertainty for a prediction, and are capable of achieving competitive predictive performance. In this study, we propose variational inference pre-trained audio neural networks (VI-PANNs).  VI-PANNs are a variational inference variant of the popular ResNet-54 architecture which are pre-trained on AudioSet, a large-scale audio event detection dataset.  We evaluate the quality of the resulting uncertainty when transferring knowledge from VI-PANNs to other downstream acoustic classification tasks using the ESC-50, UrbanSound8K, and DCASE2013 datasets. We demonstrate, for the first time, that it is possible to transfer calibrated uncertainty information along with knowledge from upstream tasks to enhance a model's capability to perform downstream tasks.
\end{abstract}


\section{Introduction}\label{sec:intro}

\Ac{TL} leverages knowledge gained from large foundation models to enhance performance on downstream tasks. In the audio domain, the feasibility of \ac{TL} has been demonstrated through the successful application of \ac{TL} techniques in numerous applications ranging from music genre classification to heart sound classification \citep{kong2020,tsalera2021TL,choi2017transfer, KoikeTLHeart, lopez2021, dimentweaklylabeledaudio}. While deterministic embeddings prevail, variational embeddings provide a promising Bayesian alternative. By using \ac{VI} to infer posterior distributions over latent features, we obtain variational embeddings which capture uncertainty and enable new analyses of transferred representations. However, the use of variational embeddings in \ac{TL} remains relatively unexplored despite their ability to capture uncertainty. Uncertainty estimation is crucial for assessing model credibility and identifying unreliable predictions \citep{beckler_multi-label_2021,FischerJoE,ortizTGRS, orescanin2021bayesian}. Specifically, the variance of variational embeddings provides epistemic uncertainty estimates that indicate when models lack knowledge \citep{FischerJoE, ortizTGRS}. This benefits building reliable artificial intelligence systems across audio domains.

In this study, we conduct a comprehensive analysis of uncertainty-aware \ac{TL} with variational audio embeddings. Bayesian versions of the popular ResNet architecture \citep{he2016deep} are first trained on the large-scale AudioSet dataset to extract robust acoustic features along with uncertainty estimates. Building off the work of \citep{kong2020}, we call our approach \acp{VI-PANN}. \acp{VI-PANN} are then transferred to three audio classification benchmark datasets - ESC-50 \citep{esc50}, UrbanSound8K \citep{urbanSound}, and DCASE2013 \citep{dcase2013}. Our goal is to rigorously evaluate the quality and utility of the resulting uncertainty from variational embeddings after transfer to sparse data scenarios.

While extensive work exists on uncertainty quantification and decomposition for classification problems \citep{Gal2016Uncertainty, pmlr-v80-depeweg18a, Chai2018Uncertainty}, the focus of this work is on the multi-class case, where each sample contains only a single label.  Unfortunately, this restriction hinders the application of existing techniques to real-world datasets where multiple labels may be present in each sample \citep{pfau_multi-label_2020, beckler_multi-label_2021, audioset}. To address this limitation, we describe a new uncertainty decomposition method for use in multi-label classification and demonstrate its efficacy on AudioSet.

This manuscript makes several contributions: 

\begin{enumerate}
    \item We modify the popular ResNet-54 architecture used in \citep{kong2020} to create two distinct \ac{VI} model variants, namely \ac{MC} dropout \citep{gal2016dropout} and Flipout \citep{wen2018flipout}. This modification enhances the model by providing access to calibrated epistemic (model) uncertainty information, which was lacking in the original model. We pre-train these \ac{VI-PANN} models on the AudioSet dataset, and share the resulting checkpoints for future research.\footnote{https://github.com/marko-orescanin-nps/VI-PANN}
    \item We derive a method for decomposing uncertainty information in the multi-label classification scenario.
    \item We apply the multi-label uncertainty decomposition technique to analyze the uncertainty of \acp{VI-PANN} on the AudioSet validation set, providing insights into the model's performance in real-world scenarios.
    \item We systematically evaluate Bayesian \ac{TL} techniques with a specific emphasis on the quality of uncertainty. This evaluation is conducted using three publicly available audio datasets (ESC-50, UrbanSound8K, and DCASE2013). 
    \item In the audio pattern recognition domain, we demonstrate, for the first time, the feasibility of transferring calibrated uncertainty information alongside knowledge from upstream tasks. This transfer markedly enhances a model's capability to excel in downstream tasks, showcasing an innovative approach in leveraging uncertainty for improved model performance.
\end{enumerate}

\section{Related Work}\label{sec:relatedwork}

\subsection{Transfer learning in audio}\label{sec:TL}

Recently, with popular deep learning frameworks like PyTorch offering pre-trained initializations for modern model architectures, \ac{TL} has become an integral part of modern model development workflows. Building upon this, research efforts have leveraged the large-scale AudioSet dataset to pre-train deep neural networks for enhanced performance on downstream audio tasks \citep{kong2020, lopez2021}. A common TL approach is to directly extract features from a pre-trained model fixed after the initial training. This method transfers general acoustic knowledge to new tasks without updating the model parameters. However, fine-tuning the pre-trained model by allowing parameter updates during training on the downstream data can further improve results by adapting to the task \citep{kong2020}. In this manuscript, we refer to these techniques as ``fixed-feature'' and ``fine-tuned,'' respectively. Leveraging large pre-trained models via either technique provides significant performance gains across various audio applications \citep{kong2020,lopez2021}.

\subsection{Bayesian deep learning}\label{sec:BDL}
The inability of modern deterministic deep learning models to communicate a measure of epistemic (model) uncertainty in prediction has led to an increased interest in \ac{BDL}, specifically in remote sensing \citep{beckler_multi-label_2021,FischerJoE,ortizTGRS}, medical \citep{kwon2020uncertainty,McClureDWC}, and safety-critical applications \citep{autonomousvehicles}. Although there are a number of different approaches to \ac{BDL}, we focus our experiments on \ac{VI}.  Due to the increased speed and the ability to scale with data and models, \ac{VI} is often favored over techniques like \ac{MCMC} \citep{JMLR:v14:hoffman13a}. In modern probabilistic machine learning libraries like BayesianTorch \citep{bayesiantorch}, the most common \ac{VI} implementations are Flipout \citep{wen2018flipout} and the Local Reparameterization Trick \citep{kingma_variational_2015}.  Due to the fact that both of these approaches represent each model weight using a Gaussian Distribution (i.e., each weight is defined using two model parameters, a mean and a variance), they effectively double the number of model parameters. In 2016, Gal \textit{et al.} \citep{gal2016dropout} showed that it was possible to perform \ac{VI} by training a model with dropout layers preceding every weight layer and activating those dropout layers during inference. This approach, called \ac{MC} dropout, does not double the number of model parameters. For this reason, along with the minimal changes required to common deep learning model architectures and training procedures, \ac{MC} dropout is often favored over other \ac{VI} approaches.  In this work, we focus on the Flipout and \ac{MC} dropout implementations of \ac{VI}.

\subsection{Uncertainty quantification and decomposition in multi-class classification}\label{sec:uncertainty}

One of the primary motivations behind using \ac{BDL} models is to gain access to high-quality uncertainty for predictions.  The existing research literature is rich with techniques for quantifying and decomposing uncertainty. In \citep{Kendall_Gal_NIPS2017_2650d608}, Kendall and Gal provide insight into two types of uncertainty that can be modeled. Aleatoric, or irreducible, uncertainty is the uncertainty inherent in the data. Epistemic, or reducible, uncertainty is the uncertainty about the prediction due to uncertainty about the model. In addition to providing a detailed description of these uncertainties, the authors describe a method for measuring predictive (total) uncertainty, based on output variance, and decomposing the total uncertainty into its aleatoric and epistemic components using the laws of total variance.  Unfortunately, the method used by Kendall and Gal requires the use of extra parameters to model the mean and variance of the model output.  Kwon \textit{et al.} \citep{kwon2020uncertainty} expands upon the work in \citep{Kendall_Gal_NIPS2017_2650d608} by proposing a method for calculating these component uncertainties without the use of additional model parameters.   

Another line of research is based on the use of entropy as a measure of predictive uncertainty \citep{pmlr-v80-depeweg18a,Chai2018Uncertainty}. We detail the approach of Chai \citep{Chai2018Uncertainty}, as we use this method for multi-class classification, and it is the basis of our multi-label classification decomposition method.  In \ac{BDL} multi-class classification problems, we approximate the predictive probabilty $p(y = c \mid x)$ using \ac{MC} integration with $M$ samples \citep{filos2019systematic}. The average probability per class $\bar{p}_{c}$ is calculated using  

\begin{equation}
    \bar{p}_{c} = \frac{1}{M}\sum_{m=1}^{M} \hat{p}_{c_{m}},
\label{eq:mean_prob}
\end{equation}

\noindent where $\hat{p}_{c_{m}} = p(y = c \mid x,\theta^m)$ and $\theta^m$ is sampled from an approximation of $p(\theta|\mathcal{D})$.  Defining $C$ as the set of all possible classes, we can then compute the entropy of a prediction with

\begin{equation}
    \mathbb{H}[y \mid x,\mathcal{D}] = - \sum_{c \in C} \bar{p}_{c} \log \bar{p}_{c}.
\label{eq:pred_entropy}
\end{equation}

\noindent Depeweg \textit{et al.} \citep{pmlr-v80-depeweg18a} and Chai \citep{Chai2018Uncertainty} use the entropy from Eq. (\ref{eq:pred_entropy}) as a measure of total uncertainty and decompose that uncertainty using the following:

\begin{equation}
    \begin{split}
       \mathbb{H}[y \mid x,\mathcal{D}] = \underbrace{\mathbb{I}[y,\theta \mid x, \mathcal{D}]}_{Epistemic} + \underbrace{\mathbb{E}_{\theta \sim p(\theta|\mathcal{D})}[\mathbb{H}[y \mid x, \theta]]}_{Aleatoric},
    \end{split}     
\end{equation}\label{eq:depuncertainty}

\noindent where $\mathbb{E}$ is expected value and $\mathbb{I}$ is mutual information.  Similar to the calculation of predictive entropy in Eq. (\ref{eq:pred_entropy}), we approximate the aleatoric uncertainty component using \ac{MC} integration to arrive at the following estimator:

\begin{equation}\label{eq:aleatoric} 
       \mathbb{E}_{\theta \sim p(\theta|\mathcal{D})}[\mathbb{H}[y~|~x,\theta]] \approx -\frac{1}{M}\sum_{m=1}^{M}\sum_{c \in C} \hat{p}_{c_{m}} \log \hat{p}_{c_{m}}.
\end{equation}

\noindent Finally, the epistemic uncertainty component is calculated by finding the difference between Eq. (\ref{eq:pred_entropy}) and (\ref{eq:aleatoric}).  

\section{Methodology}\label{sec:methods}

\subsection{Architecture}\label{sec:architecture}

As a starting point, we adopt the ResNet-54 architecture described in \citep{kong2020} and make use of the source code provided by the authors. In order to evaluate \ac{VI-PANN}, we implement \ac{MC} dropout \citep{gal2016dropout} and Flipout \citep{wen2018flipout} variants of the \ac{PANN} architecture in \citep{kong2020}.  

For the \ac{MC} dropout variant, the architecture of \citep{kong2020} is left unmodified during training.  However, during inference, we explicitly keep dropout layers active.

In order to implement the Flipout model, we utilize the Bayesian-Torch \citep{bayesiantorch} software package.  Using Bayesian-Torch, we convert deterministic layers to Bayesian layers.  More specifically, linear layers are converted to LinearFlipout layers and Conv2d layers are converted to Conv2dFlipout. These weight layers are initialized using the MOPED methodology described in \citep{moped}. In our case, initialization is done by calling the Bayesian-Torch dnn\_to\_bnn() function with our pre-trained deterministic model and the default \textit{moped\_delta} parameter of 0.5. We then modify the cross entropy loss function from \citep{kong2020} to a loss function based on the following form of the negative Evidence Lower Bound (ELBO): 

\begin{equation}
    \mathcal{L}_q = \text{KL}[q_{\phi}(\theta)\mid\mid p(\theta)] - \mathbb{E}_{q} \left[\log p\left(\mathcal{D}|\theta\right) \right],
\label{eq:ELBO}
\end{equation}

\noindent where KL corresponds to the \ac{KL} divergence, and $\mathbb{E}_q$ represents the expected value under the probability distribution $q_{\phi}(\theta)$.  A detailed discussion and derivation of this objective can be found in \citep{FischerJoE}.

\subsection{Uncertainty quantification and decomposition in multi-label classification}\label{sec:uncertaintymethods}

In this work, we train and evaluate \ac{BDL} models on both multi-class and multi-label classification tasks. In the multi-class case (ESC-50, UrbanSound8K, and DCASE2013), we can directly apply the techniques described in Depeweg \textit{et al.}~\citep{pmlr-v80-depeweg18a} and Chai \citep{Chai2018Uncertainty}. In the multi-label case (AudioSet), however, we must modify the multi-class uncertainty decomposition technique to account for the fact that each class is an independent binary classification problem.  Following the methodology in \citep{Chai2018Uncertainty}, we start by calculating the predictive entropy (i.e., total uncertainty).  To calculate the predictive entropy, we first calculate the entropy for each class   

\begin{equation}
    \mathbb{H}[y \mid x,\mathcal{D}] = - \bar{p}_{c}\log \bar{p}_{c} - (1-\bar{p}_{c})\log (1-\bar{p}_{c}), 
\label{eq:binary_pred_entropy}
\end{equation}

\noindent where $\bar{p}_{c}$ is defined in Eq.~(\ref{eq:mean_prob}).  Next, to capture the total entropy of the prediction, we sum over all classes $\sum_{c \in C} \mathbb{H}[y \mid x,\mathcal{D}]$. Borrowing the definition of $\hat{p}_{c_{m}}$ from Section~\ref{sec:uncertainty}, and modifying Eq.~(2.21) from \citep{Chai2018Uncertainty} for the multi-label case, we are left with the following estimator for aleatoric uncertainty

\begin{equation}\label{eq:aleatoric_binary} 
       \mathbb{E}_{\theta \sim p(\theta|\mathcal{D})}[\mathbb{H}[y \mid x, \theta]] \approx -\sum_{c \in C} \frac{1}{M} \sum_{m=1}^{M} \hat{p}_{c_{m}} \log \hat{p}_{c_{m}}  
            + (1-\hat{p}_{c_{m}}) \log (1-\hat{p}_{c_{m}}).
\end{equation}

\noindent Finally, in order to compute epistemic uncertainty, we calculate the difference between Eq. (\ref{eq:binary_pred_entropy}) and (\ref{eq:aleatoric_binary}),

\begin{equation}
    \begin{split}
       &\mathbb{I}[y,\theta \mid x, \mathcal{D}] \approx \mathbb{H}[y \mid x,\mathcal{D}] - \mathbb{E}_{\theta \sim p(\theta|\mathcal{D})}[\mathbb{H}[y \mid x, \theta]].
    \end{split}     
\end{equation}\label{eq:epistemic}

\subsection{Model evaluation}
In order to align with \citep{kong2020} and \citep{audioset}, we present our pre-training results using \ac{mAP}, \ac{AUC}, and d-prime. Similar to \citep{kong2020}, we calculate each metric using macro-averaging (i.e., we calculate each class individually and average across classes).

For assessing model calibration, we draw on the insights from Filos \textit{et al.} \citep{filos2019systematic}, who demonstrated that a well-calibrated model's performance improves when high-uncertainty predictions are discarded. Furthermore, Ortiz \textit{et al.} \citep{ortizTGRS, ortiz2023uncertainty} demonstrated on large scale multispectral satellite datasets (multi-year data) for both classification and regression applications that proper calibration and uncertainty quantification are critical for operational use of neural network models in geoscience applications. Consequently, we employ \ac{mAP} and accuracy versus data retained curves to evaluate model calibration based on predictive entropy, aleatoric uncertainty, and epistemic uncertainty. Plot shading represents a 95\% \ac{CI} calculated over 20 replications.  

To illustrate the practicality of calibrated model uncertainty, we assess each of our \acp{VI-PANN} on the ShipsEar dataset \citep{santos-dominguez_shipsear_2016}. ShipsEar is a multi-class classification dataset comprising 90 underwater sound recordings of 11 different types of ships. This dataset was chosen because each sample is \ac{OOD}, and the recordings, captured underwater with hydrophones, differ from the microphones used in the \ac{TL} datasets in this study. Consequently, we can analyze the change in model uncertainty (total, aleatoric, and epistemic) when each model is evaluated on data types and distributions it hasn't been trained on. 

Due to the fact that our \ac{TL} datasets require cross-fold validation, we present all results averaged across folds.

\subsection{Bayesian deep learning model pre-training}\label{sec:pretrain}

In order to pre-train our models on AudioSet, we adopt the approach and hyperparameters from \citep{kong2020}. Specifically, to standardize and control for the acoustic pre-processing hyperparameters, enabling a direct and meaningful comparison of model performance between our \acp{VI-PANN} and the deterministic models detailed in \citep{kong2020}, AudioSet acoustic segments are resampled to 32kHz and converted to log-mel spectrograms using a Hamming window of 1024, a hop size of 320, and 64 mel filter banks. Additionally, following the approach in \citep{kong2020}, we remove frequencies above 14kHz and below 50Hz from the samples. For additional details on acoustic pre-processing hyperparameter selection, we refer the interested reader to \citep{kong2020}. We use a batch size of 32, and an Adam optimizer with a learning rate of 0.001.  For the \ac{MC} dropout variants, we use a dropout rate of 0.2 for convolutional layers, and 0.5 for linear layers.

We apply this training setup to our YouTube-curated repository of approximately 1.7M 10-second, unbalanced audio clips.  Similar to \citep{kong2020}, we employ mixup \citep{zhang2018mixup} augmentation with $\alpha=1.0$; however, we make no effort to balance the training dataset. As the goal of our investigation is not to match state-of-the-art performance on the AudioSet tagging task but rather construct large-parameter probabilistic versions of AudioSet pre-trained networks to investigate the benefits they confer to uncertainty analysis in the acoustic domain, we train our deterministic \ac{PANN} and \ac{MC} dropout \ac{VI-PANN} for approximately 3M steps. In order to train our Flipout \ac{VI-PANN}, we initialize the network priors and posteriors using MOPED \citep{moped} with the learned weights from our deterministic \ac{PANN}.  We then train the Flipout \ac{VI-PANN} for an additional 2M steps.  The deterministic \ac{PANN} and both \acp{VI-PANN} are evaluated using the AudioSet balanced evaluation split.

\subsection{Bayesian transfer learning}\label{sec:BTL}

In our \ac{TL} experiments, we explore three distinct \ac{TL} strategies.

\begin{enumerate}
    \item \textbf{Flip Strategy}:
    \begin{itemize}
        \item Initialize a Flipout model with parameters from our Flipout \ac{VI-PANN}.
        \item Replace the classification head with a new Flipout head, using the Bayesian-Torch LinearFlipout layer defaults.
        \item Freeze the backbone, train the classification head for 200 epochs with a learning rate of 0.001 (referred to as "fixed-feature"), then unfreeze the backbone, reduce the learning rate by a factor of 10, and train for an additional 200 epochs (referred to as "fine-tuned").
    \end{itemize}
    \item \textbf{Det-Flip Strategy}:
    \begin{itemize}
        \item Initialize a deterministic model with our deterministic \ac{PANN} using MOPED (\textit{moped\_delta = 0.5}), as described in \ref{sec:architecture}.
        \item Replace the deterministic head with a Flipout head, following the Flip strategy workflow.
    \end{itemize}
    \item \textbf{Drop Strategy}:
    \begin{itemize}
        \item Initialize an \ac{MC} dropout model with our \ac{MC} dropout \ac{VI-PANN}.
        \item Replace the head with an \ac{MC} dropout head (dropout rate: 0.5), and follow the Flip strategy workflow. 
    \end{itemize}
\end{enumerate}

For comparison, we include results from a deterministic baseline (\textbf{Det}):

\begin{itemize}
    \item Initialize a deterministic model with our deterministic \ac{PANN}.
    \item Replace the head with a deterministic head and follow the Flip strategy training workflow.
\end{itemize}

These diverse strategies allow us to assess the impact of different transfer learning approaches on model performance.

For comparison, Table \ref{tab:complexity} contains a summary of \ac{TL} model variant, number of parameters, and the number of multiply-accumulate operations (MACs). The presentation is segmented by dataset, as the input feature shape and the number of classes have distinct impacts on the MACs and number of model parameters, respectively.

\begin{table}[ht!]
    \centering
    \caption{Model parameter counts and multiply-accumulate operations (MACs) of the three VI model variants used in the transfer learning experiments}\label{tab:complexity}
    \begin{tabular}{|c|c|c|c|}
    \hline
         Dataset & Model & Params & MACs (per-sample)  \\ \hline
         \multirow{3}{*}{ESC-50} & Drop & 103,340,786 & 13,536,497,600 \\ \cline{2-4}
         & Det-Flip & 107,639,588 & 13,545,306,248\\ \cline{2-4}
         & Flip & 206,620,004 & 26,653,313,032\\ \hline
         \multirow{3}{*}{UrbanSound8k} & Drop & 103,258,826 & 13,536,333,760 \\ \cline{2-4}
         & Det-Flip & 107,475,668 & 13,544,814,568 \\ \cline{2-4}
         & Flip & 206,456,084 & 26,652,821,352 \\ \hline
         \multirow{3}{*}{DCASE2013} & Drop & 103,258,826 & 81,862,378,176 \\ \cline{2-4}
         & Det-Flip & 107,475,668 & 81,870,858,984 \\ \cline{2-4}
         & Flip & 206,456,084 & 160,695,176,040\\ \hline
    \end{tabular}    
\end{table}


\section{Results and Discussion}\label{sec:results}

\subsection{Datasets}\label{sec:data}

To comprehensively evaluate the uncertainty-aware transfer learning approach, we select a single foundation dataset, AudioSet, and three diverse audio classification datasets - ESC-50, UrbanSound8K, and DCASE2013. These datasets offer various sound recognition tasks to assess the generalization of variational embeddings. A summary of the TL dataset details is presented in Table \ref{tab:TLdatasets}.

\begin{table}[ht]
    \centering
    \caption{Characteristics of the datasets used to evaluate \ac{VI-PANN} embeddings in transfer learning}
    \resizebox{.5\columnwidth}{!}{
        \label{tab:TLdatasets}
        
        \begin{tabular}{|c|c|c|c|c|}
            \hline
            Dataset      & Num Clips & Num Classes & Duration & Folds \\ \hline
            ESC-50        & 2,000     &  50   & 5 sec & 5     \\ \hline
            UrbanSound8K  & 8,732     &  10   & $\leq{4}$ sec & 10    \\ \hline
            DCASE2013   & 100       &  10   & 30 sec & 5     \\ \hline
        \end{tabular}}
\end{table}

\textbf{AudioSet} \citep{audioset}: A large-scale audio event recognition dataset consisting of 2.1M 10-second audio samples. Each of these samples were extracted from videos on YouTube and hand-annotated.  Of the approximately 2.1M videos listed in the original AudioSet paper \citep{audioset}, we were only able to obtain approximately 1.7M (many videos from the published dataset are no longer available via the links from \citep{audioset}).  AudioSet has an ontology of 527 classes and is a heavily imbalanced, multi-label dataset (i.e., one or more labels can be present in a given sample). It is essential to note that the label quality varies significantly across classes, with some having noisy labels and others exhibiting high-quality annotations. Within the AudioSet data, there are three splits: a balanced evaluation split, a balanced training split, and an unbalanced training split.

\textbf{ESC-50} \citep{esc50}: A multi-class classification dataset which consists of 2000 five-second recordings organized into 50 classes. Split into 5-folds for cross-validation, it covers a variety of environmental sound events like gunshots, dogs barking, and applause. The ESC-50 dataset is suitable for evaluating fine-grained event recognition capabilities.

\textbf{UrbanSound8K} \citep{urbanSound}: A multi-class classification dataset containing 8732 urban sound excerpts up to 4 seconds, categorized into 10 classes. Split into 10-folds for cross-validation, this dataset is commonly used to evaluate a model's ability to identify ambient urban noises such as air conditioner, car horn, and children playing.

\textbf{DCASE2013} \citep{dcase2013}: A multi-class dataset with 10 classes representing various acoustic scenes and events. It consists of 100 audio samples, each 30 seconds in duration, and is split into 5-folds for cross-validation. DCASE2013 is commonly used to evaluate acoustic scene/event classification in medium duration recordings.

Together, these datasets enable a rigorous evaluation of our approach on diverse audio classification tasks with labelled data far more scarce than the large foundation dataset. Furthermore, the variety of sounds and context shifts across datasets allows us to evaluate the ability of \acp{VI-PANN} to transfer and generalize their learned variational embeddings. Analyzing uncertainties on these datasets will reveal how embedding distributions capture model credibility across different acoustic environments and events.

\subsection{Foundation model (AudioSet)}\label{sec:audiosetresults}

\begin{figure*}[ht!]
    \centering
        \subfloat{
            \centering
            \includegraphics[width=.32\textwidth]{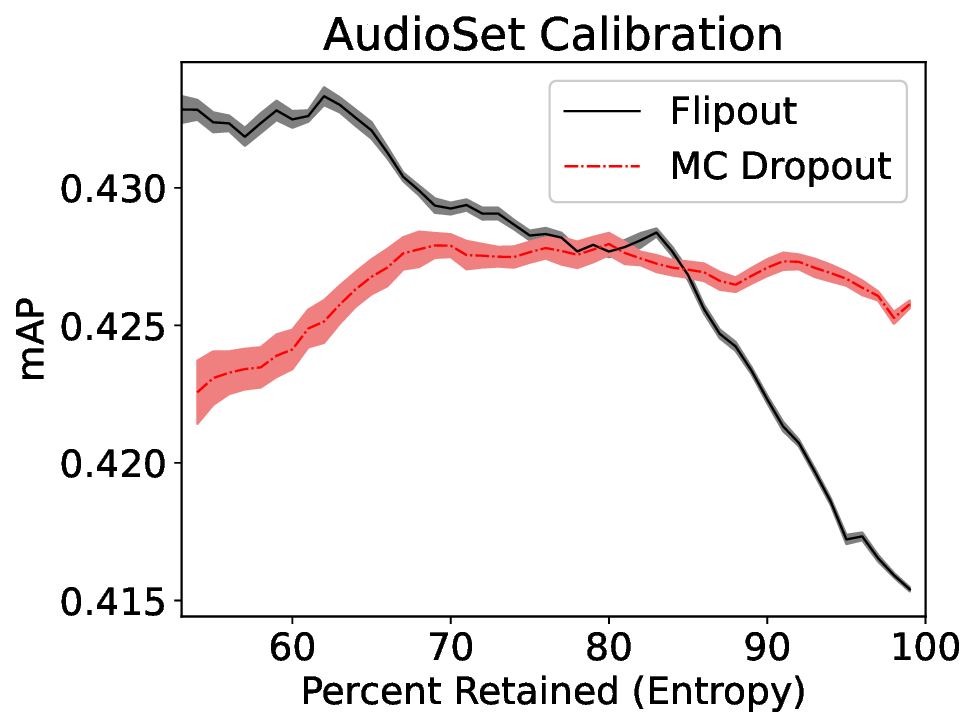}
        }
        \subfloat{
            \centering
            \includegraphics[width=.32\textwidth]{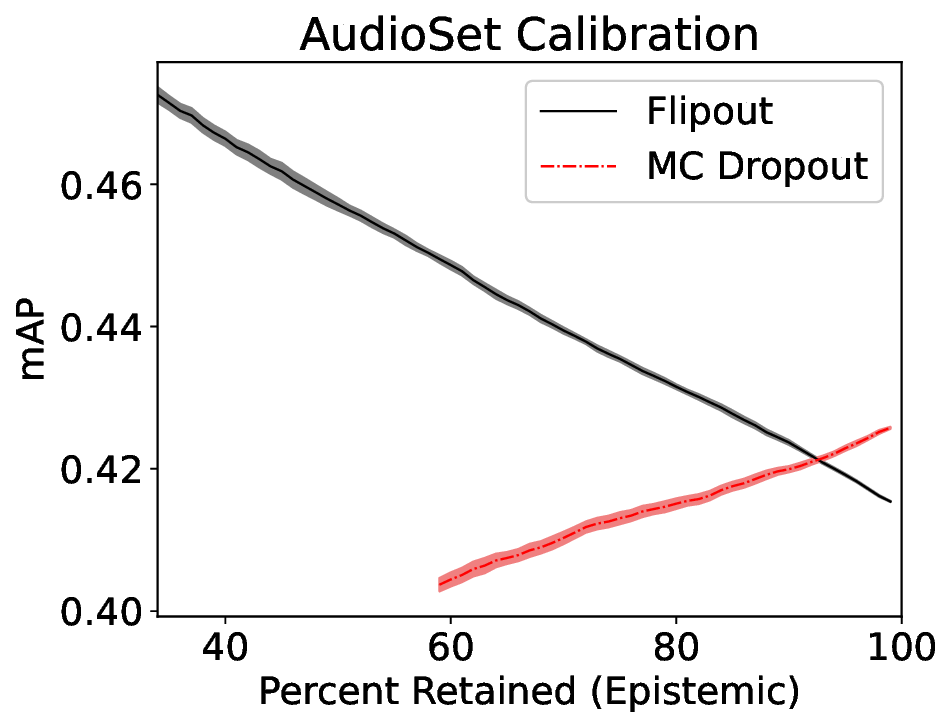}
        }
        \subfloat{
            \centering
            \includegraphics[width=.32\textwidth]{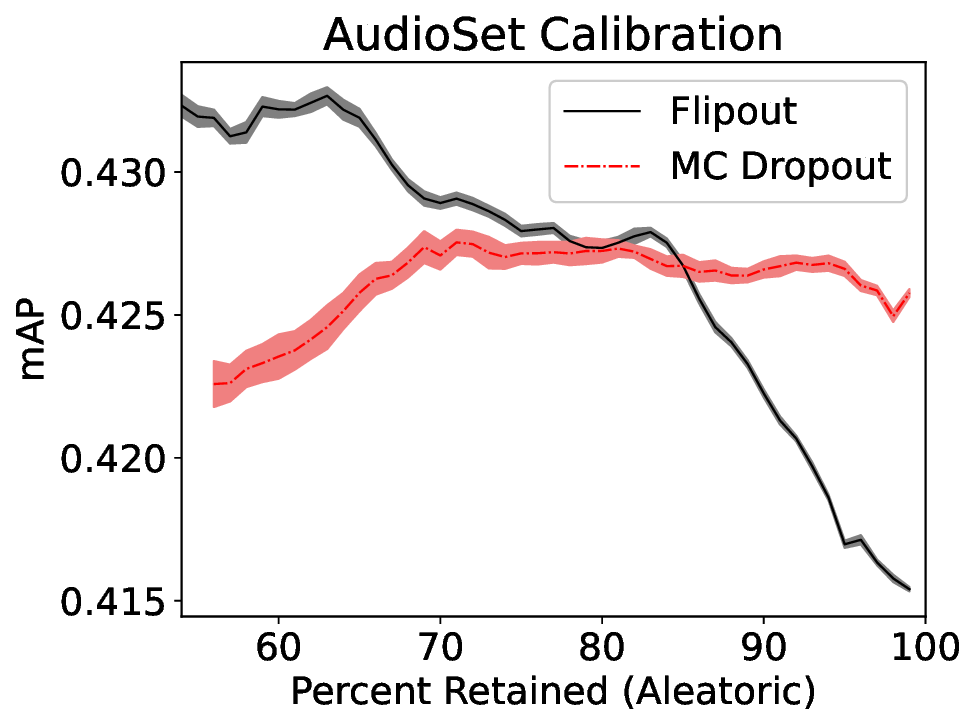}
        } \hfill
\caption{Uncertainty calibration plots for foundation model training on AudioSet.  Comparison plot of test set accuracy vs. percentage of evaluation data retained based on entropy (left), epistemic uncertainty (center), and aleatoric uncertainty (right). Shading represents a 95\% \ac{CI}.}  
    \label{fig:AudioSet Calibration}
\end{figure*}

The results of our AudioSet pre-training are summarized in Table \ref{tab:AudioSetResults}. Each of our models exhibits comparable performance, as measured by mAP, AUC, and d-prime, to the ResNet-54 \ac{PANN} presented in \citep{kong2020}. Alongside performance metrics, we provide predictive entropy (total uncertainty), epistemic uncertainty, and aleatoric uncertainty calibration plots for our \acp{VI-PANN} in Fig.~\ref{fig:AudioSet Calibration}.

\begin{table}[ht!]
    \small
    \begin{center}
        \caption{Model mean average precision (mAP), area under the receiver operating characteristic curve (AUC), and d-prime after pre-training on the AudioSet dataset}
        \label{tab:AudioSetResults}
        \begin{tabular}{|c|c|c|c|}
        \hline
            Model & mAP & AUC & d-prime \\ \hline
            Det & 0.418 & 0.973 & 2.732  \\ \hline
            Drop & 0.426 & 0.975 & 2.762  \\ \hline
            Flip & 0.415 & 0.972 & 2.708  \\ \hline
            ResNet-54 \ac{PANN} \citep{kong2020} &  0.429   & 0.971    & 2.675 \\ \hline
        \end{tabular}
    \end{center}
\end{table}

Our Flipout \ac{VI-PANN} demonstrates well-calibrated uncertainty across all three measures. In contrast, the \ac{MC} dropout \ac{VI-PANN} exhibits poor calibration. A well-calibrated model typically shows improved performance as high-uncertainty predictions are discarded \citep{filos2019systematic}. The observed poor calibration is likely attributed to the complexity and significant class imbalance within the AudioSet dataset. Additionally, the lack of explicit learning of the dropout parameter during the training process may contribute to this issue \citep{gal2017concrete}.

\begin{figure*}[ht!]
    \centering
        \subfloat{
            \centering
            \includegraphics[width=.32\textwidth]{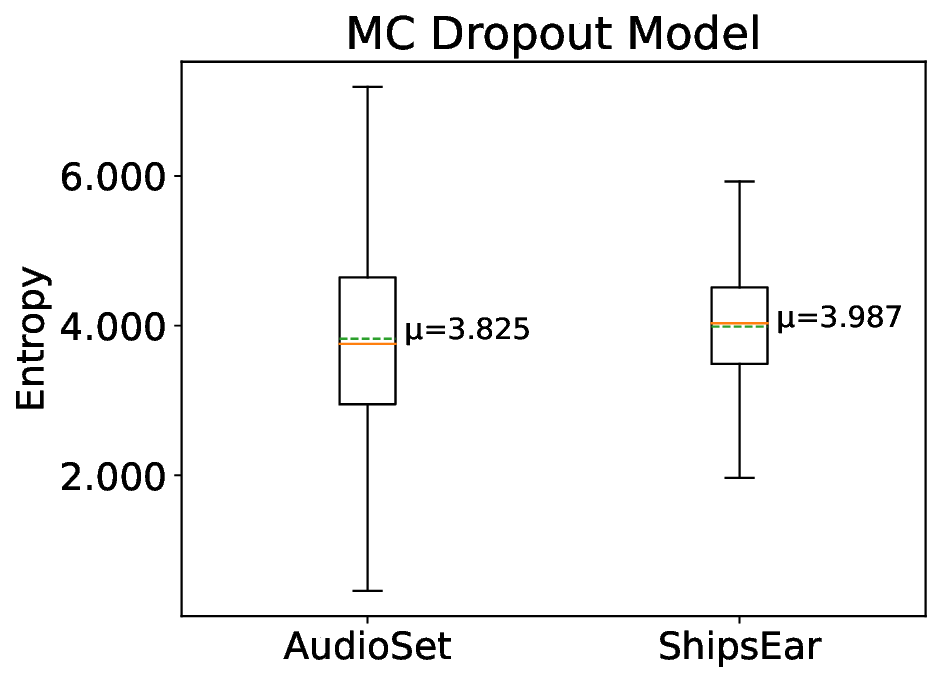}
        }
        \subfloat{
            \centering
            \includegraphics[width=.32\textwidth]{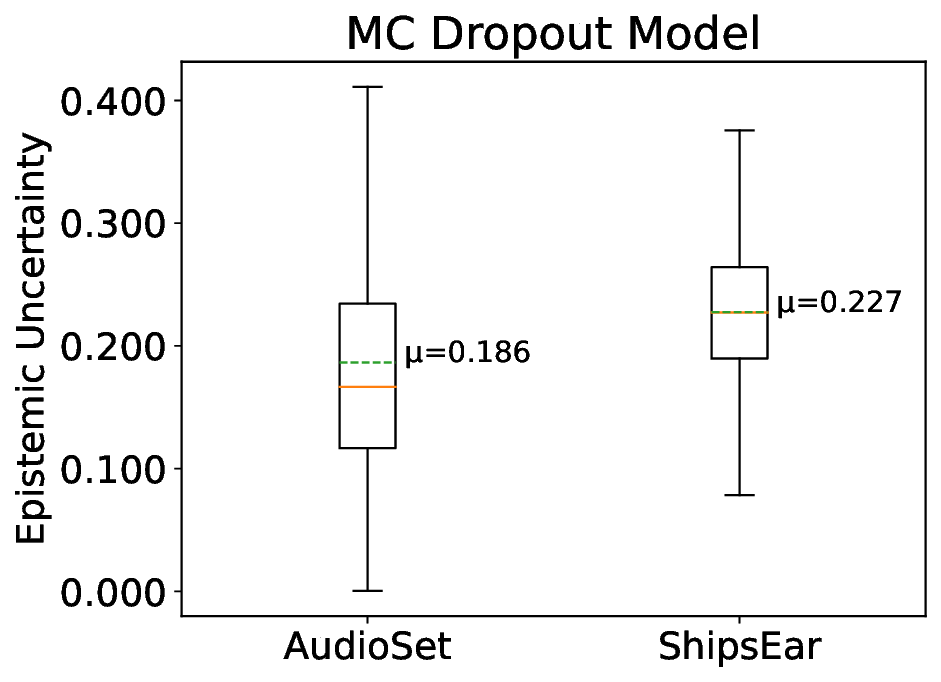}
        }
        \subfloat{
            \centering
            \includegraphics[width=.32\textwidth]{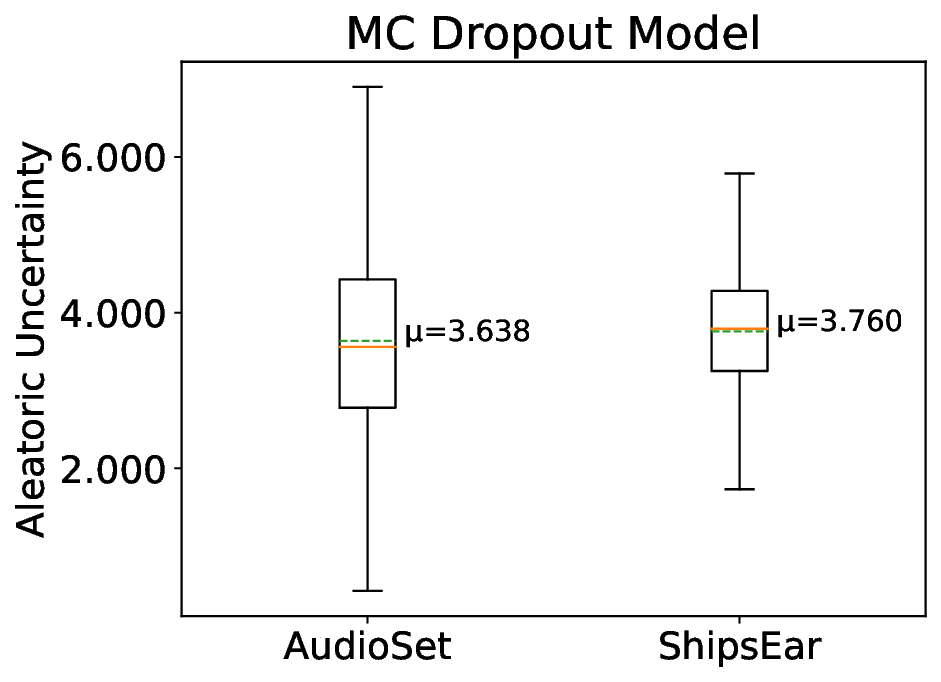}
        } \hfill
        \subfloat{
            \centering
            \includegraphics[width=.32\textwidth]{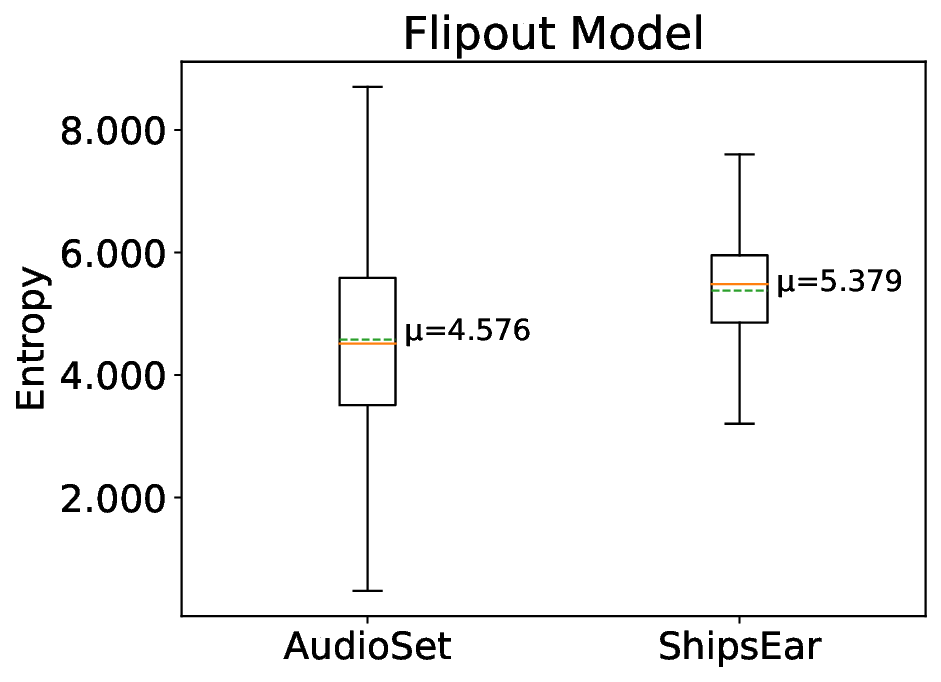}
        }
        \subfloat{
            \centering
            \includegraphics[width=.32\textwidth]{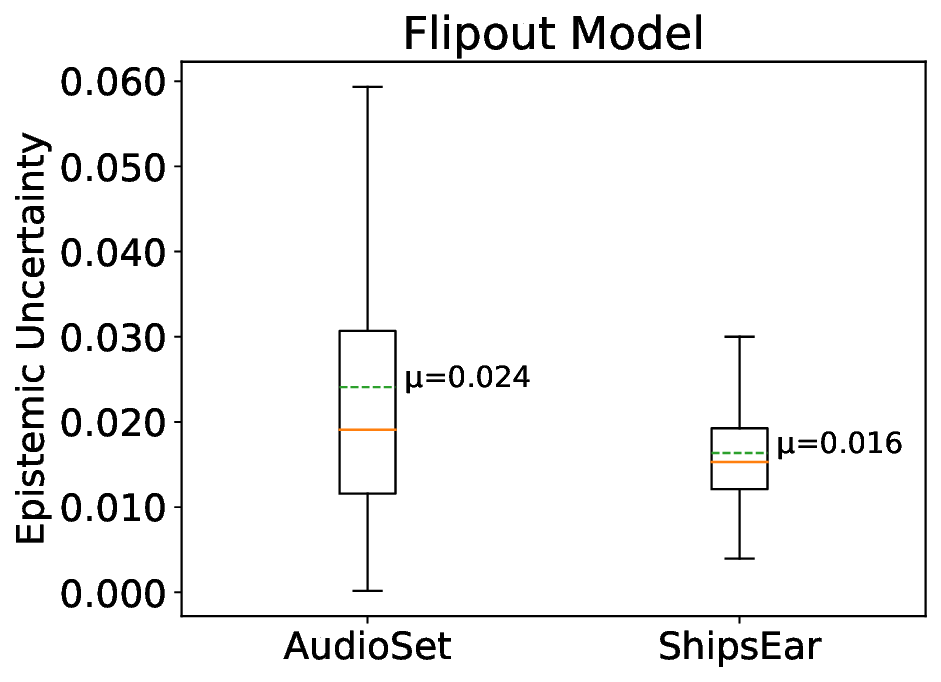}
        }
        \subfloat{
            \centering
            \includegraphics[width=.32\textwidth]{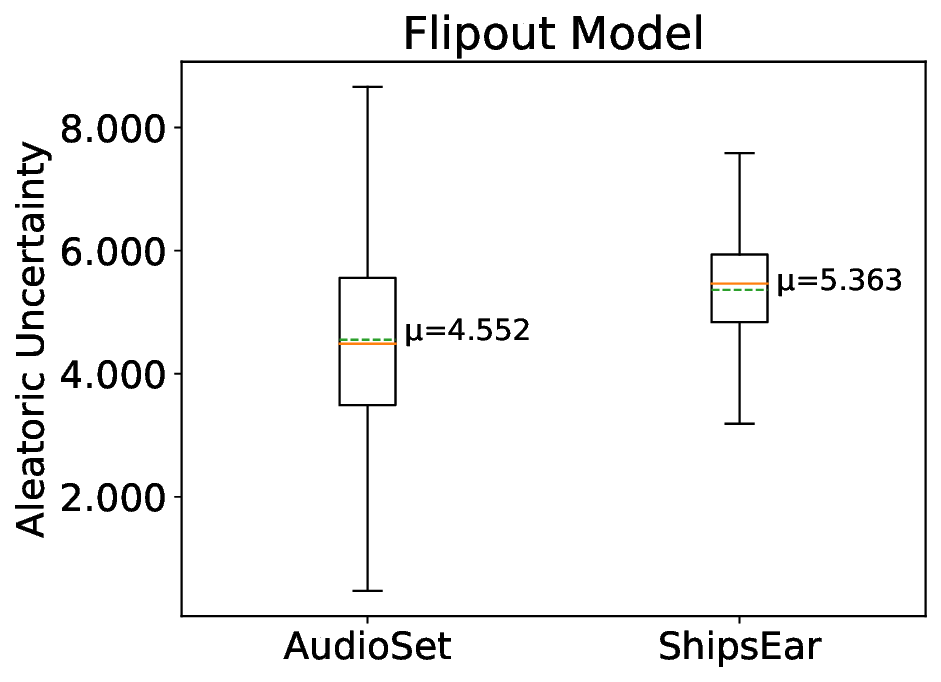}
        } \hfill

    \caption{Uncertainty box plots depicting results of \ac{MC} Dropout model (top row) and Flipout model(bottom row) trained on AudioSet. The plots compare predictive entropy (left), epistemic uncertainty (middle), and aleatoric uncertainty (right) as the models are evaluated on both the AudioSet test set and the ShipsEar dataset. Both the median (orange line) and mean (dashed green line) are presented.} 
    \label{fig:AudioSetBoxPlots}
\end{figure*}

In Fig. \ref{fig:AudioSetBoxPlots}, we present box plots that compare the model uncertainty on both the AudioSet test set and the ShipsEar dataset. Although results from both the Flipout and \ac{MC} Dropout models are included, our primary focus is on the Flipout model due to its demonstrated calibration across all three types of uncertainty. Upon analyzing the model's response to samples from the ShipsEar dataset, we observe a subtle increase in both average entropy and aleatoric uncertainty when compared to the AudioSet test set. Conversely, the average epistemic uncertainty remains consistent across both datasets. Notably, the plots reveal a tighter distribution of uncertainty on the ShipsEar dataset in contrast to the AudioSet dataset. This phenomenon is likely attributed to the diversity of the input data, extreme class imbalance, and unsatisfactory label quality observed in many underrepresented classes within the AudioSet dataset.

\subsection{Transfer Learning}\label{sec:TLresults}

The transfer learning (\ac{TL}) results for ESC-50, UrbanSound8K, and DCASE2013 are detailed in Table \ref{tab:TLResults}. For context, the results of training models from scratch (i.e., without \ac{TL}) are also provided in Table \ref{tab:scratchresults}. Each model variant (Det, Det-Flip, Drop, Flip) exhibits comparable performance across the three datasets. Notably, fine-tuned models demonstrate a substantial increase in accuracy compared to fixed-feature (fixed) models and both \ac{TL} techniques provide significant performance increases over the models trained from scratch on the \ac{TL} datasets. As expected, when trained from scratch on the relatively small \ac{TL} datasets, the high-capacity ResNet-54 models perform relatively poorly and suffer from overfitting. The Flip variant which contains 2x the learnable parameters, when compared to the other variants, performs particularly poorly. Although the Det \ac{PANN} slightly outperforms others in accuracy on ESC-50 and DCASE2013, it lacks the capability to provide access to epistemic uncertainty in predictions.

For reference, we present the results alongside the \ac{SOTA} results for each dataset. It is essential to clarify that the primary aim of this study was not to achieve \ac{SOTA} performance on these datasets. Instead, our goal was to demonstrate performance comparable to existing methods while also offering calibrated epistemic uncertainty information. Nonetheless, our \acp{VI-PANN} demonstrate comparable performance (within 2 to 3\% accuracy) to these \ac{SOTA} approaches. It's worth noting that many of the model architectures employed to achieve \ac{SOTA} performance on these datasets are Transformer-based. In contrast to our approach, these Transformer-based architectures are deterministic and do not provide access to calibrated epistemic uncertainty information. 

\begin{table}[ht]
    \small
    \begin{center}
        \caption{Transfer learning experiment model accuracies on ESC-50, UrbanSound8k, and DCASE2013}
        \resizebox{0.7\columnwidth}{!}{
            \label{tab:TLResults}
                \begin{tabular}{|c|cc|cc|cc|}
                    \hline
                    \multirow{2}{*}{Model} & \multicolumn{2}{c|}{ESC-50} & \multicolumn{2}{c|}{UrbanSound8K} & \multicolumn{2}{c|}{DCASE2013} \\ \cline{2-7} 
                    
                    & \multicolumn{1}{c|}{Fixed} & Fine-Tuned & \multicolumn{1}{c|}{Fixed} & Fine-Tuned & \multicolumn{1}{c|}{Fixed} & Fine-Tuned \\ \hline
                                    
                    Flip & \multicolumn{1}{c|}{0.934}  & 0.950 & \multicolumn{1}{c|}{0.857} & 0.873 &  \multicolumn{1}{c|}{0.860} & 0.870 \\ \hline
                    
                    Det-Flip  & \multicolumn{1}{c|}{0.951}  & 0.962 & \multicolumn{1}{c|}{0.858} & \textbf{0.874} & \multicolumn{1}{c|}{0.790} & 0.850 \\ \hline
                    
                    Drop  & \multicolumn{1}{c|}{0.952}  & 0.958  & \multicolumn{1}{c|}{0.864} & 0.873 & \multicolumn{1}{c|}{0.830} & 0.860  \\ \hhline{=======}
                    
                    Det & \multicolumn{1}{c|}{0.951}  & \textbf{0.964}  & \multicolumn{1}{c|}{0.858} & 0.870 & \multicolumn{1}{c|}{0.870} & \textbf{0.880} \\ \hhline{=======}

                    SOTA & \multicolumn{2}{c|}{0.984 \citep{srivastava2024omnivec} } & \multicolumn{2}{c|}{0.908 \citep{verbitskiy2022}} & \multicolumn{2}{c|}{0.916 \citep{lopez2021}} \\ \hline
                \end{tabular}}
    \end{center}            
\end{table}

\begin{table}[ht!]
    \centering
    \caption{Baseline model accuracies after training on ESC-50, UrbanSound8k, and DCASE2013 without transfer learning}
    \begin{tabular}{|c|c|c|c|}
        \hline
        Model & ESC-50 & UrbanSound8k & DCASE2013  \\ \hline
        Flip & 0.422 & 0.794 & 0.400 \\ \hline
        Det-Flip & 0.834 & 0.815 & 0.610 \\ \hline
        Drop & 0.861 & 0.827 & 0.680 \\ \hhline{====}
        Det & 0.858 & 0.827 & 0.730 \\ \hline
    \end{tabular}
    \label{tab:scratchresults}
\end{table}

\begin{figure*}[ht!]
    \centering
        \subfloat{
            \centering
            \includegraphics[width=.32\textwidth]{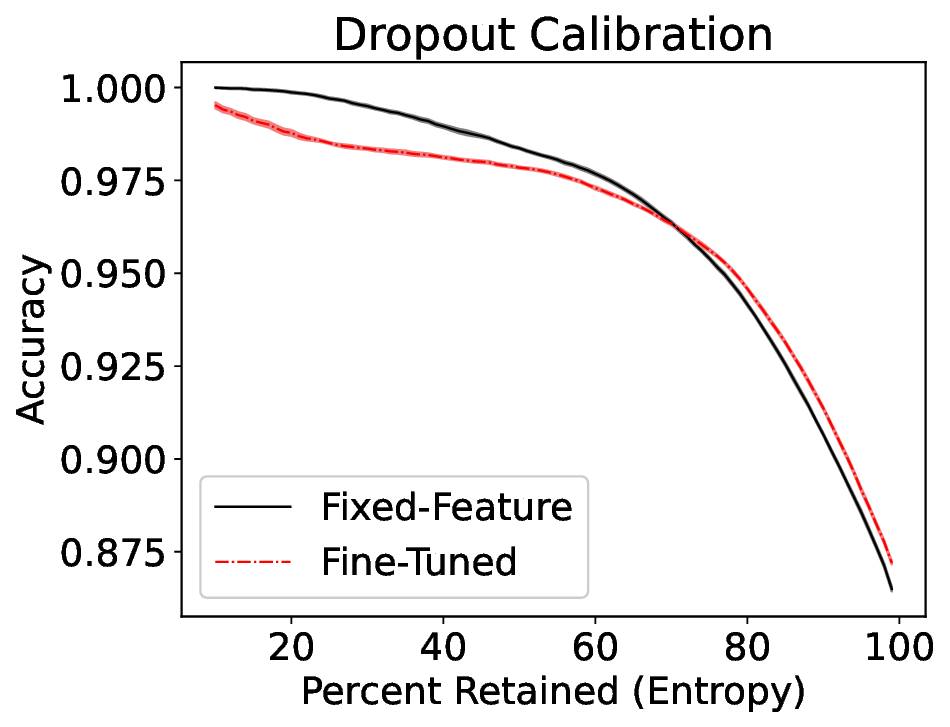}
        }
        \subfloat{
            \centering
            \includegraphics[width=.32\textwidth]{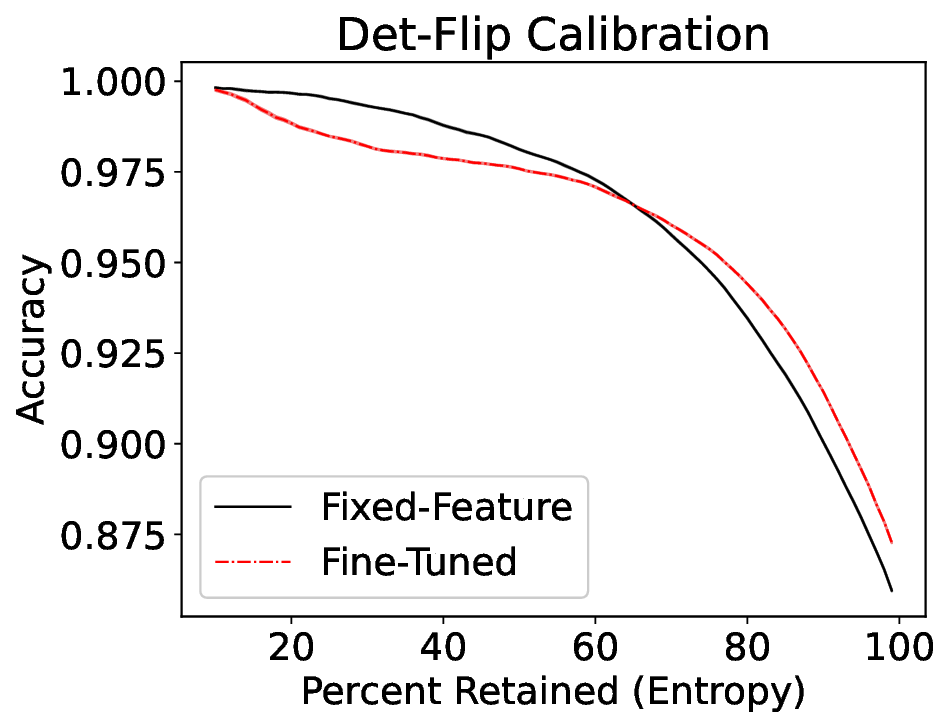}
        }
        \subfloat{
            \centering
            \includegraphics[width=.32\textwidth]{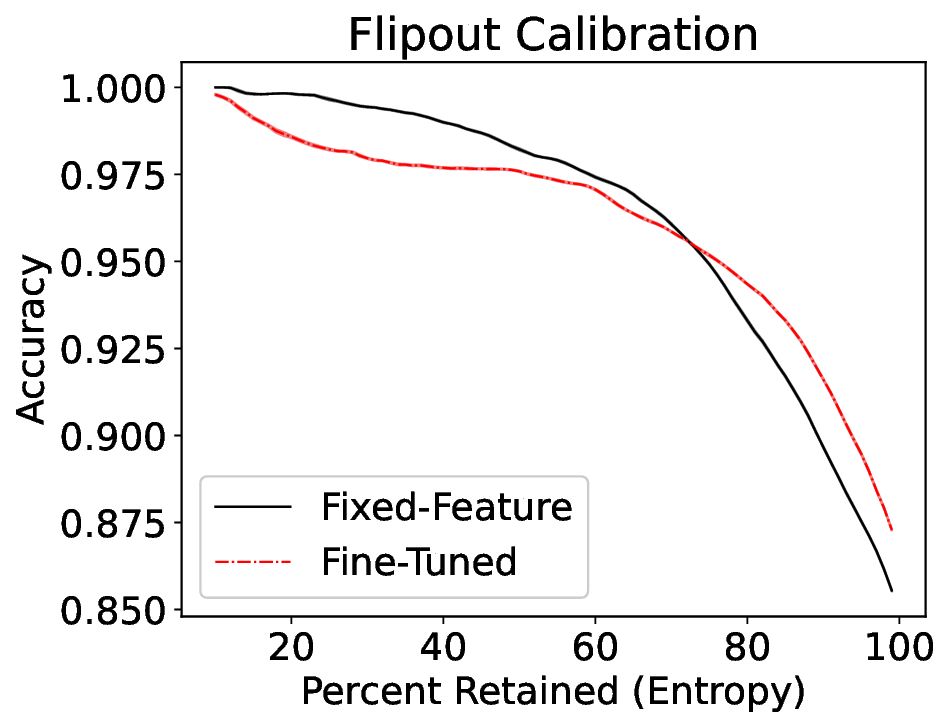}
        } \hfill
        \subfloat{
            \centering
            \includegraphics[width=.32\textwidth]{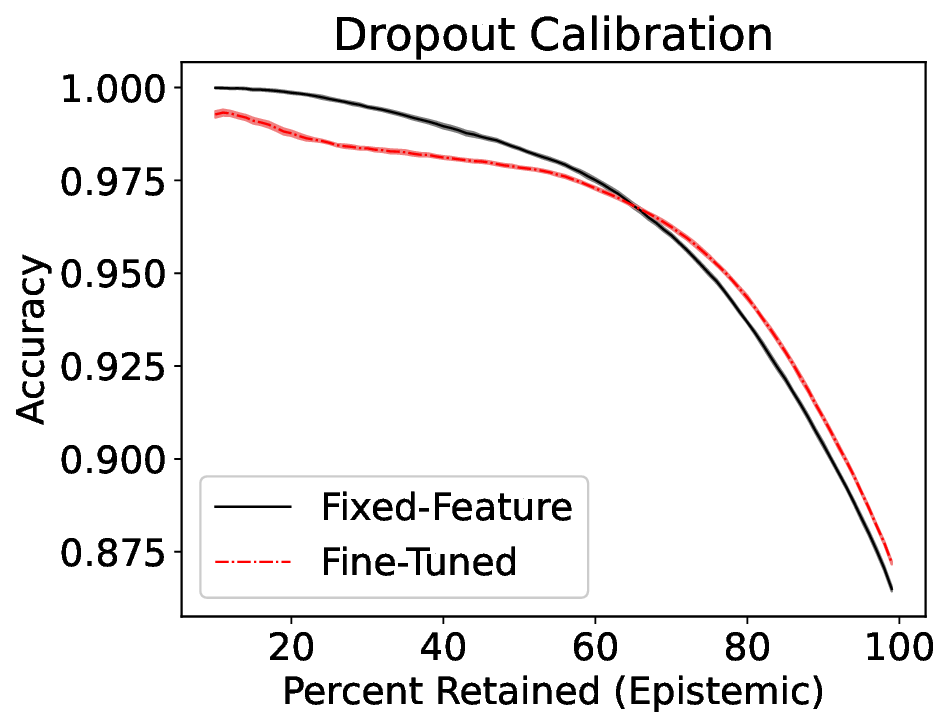}
        }
        \subfloat{
            \centering
            \includegraphics[width=.32\textwidth]{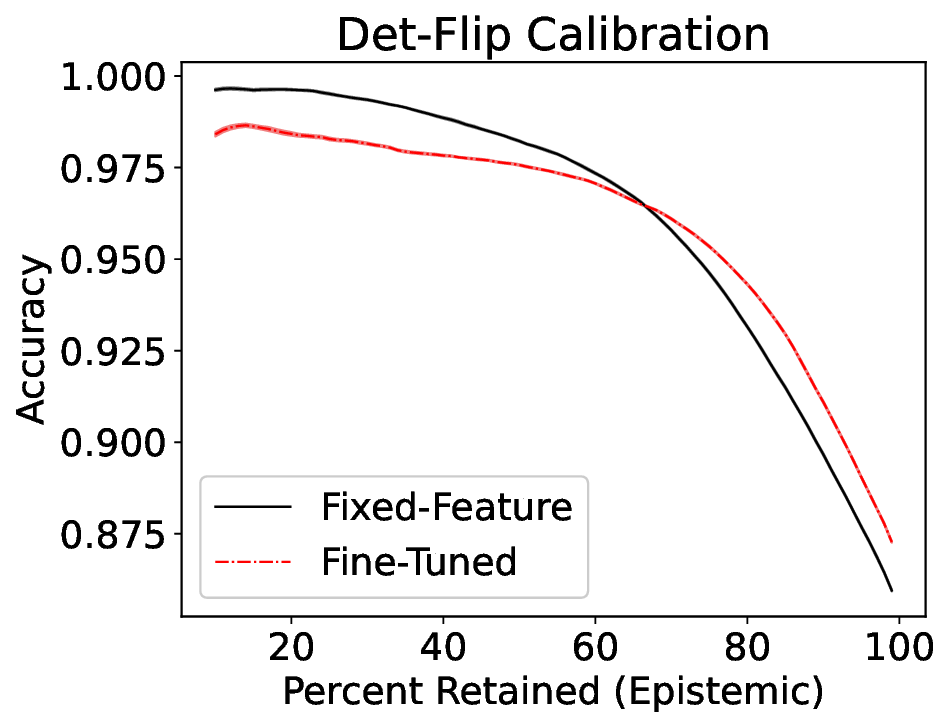}
        }
        \subfloat{
            \centering
            \includegraphics[width=.32\textwidth]{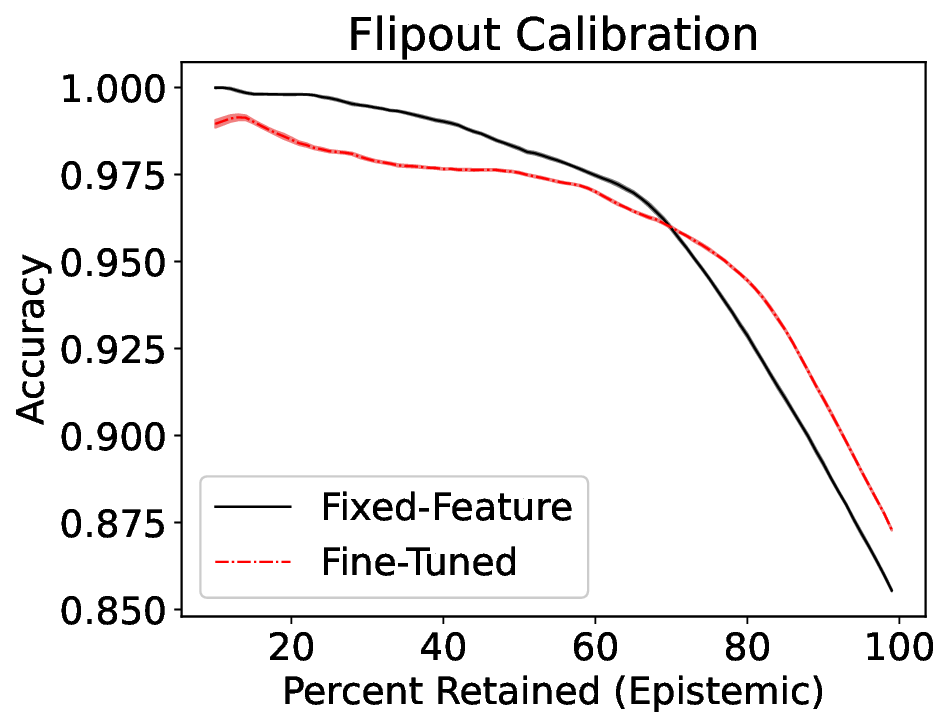}
        } \hfill
        \subfloat{
            \centering
            \includegraphics[width=.32\textwidth]{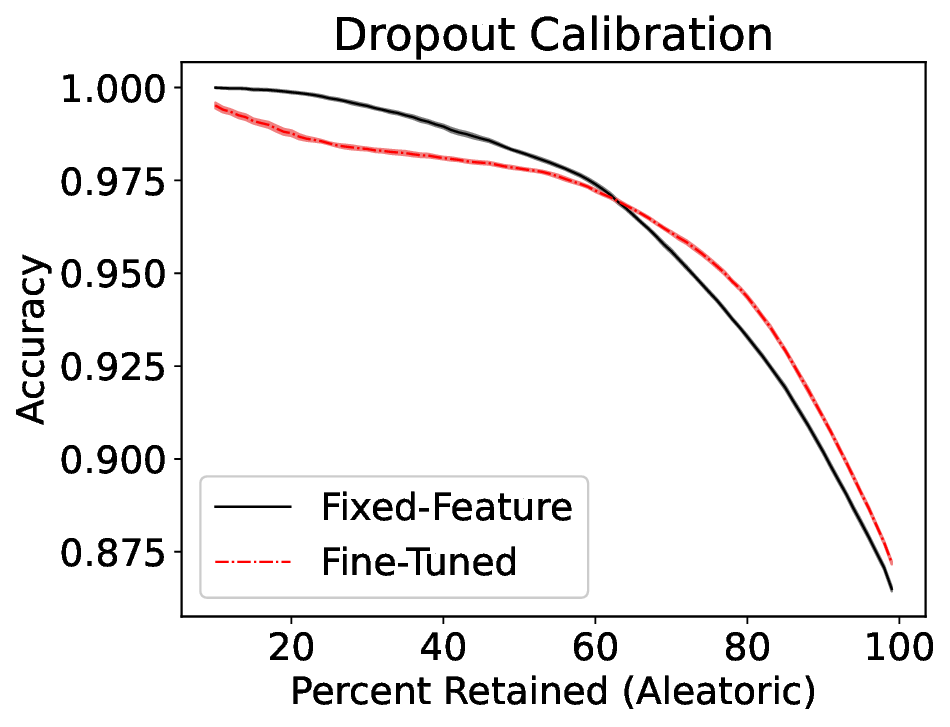}
        }
        \subfloat{
            \centering
            \includegraphics[width=.32\textwidth]{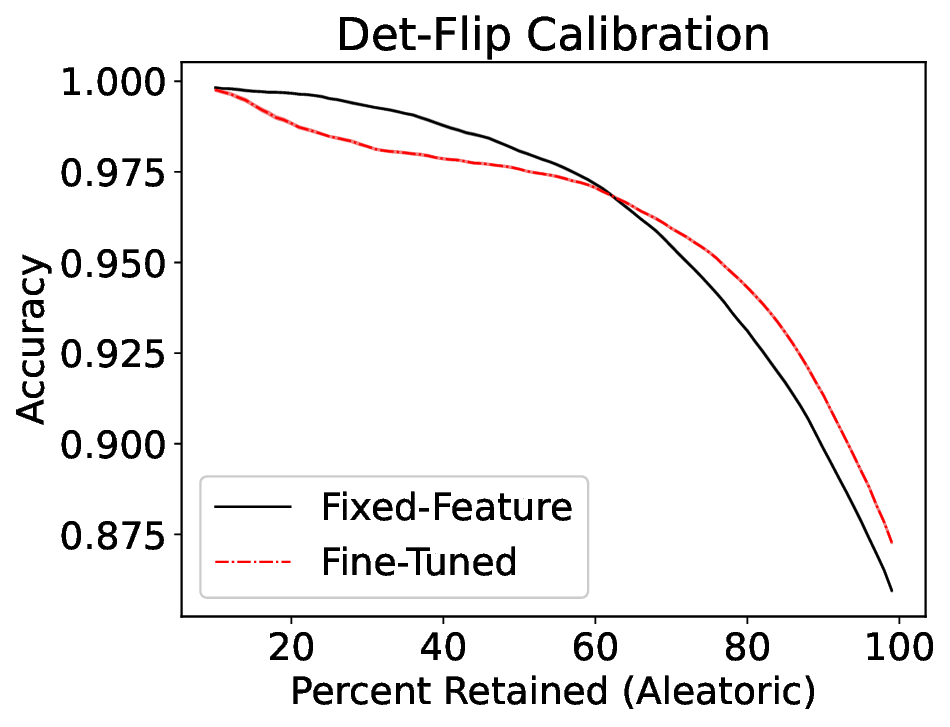}
        }
        \subfloat{
            \centering
            \includegraphics[width=.32\textwidth]{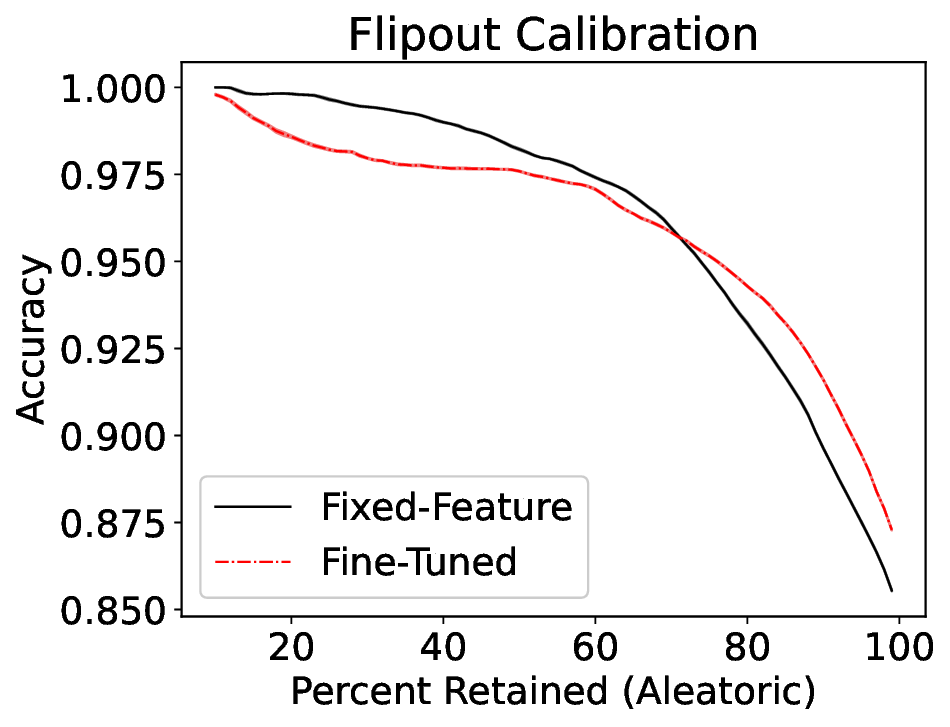}
        }

    \caption{Uncertainty calibration plots comparing fixed-feature and fine-tuning \ac{TL} techniques on UrbanSound8K.  Comparison plots of test set accuracy vs. percentage of evaluation data retained based on Entropy (top), Epistemic Uncertainty (middle) and Aleatoric Uncertainty (bottom).  Drop \ac{VI-PANN} is on the left, Det-Flip \ac{VI-PANN} in the center, and Flip \ac{VI-PANN} on the right. Shading represents a 95\% CI.}  
    \label{fig:UrbanSound8kCalibration}
\end{figure*}

\begin{figure*}[ht!]
    \centering
        \subfloat{
            \centering
            \includegraphics[width=.32\textwidth]{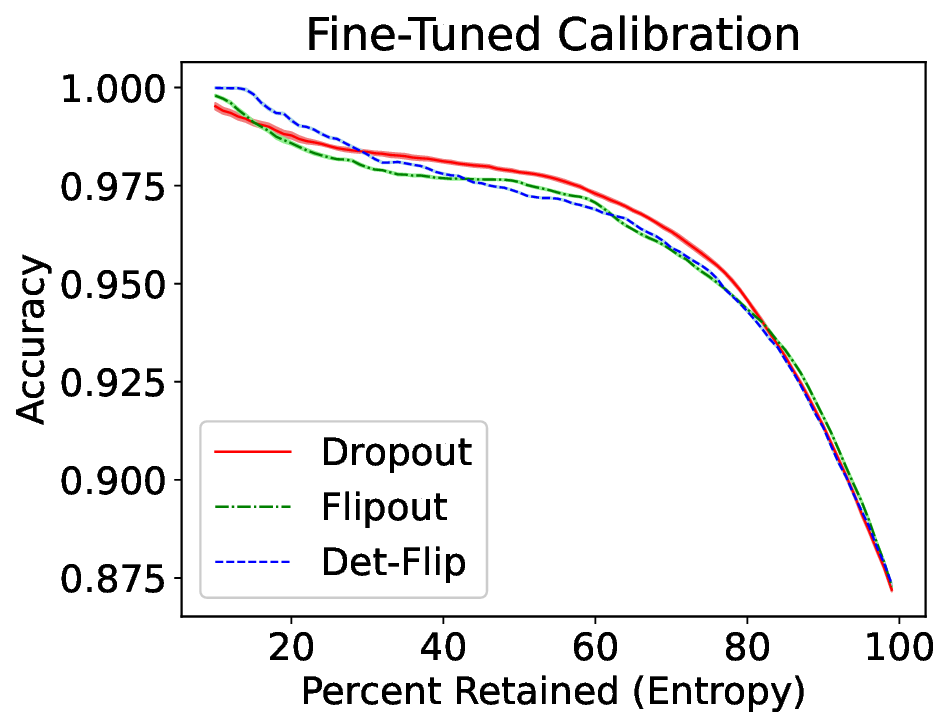}
        }
        \subfloat{
            \centering
            \includegraphics[width=.32\textwidth]{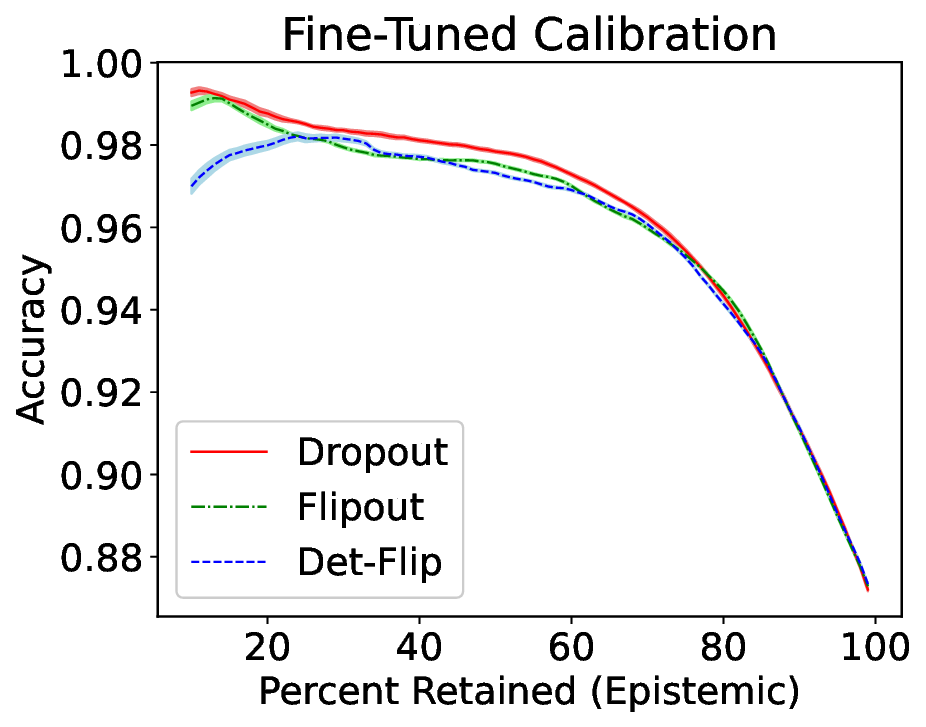}
        }
        \subfloat{
            \centering
            \includegraphics[width=.32\textwidth]{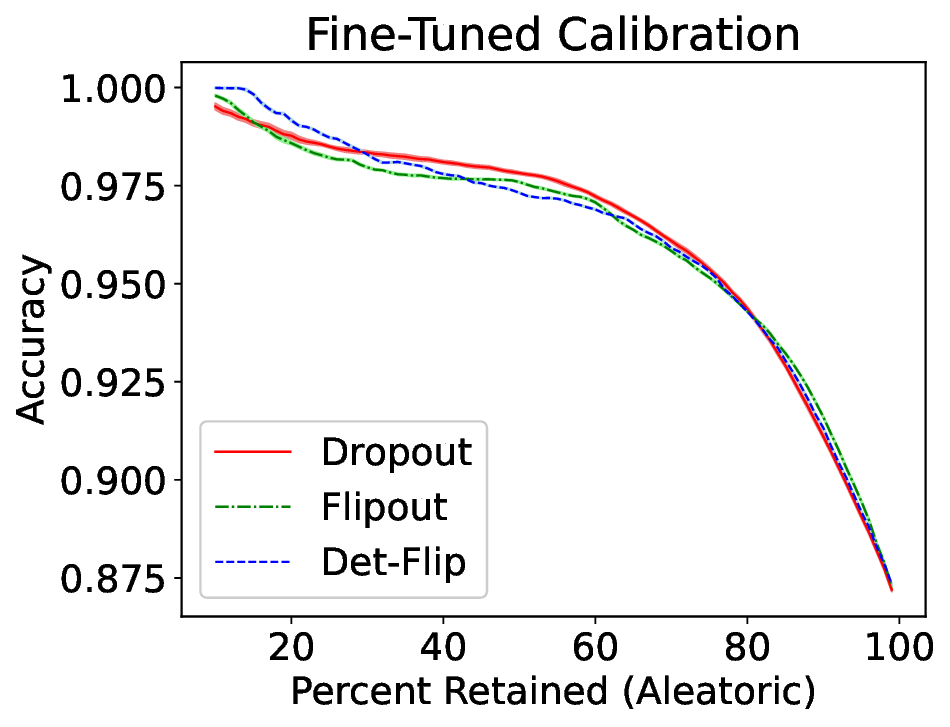}
        } \hfill
        \subfloat{
            \centering
            \includegraphics[width=.32\textwidth]{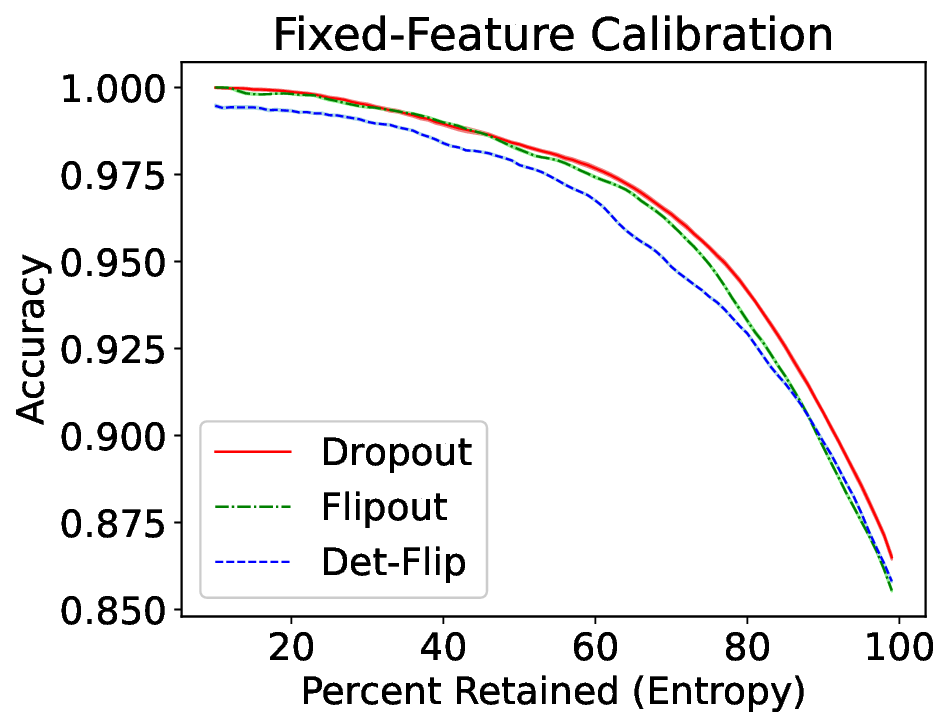}
        }
        \subfloat{
            \centering
            \includegraphics[width=.32\textwidth]{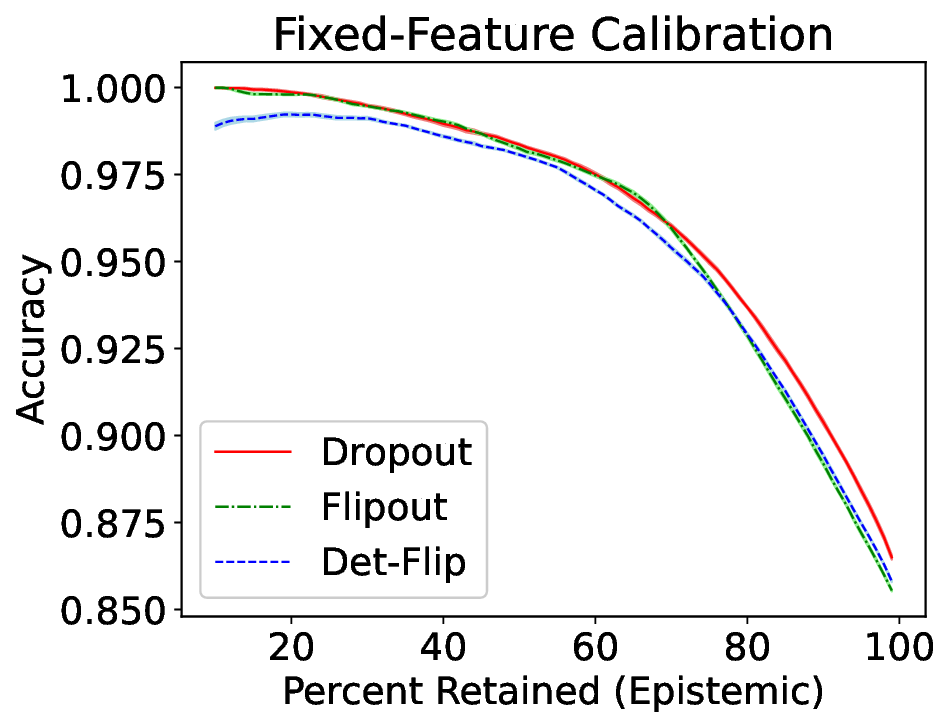}
        }
        \subfloat{
            \centering
            \includegraphics[width=.32\textwidth]{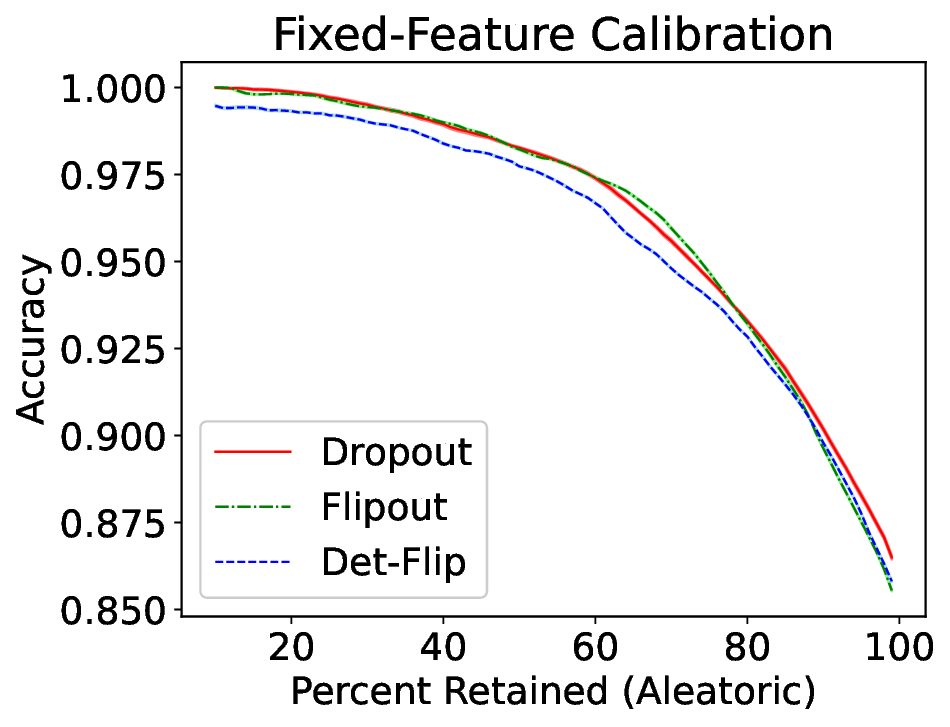}
        } \hfill

    \caption{Uncertainty calibration plots comparing Drop, Flip, and Det-Flip \ac{VI-PANN} variants on UrbanSound8k.  Comparison plots of test set accuracy vs. percentage of evaluation data retained based on Entropy (left), Epistemic Uncertainty (center) and Aleatoric Uncertainty (right).  Plots corresponding to fine-tuned models are on the top, fixed-feature model plots are on the bottom. Shading represents a 95\% CI.}  
    \label{fig:UrbanSound8kModelComparison}
\end{figure*}

\begin{figure*}[ht!]
    \centering
        \subfloat{
            \centering
            \includegraphics[width=.32\textwidth]{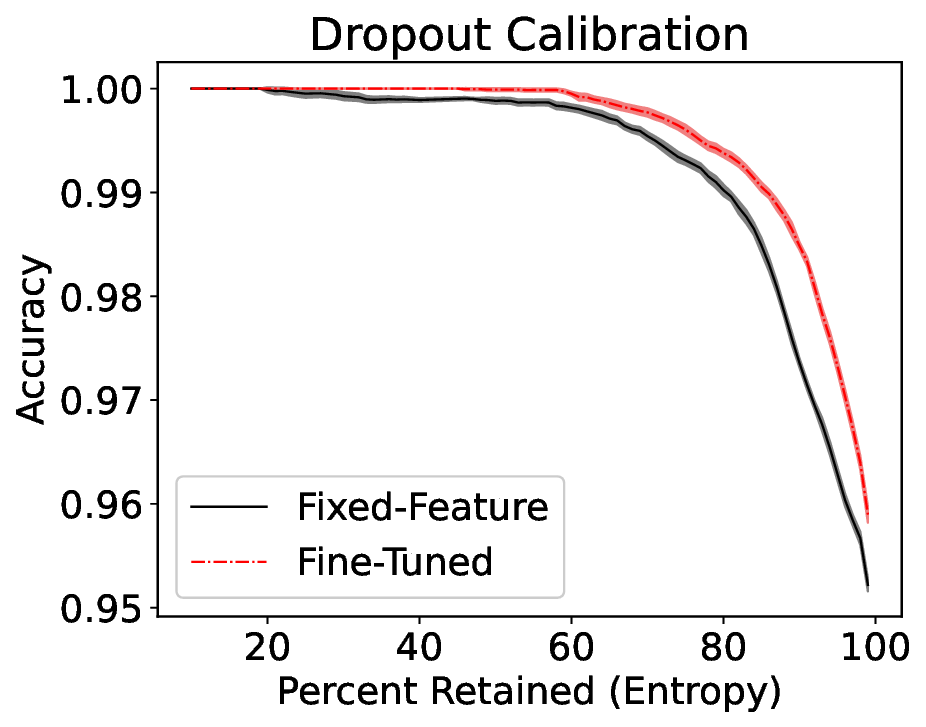}
        }
        \subfloat{
            \centering
            \includegraphics[width=.32\textwidth]{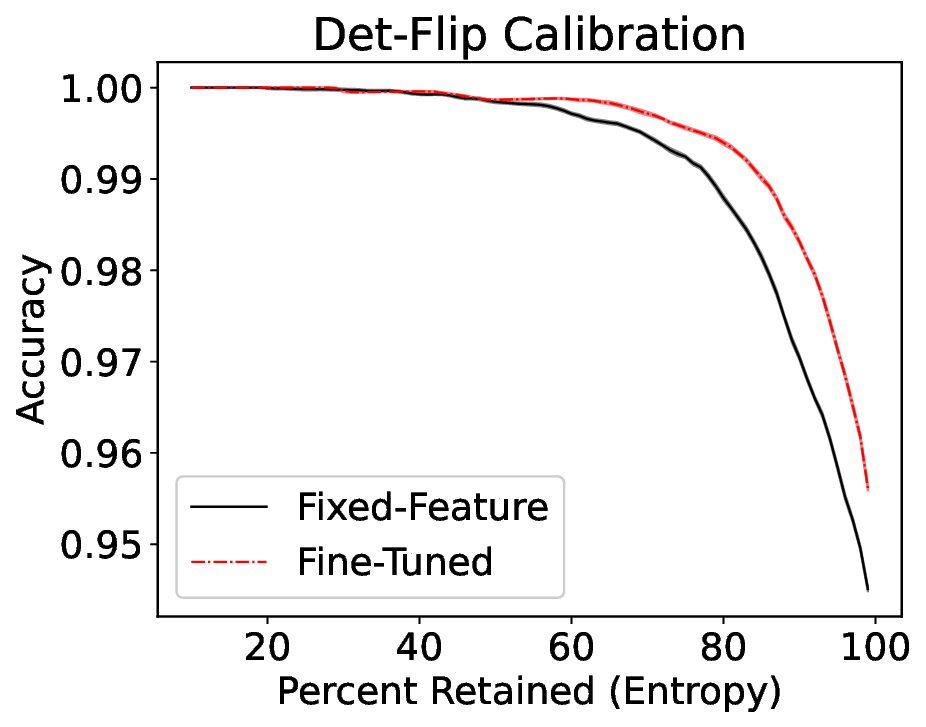}
        }
        \subfloat{
            \centering
            \includegraphics[width=.32\textwidth]{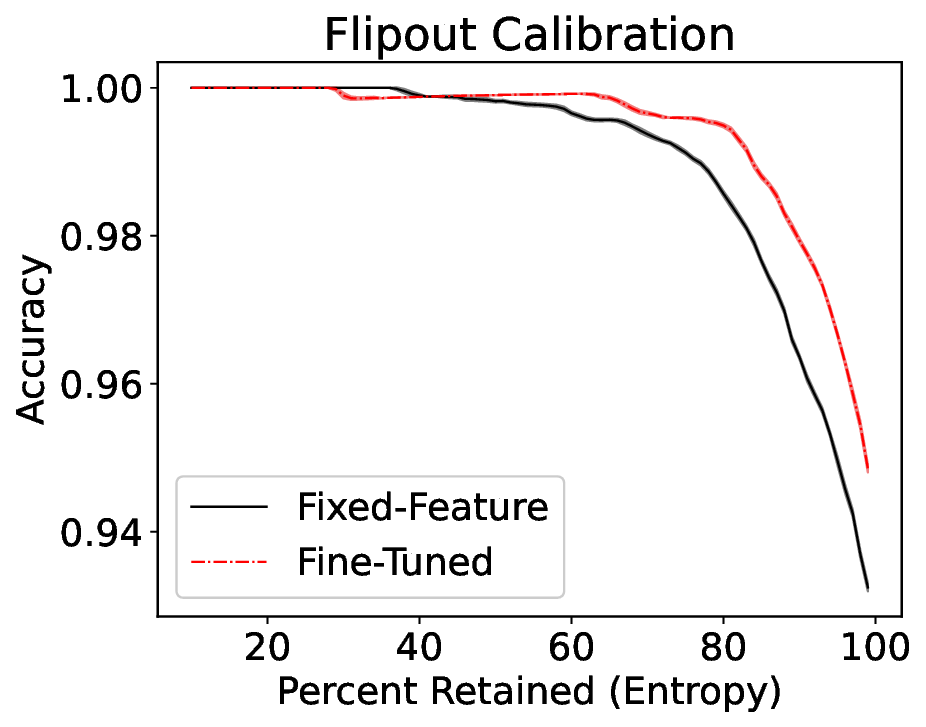}
        } \hfill
        \subfloat{
            \centering
            \includegraphics[width=.32\textwidth]{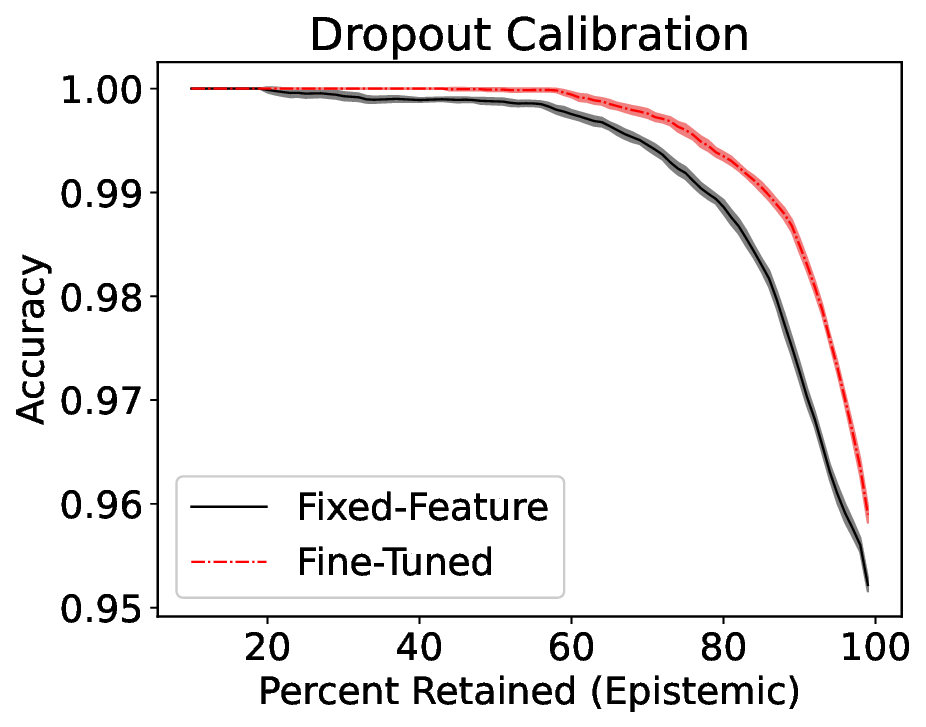}
        }
        \subfloat{
            \centering
            \includegraphics[width=.32\textwidth]{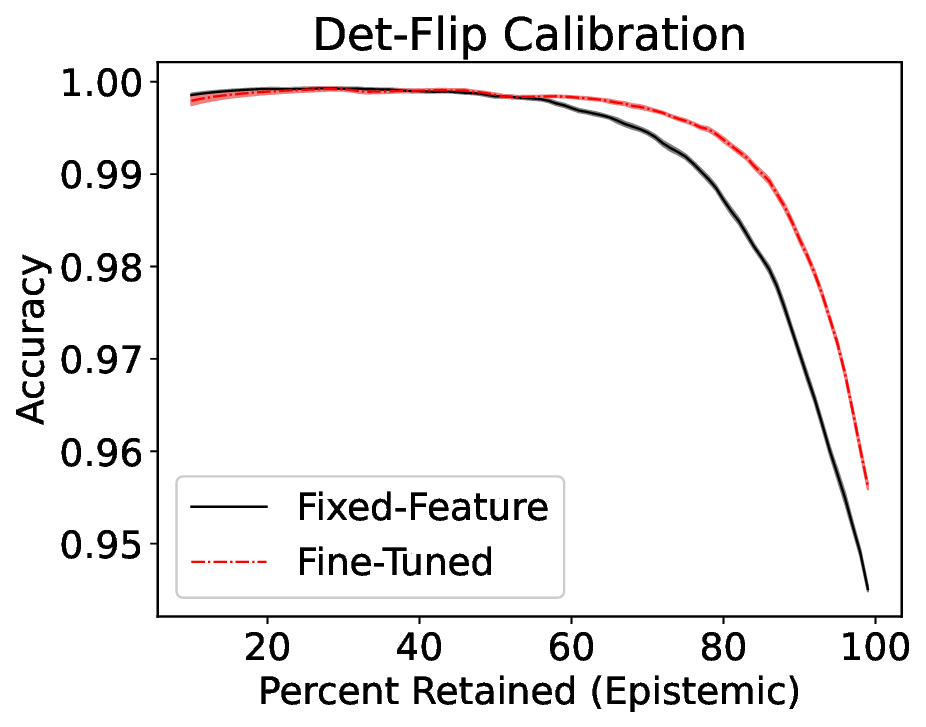}
        }
        \subfloat{
            \centering
            \includegraphics[width=.32\textwidth]{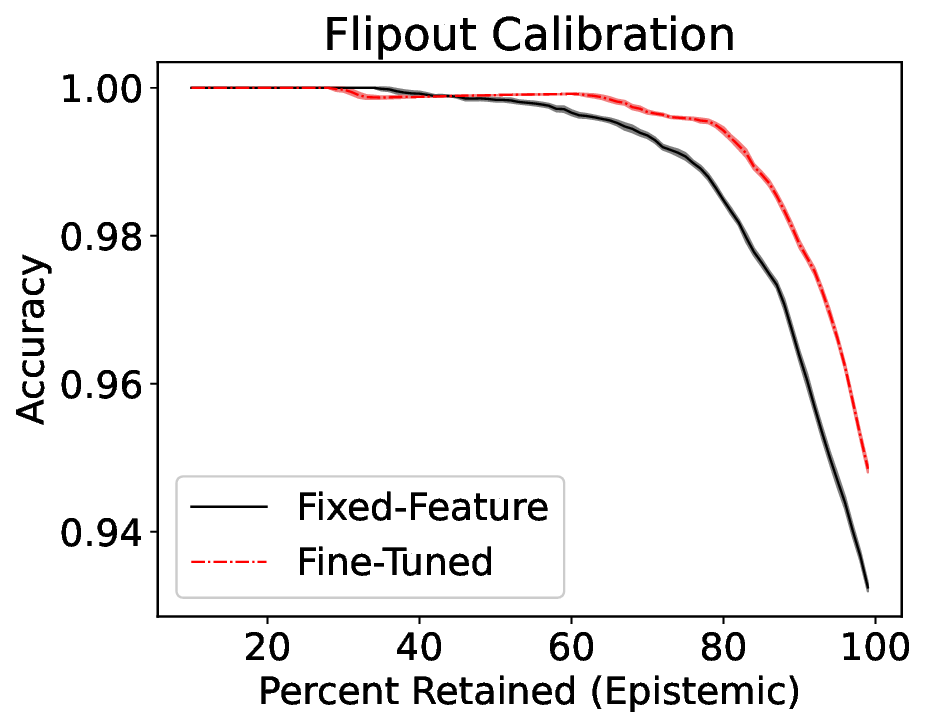}
        } \hfill
        \subfloat{
            \centering
            \includegraphics[width=.32\textwidth]{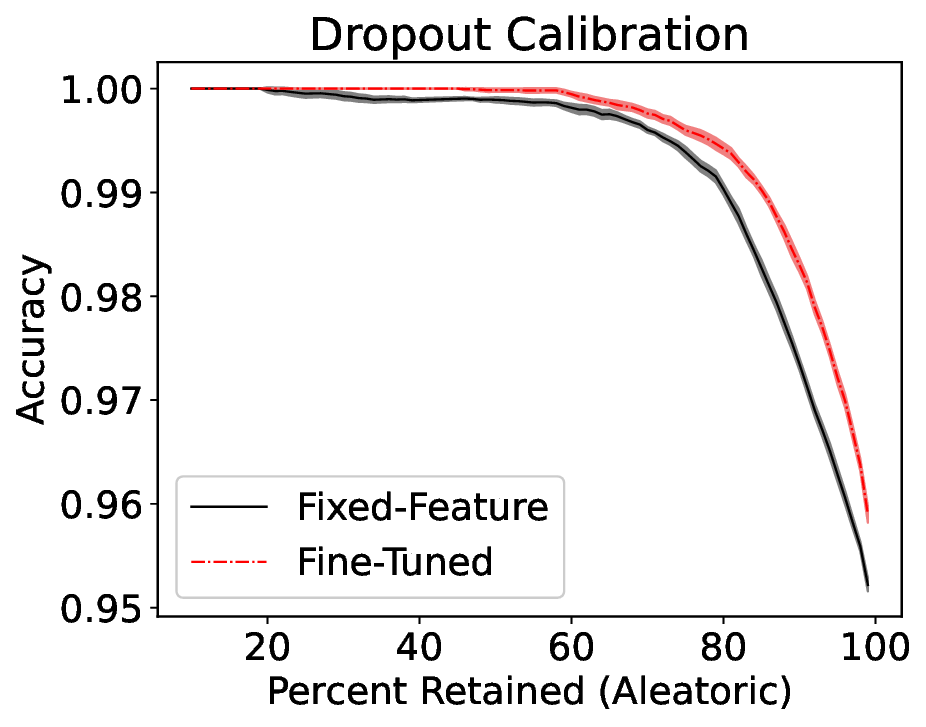}
        }
        \subfloat{
            \centering
            \includegraphics[width=.32\textwidth]{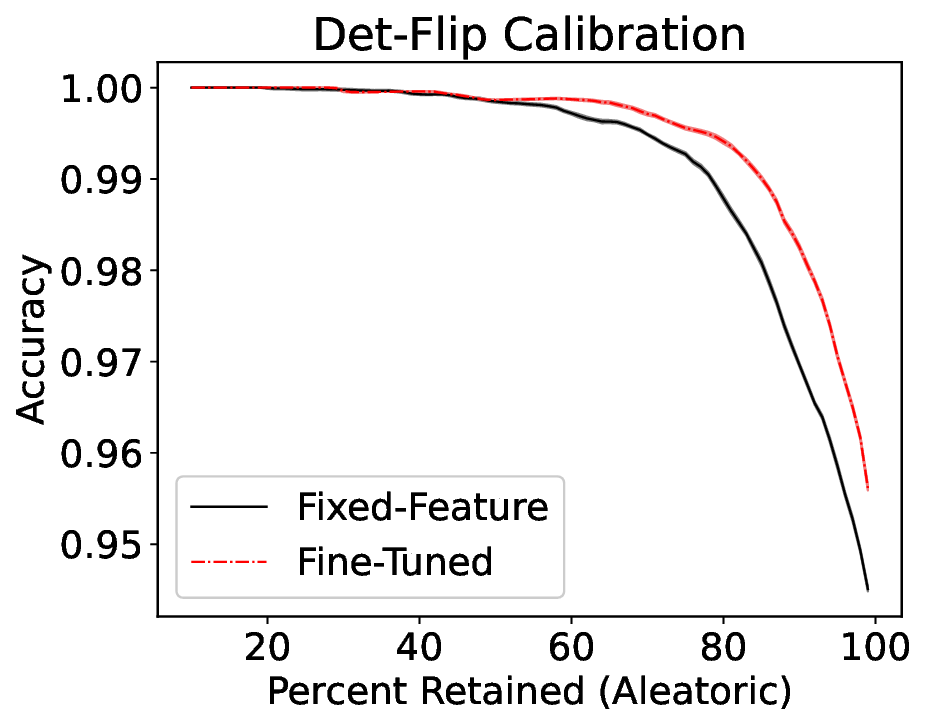}
        }
        \subfloat{
            \centering
            \includegraphics[width=.32\textwidth]{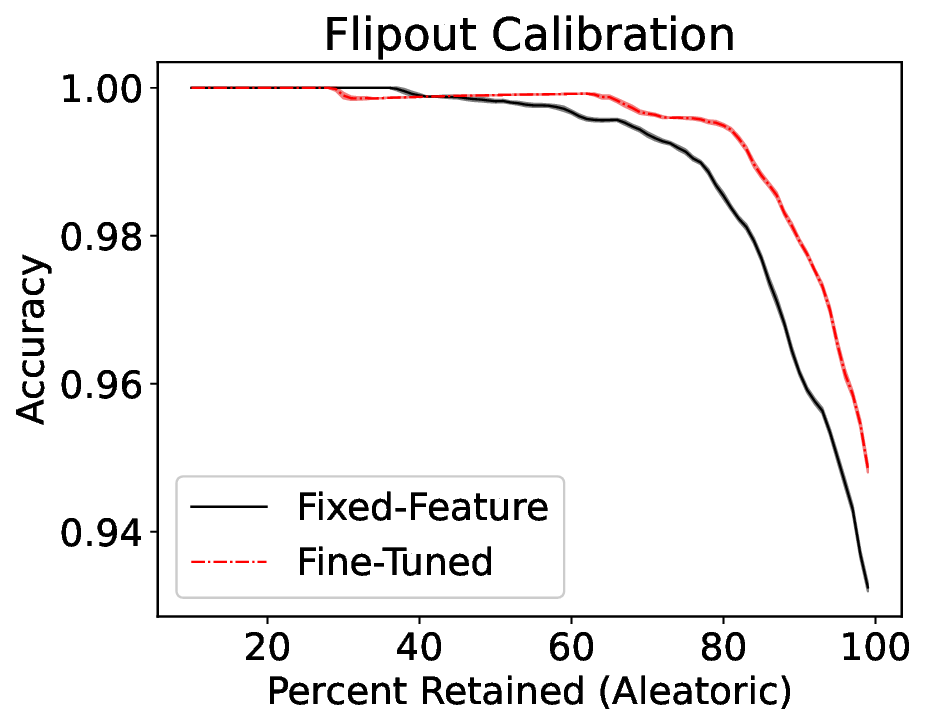}
        }

    \caption{Uncertainty calibration plots comparing fixed-feature and fine-tuning \ac{TL} techniques on ESC-50.  Comparison plots of test set accuracy vs. percentage of evaluation data retained based on Entropy (top), Epistemic Uncertainty (middle) and Aleatoric Uncertainty (bottom).  Drop \ac{VI-PANN} is on the left, Det-Flip \ac{VI-PANN} in the center, and Flip \ac{VI-PANN} on the right. Shading represents a 95\% CI.}  
    \label{fig:ESC50Calibration}
\end{figure*}

\begin{figure*}[ht!]
    \centering
        \subfloat{
            \centering
            \includegraphics[width=.32\textwidth]{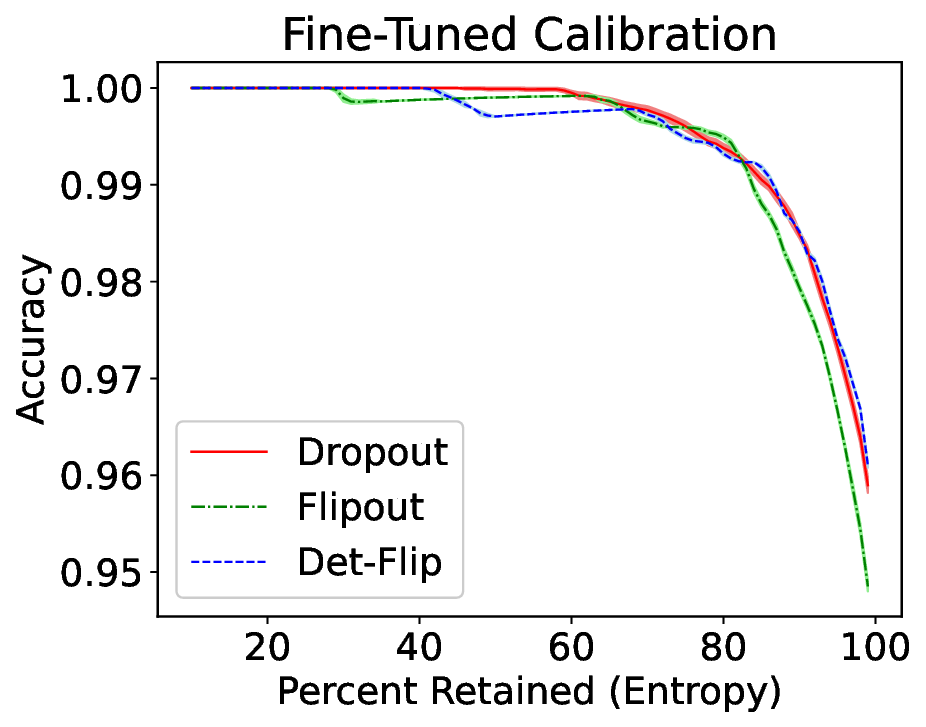}
        }
        \subfloat{
            \centering
            \includegraphics[width=.32\textwidth]{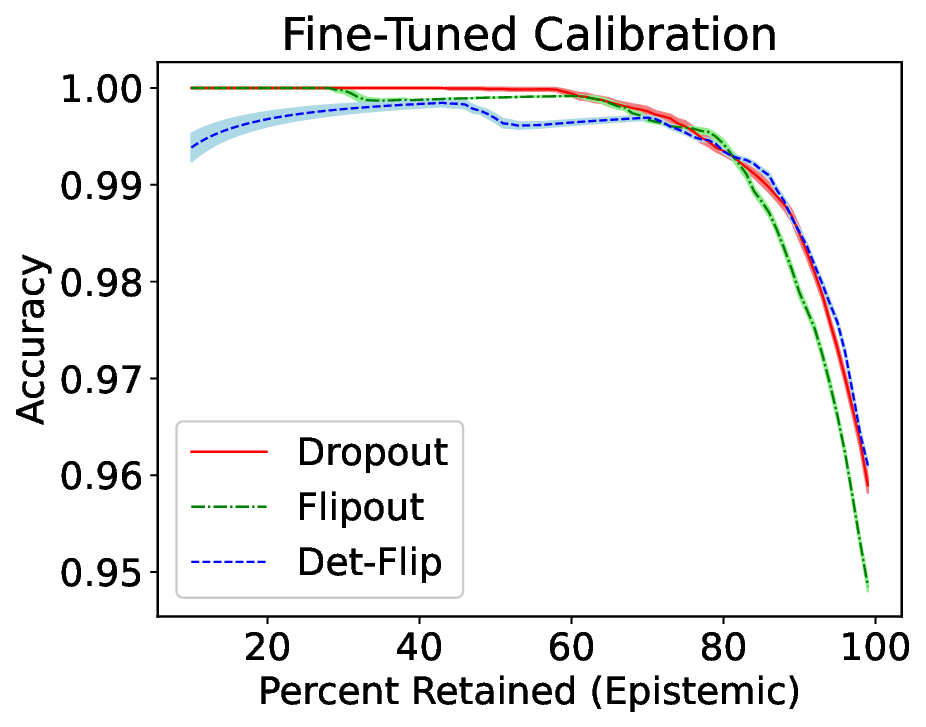}
        }
        \subfloat{
            \centering
            \includegraphics[width=.32\textwidth]{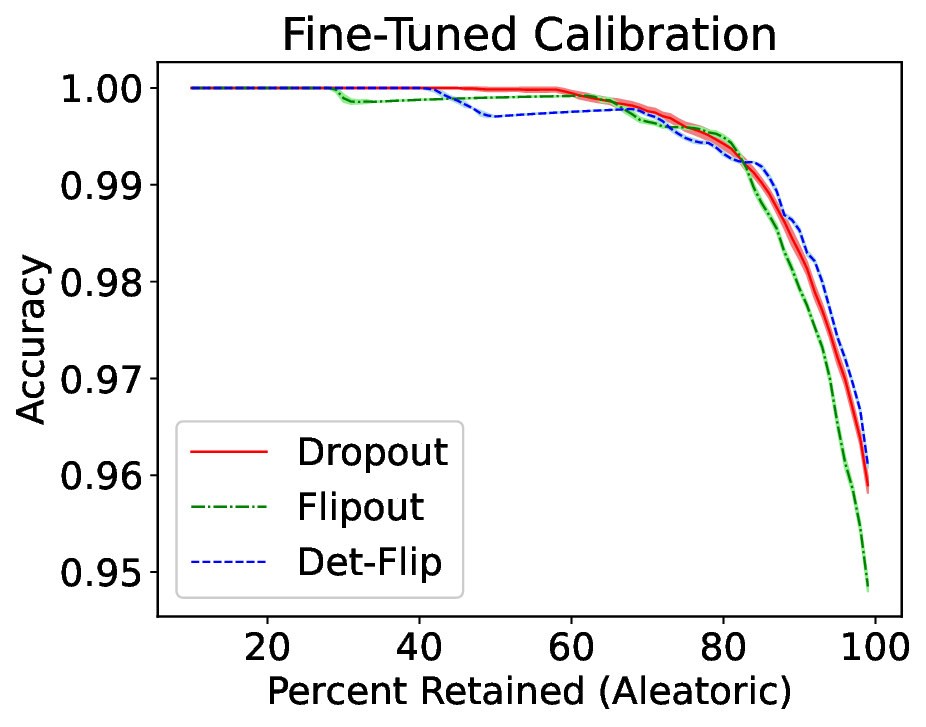}
        } \hfill
        \subfloat{
            \centering
            \includegraphics[width=.32\textwidth]{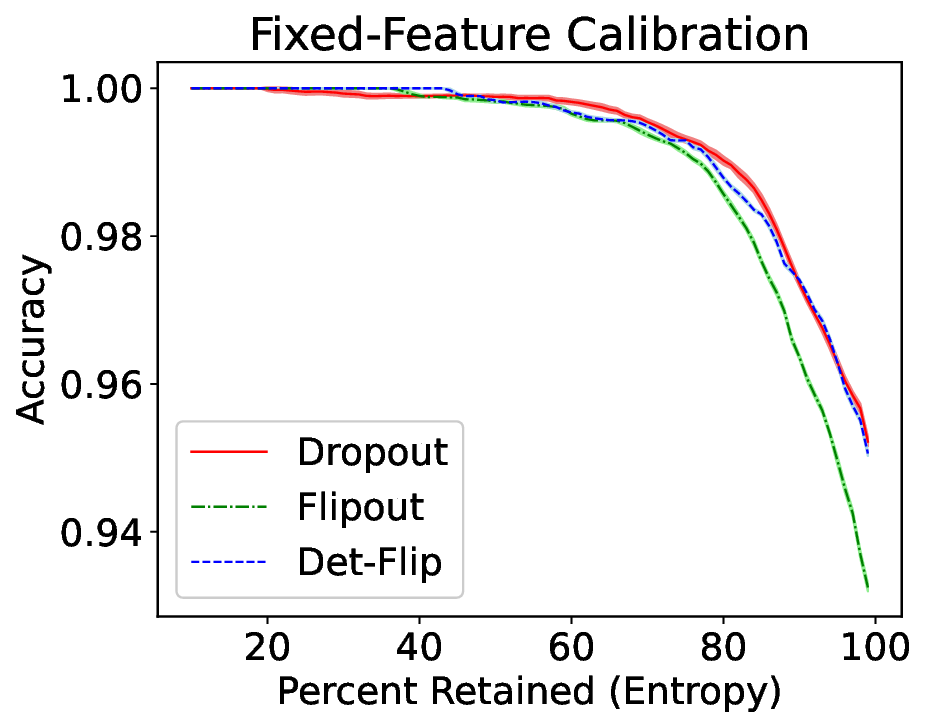}
        }
        \subfloat{
            \centering
            \includegraphics[width=.32\textwidth]{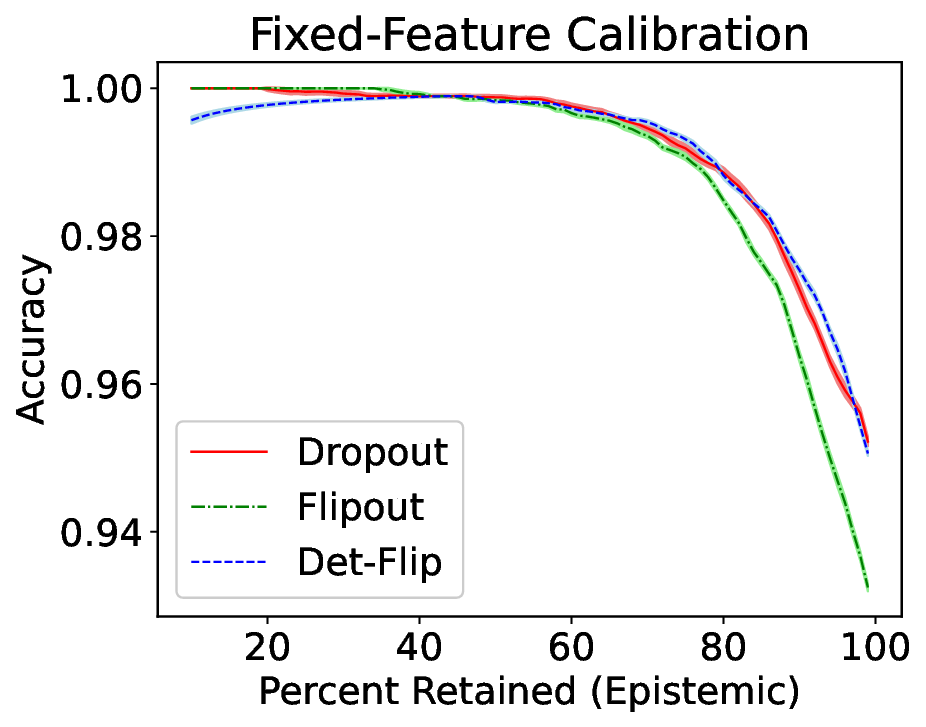}
        }
        \subfloat{
            \centering
            \includegraphics[width=.32\textwidth]{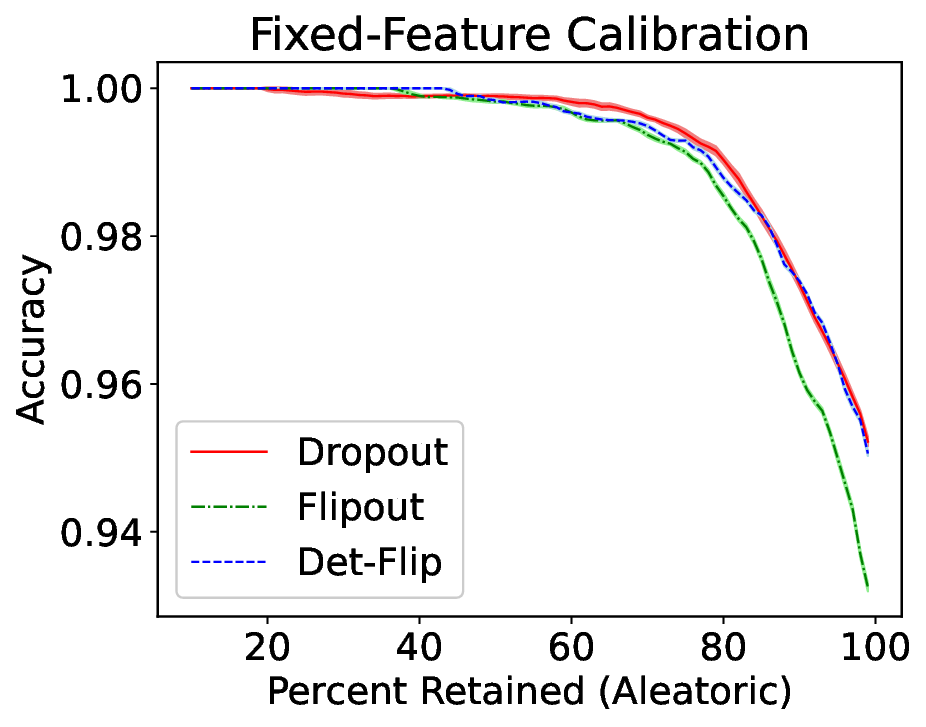}
        } \hfill

    \caption{Uncertainty calibration plots comparing Drop, Flip, and Det-Flip \ac{VI-PANN} variants on ESC-50.  Comparison plots of test set accuracy vs. percentage of evaluation data retained based on Entropy (left), Epistemic Uncertainty (center) and Aleatoric Uncertainty (right).  Plots corresponding to fine-tuned models are on the top, fixed-feature model plots are on the bottom. Shading represents a 95\% CI.}  
    \label{fig:ESC50ModelComparison}
\end{figure*}

\begin{figure*}[ht!]
    \centering
        \subfloat{
            \centering
            \includegraphics[width=.32\textwidth]{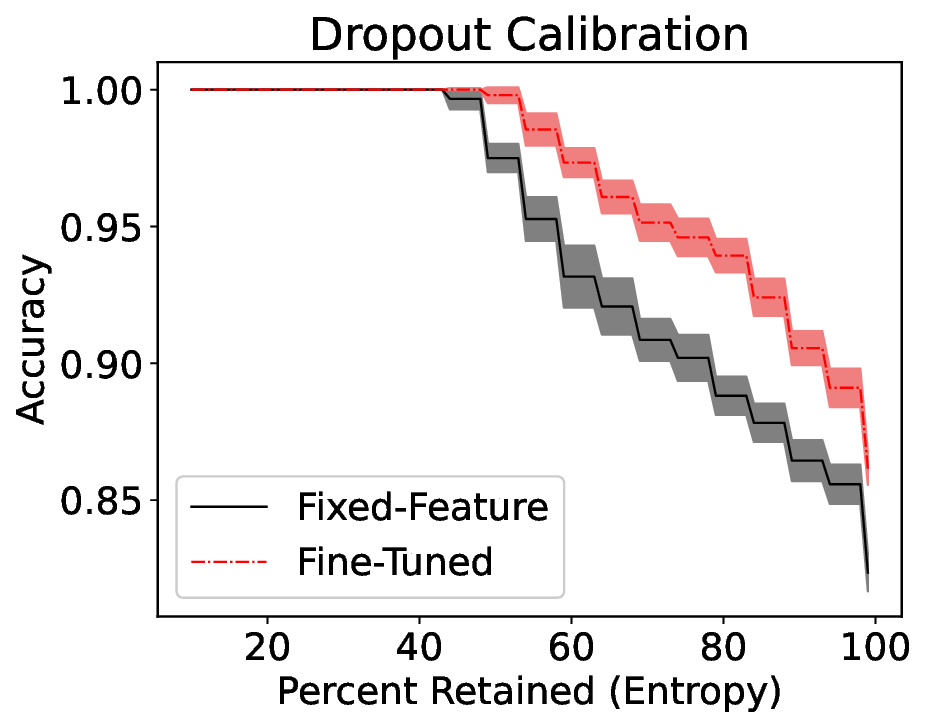}
        }
        \subfloat{
            \centering
            \includegraphics[width=.32\textwidth]{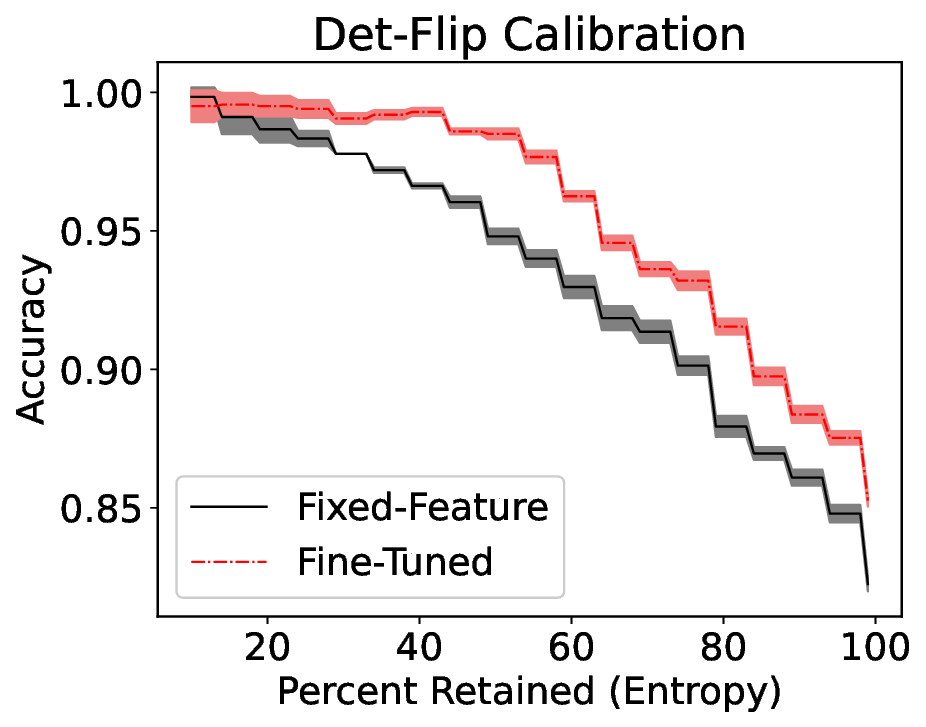}
        }
        \subfloat{
            \centering
            \includegraphics[width=.32\textwidth]{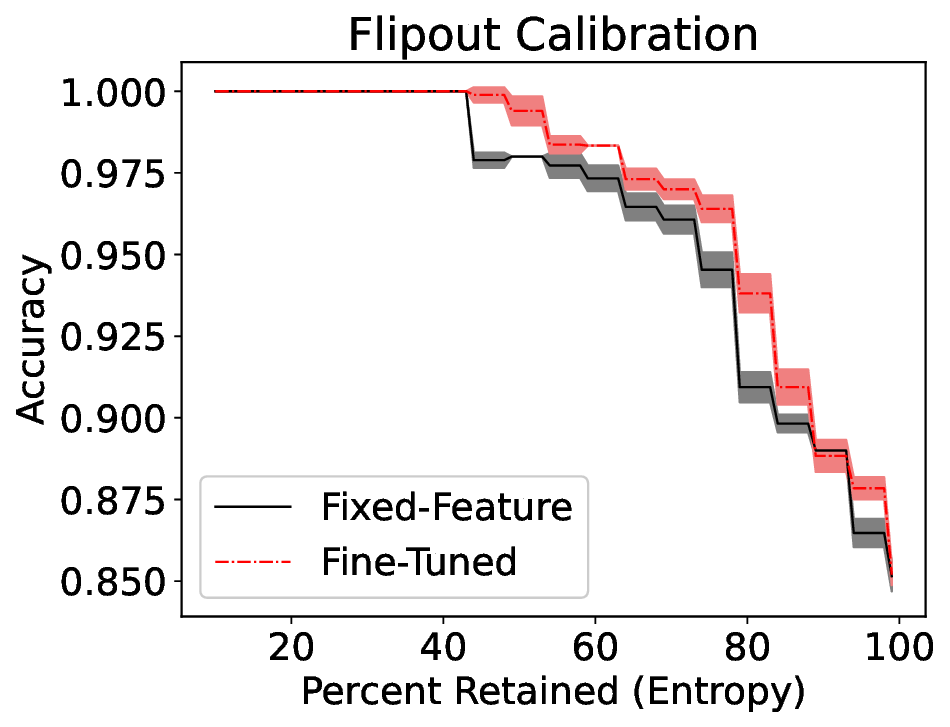}
        } \hfill
        \subfloat{
            \centering
            \includegraphics[width=.32\textwidth]{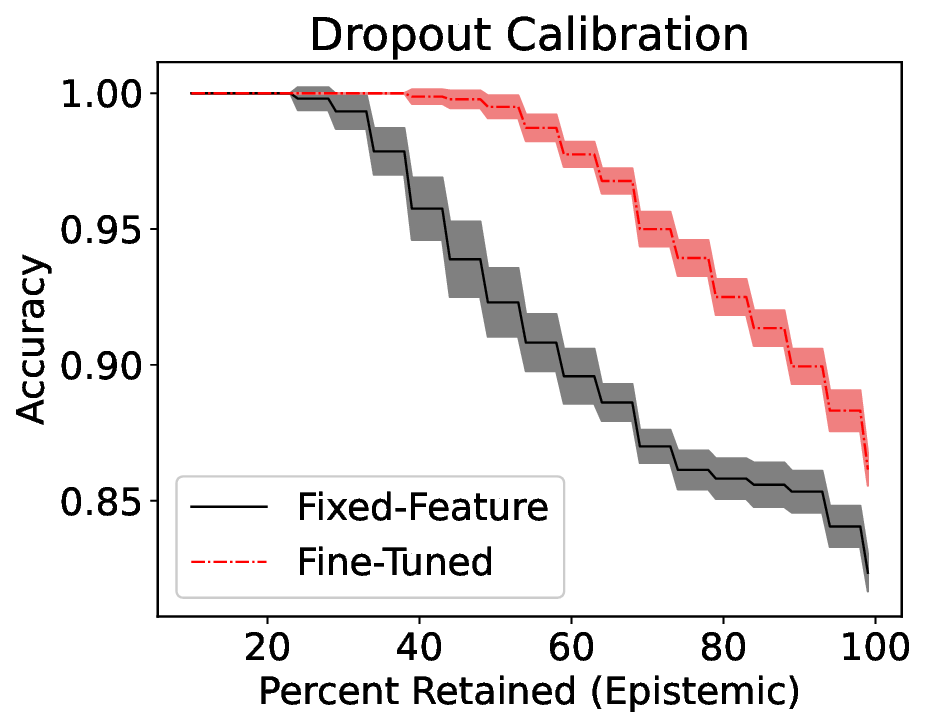}
        }
        \subfloat{
            \centering
            \includegraphics[width=.32\textwidth]{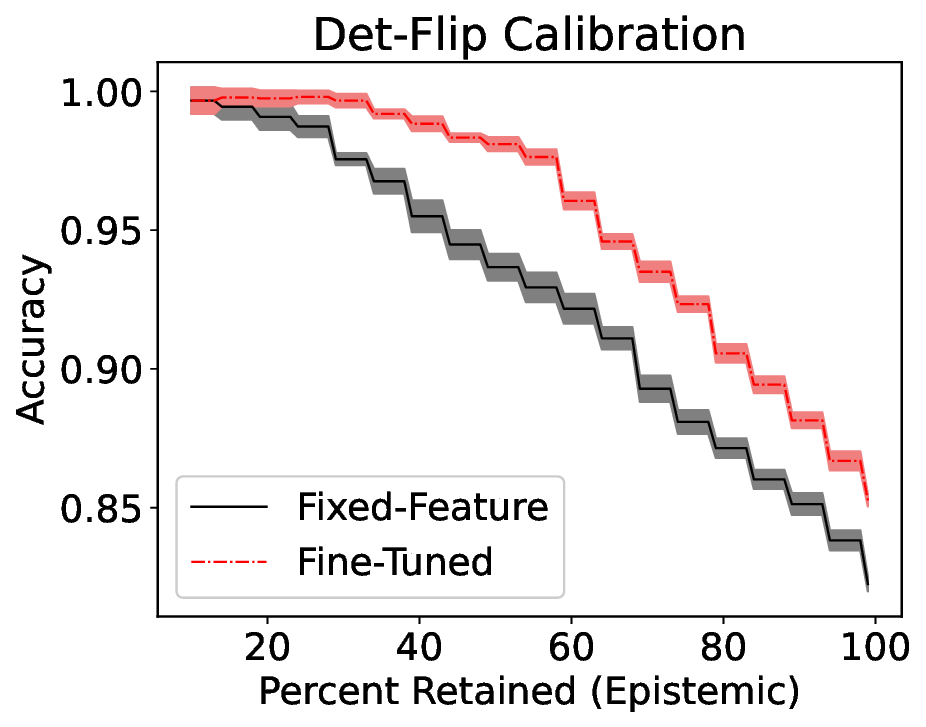}
        }
        \subfloat{
            \centering
            \includegraphics[width=.32\textwidth]{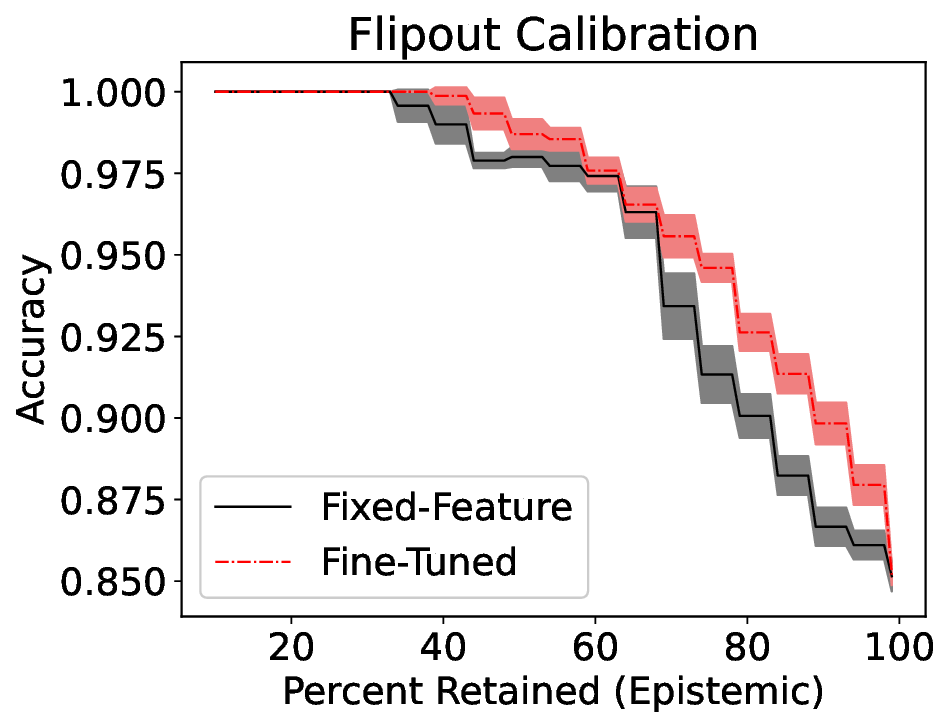}
        } \hfill
        \subfloat{
            \centering
            \includegraphics[width=.32\textwidth]{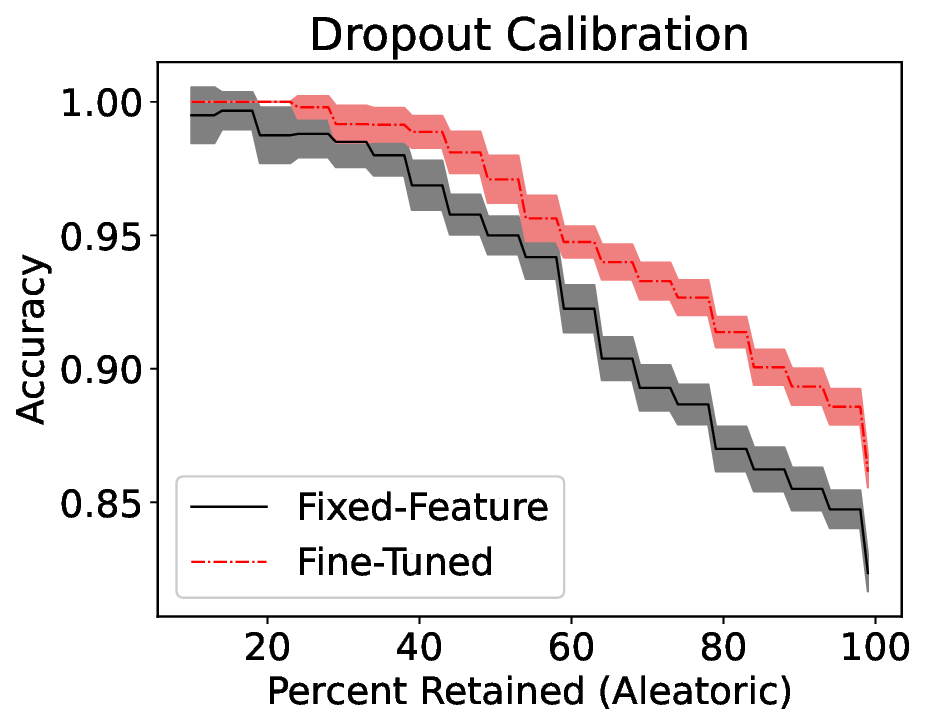}
        }
        \subfloat{
            \centering
            \includegraphics[width=.32\textwidth]{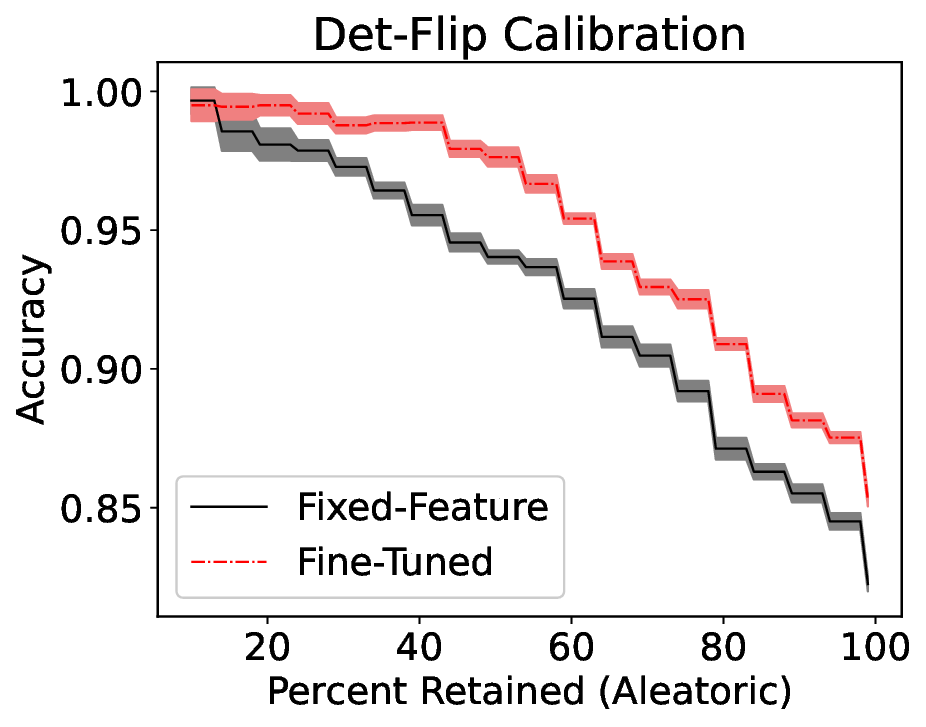}
        }
        \subfloat{
            \centering
            \includegraphics[width=.32\textwidth]{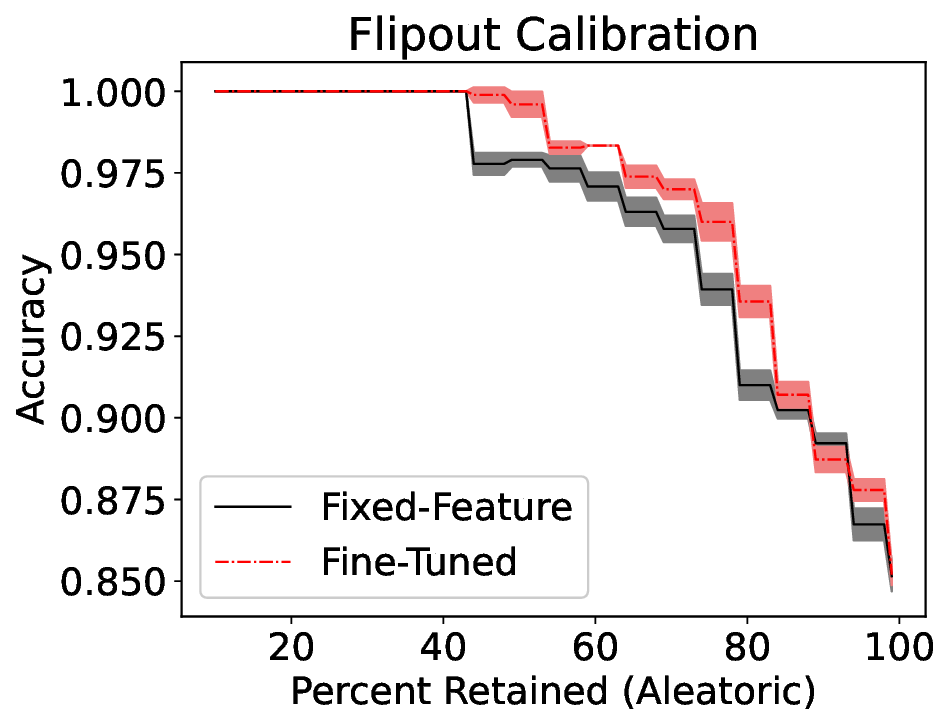}
        }

    \caption{Uncertainty calibration plots comparing fixed-feature and fine-tuning \ac{TL} techniques on DCASE2013.  Comparison plots of test set accuracy vs. percentage of evaluation data retained based on Entropy (top), Epistemic Uncertainty (middle) and Aleatoric Uncertainty (bottom).  Drop \ac{VI-PANN} is on the left, Det-Flip \ac{VI-PANN} in the center, and Flip \ac{VI-PANN} on the right. Shading represents a 95\% CI.}  
    \label{fig:DCASE2013Calibration}
\end{figure*}

\begin{figure*}[ht!]
    \centering
        \subfloat{
            \centering
            \includegraphics[width=.32\textwidth]{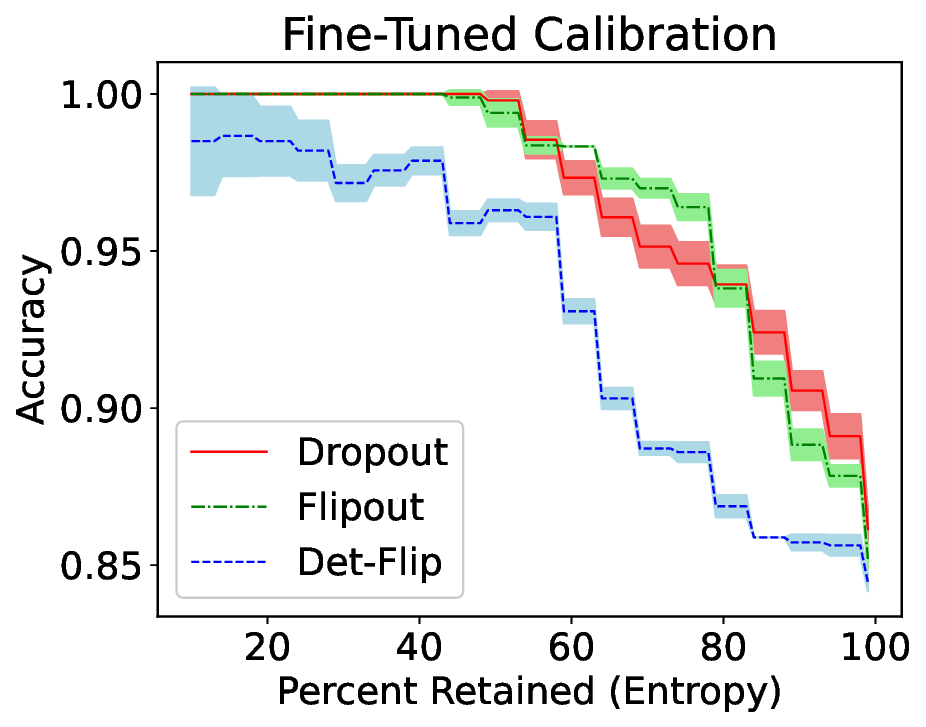}
        }
        \subfloat{
            \centering
            \includegraphics[width=.32\textwidth]{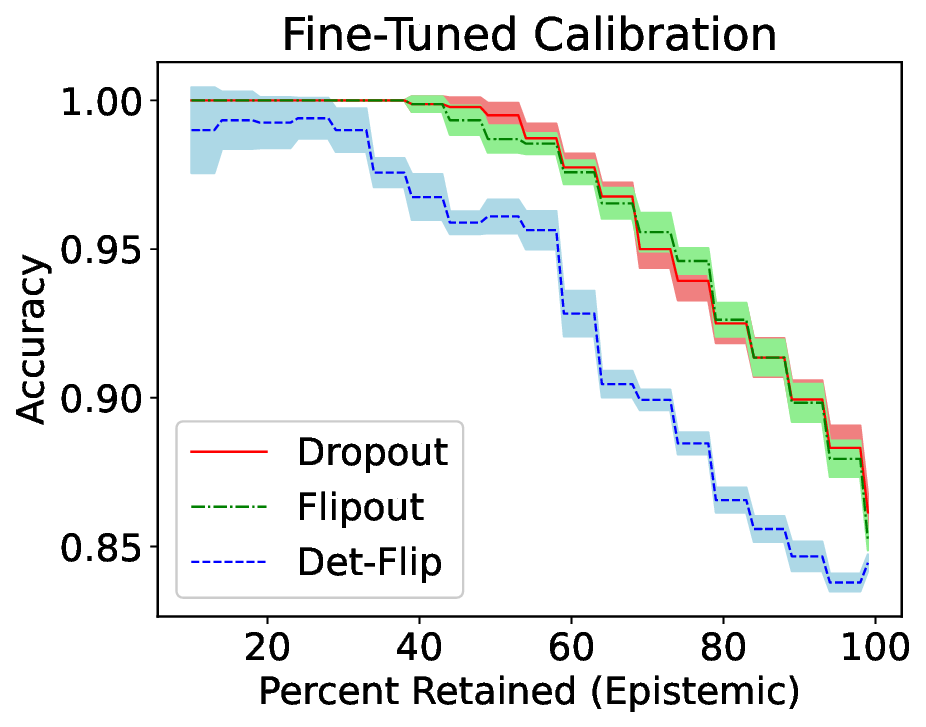}
        }
        \subfloat{
            \centering
            \includegraphics[width=.32\textwidth]{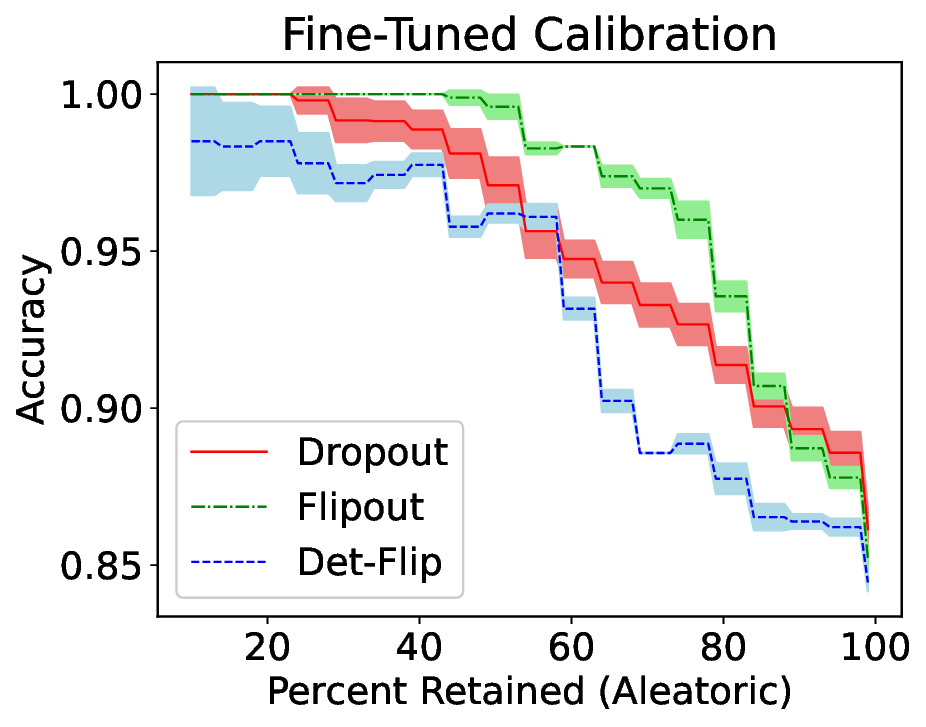}
        } \hfill
        \subfloat{
            \centering
            \includegraphics[width=.32\textwidth]{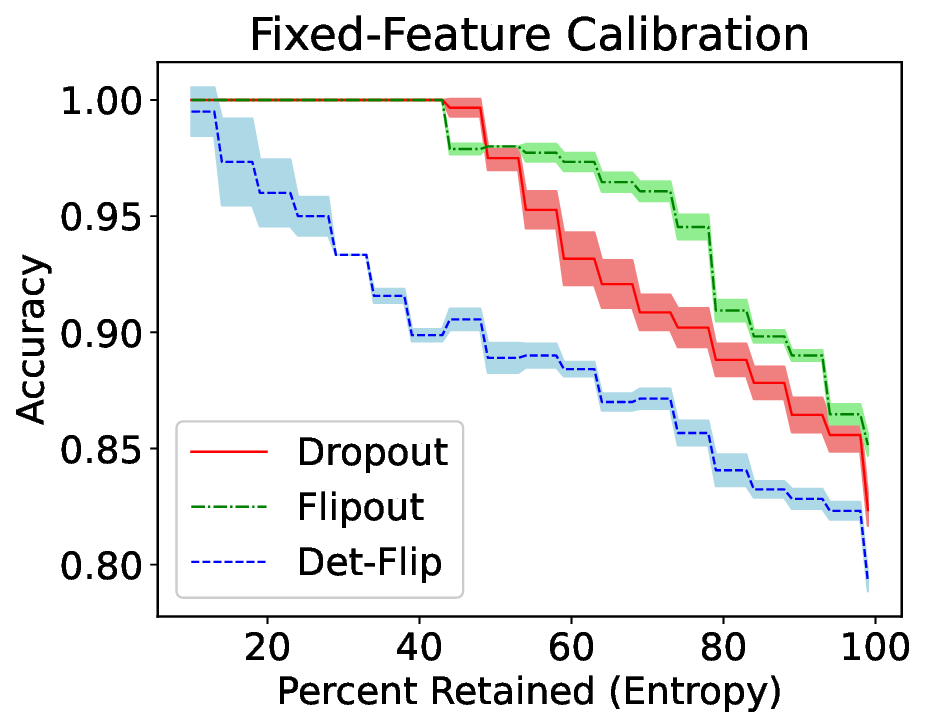}
        }
        \subfloat{
            \centering
            \includegraphics[width=.32\textwidth]{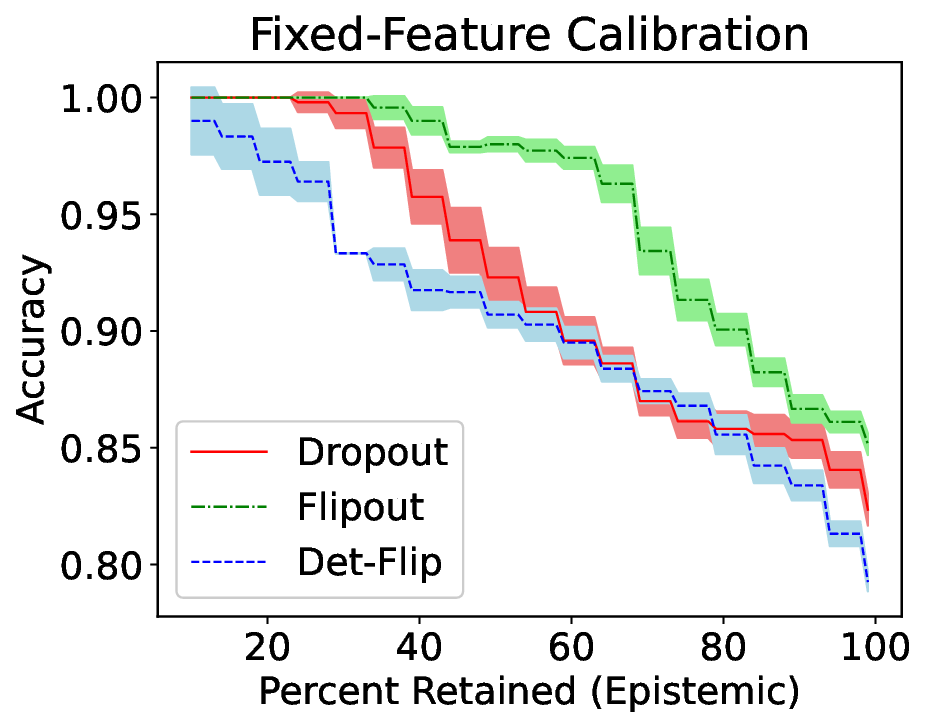}
        }
        \subfloat{
            \centering
            \includegraphics[width=.32\textwidth]{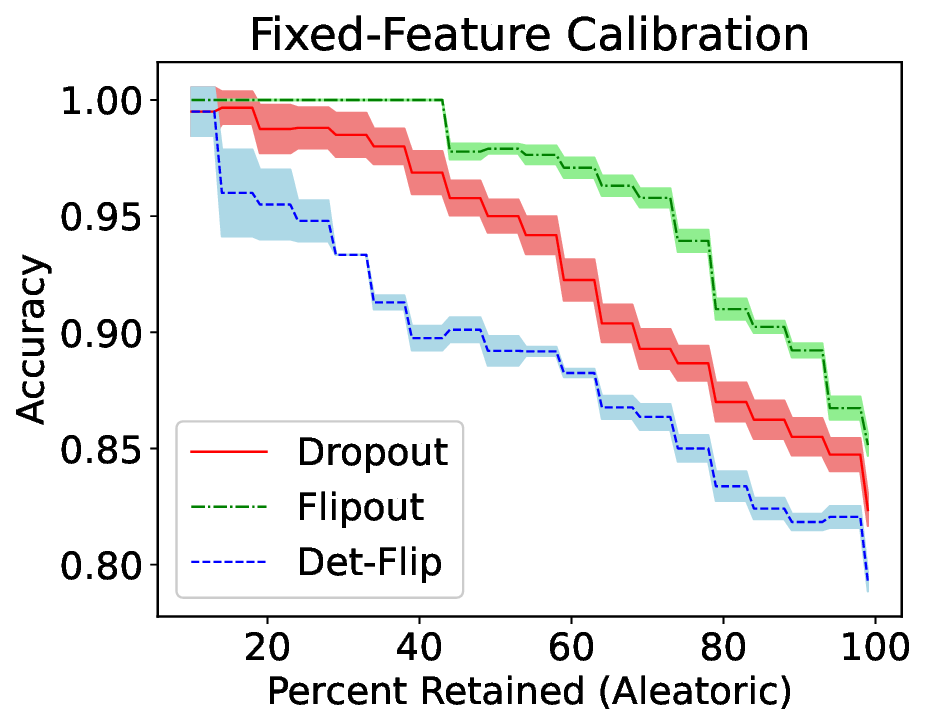}
        } \hfill

    \caption{Uncertainty calibration plots comparing Drop, Flip, and Det-Flip \ac{VI-PANN} variants on DCASE2013.  Comparison plots of test set accuracy vs. percentage of evaluation data retained based on Entropy (left), Epistemic Uncertainty (center) and Aleatoric Uncertainty (right).  Plots corresponding to fine-tuned models are on the top, fixed-feature model plots are on the bottom. Shading represents a 95\% CI.}  
    \label{fig:DCASE2013ModelComparison}
\end{figure*}

Furthermore, we provide calibration plots for UrbanSound8K (Figs. \ref{fig:UrbanSound8kCalibration} and \ref{fig:UrbanSound8kModelComparison}), ESC-50 (Figs. \ref{fig:ESC50Calibration} and \ref{fig:ESC50ModelComparison}), and DCASE2013 (Figs. \ref{fig:DCASE2013Calibration} and \ref{fig:DCASE2013ModelComparison}). These plots reveal that, following \ac{TL}, when fixing the features of the base model and after fine-tuning, all three variants of our \acp{VI-PANN} result in well-calibrated models. The stairstep pattern evident in the DCASE2013 plots is attributed to the comparatively small size of the DCASE2013 dataset.

In each of the UrbanSound8k calibration plots, the curve of the \ac{TL} model learned from fixed-features starts at a lower accuracy and crosses over that of the \ac{TL} model fine-tuned, achieving a higher accuracy at the same percentage of data retained. These results suggest that fine-tuning may have an adverse effect on model calibration if care is not taken in the fine-tuning process. Intuitively, one might extend the hyperparameter evaluation of fine-tuning and model selection to include calibration curve.

\begin{figure*}[ht!]
    \centering
        \subfloat{
            \centering
            \includegraphics[width=.32\textwidth]{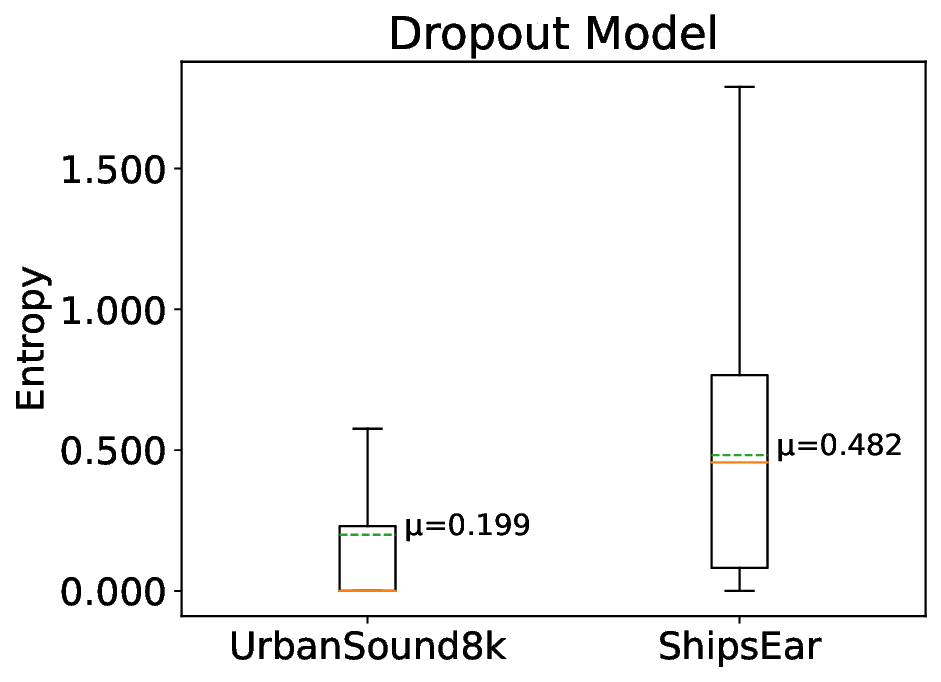}
        }
        \subfloat{
            \centering
            \includegraphics[width=.32\textwidth]{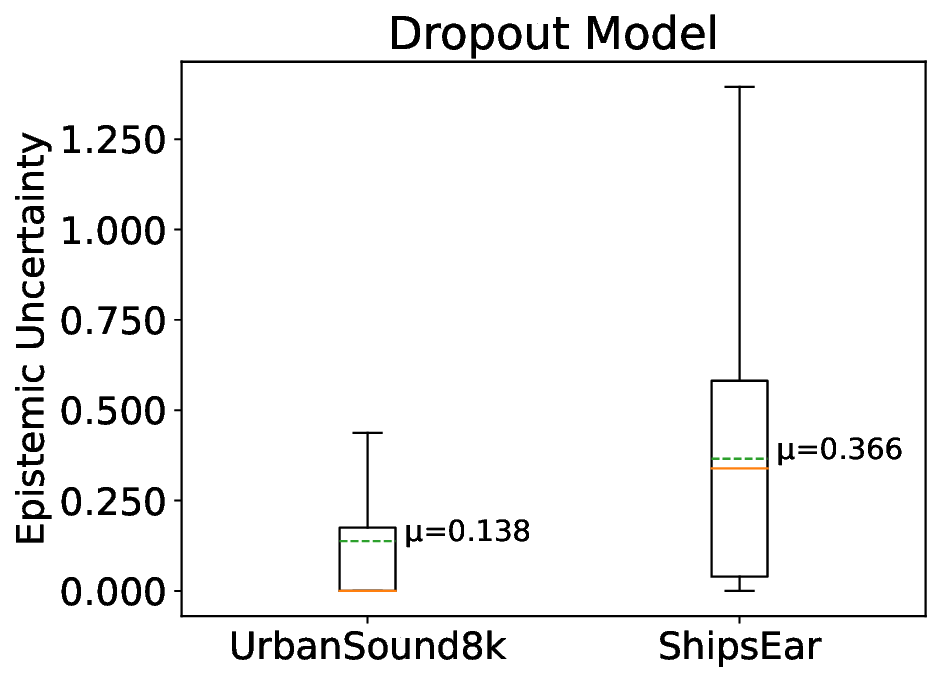}
        }
        \subfloat{
            \centering
            \includegraphics[width=.32\textwidth]{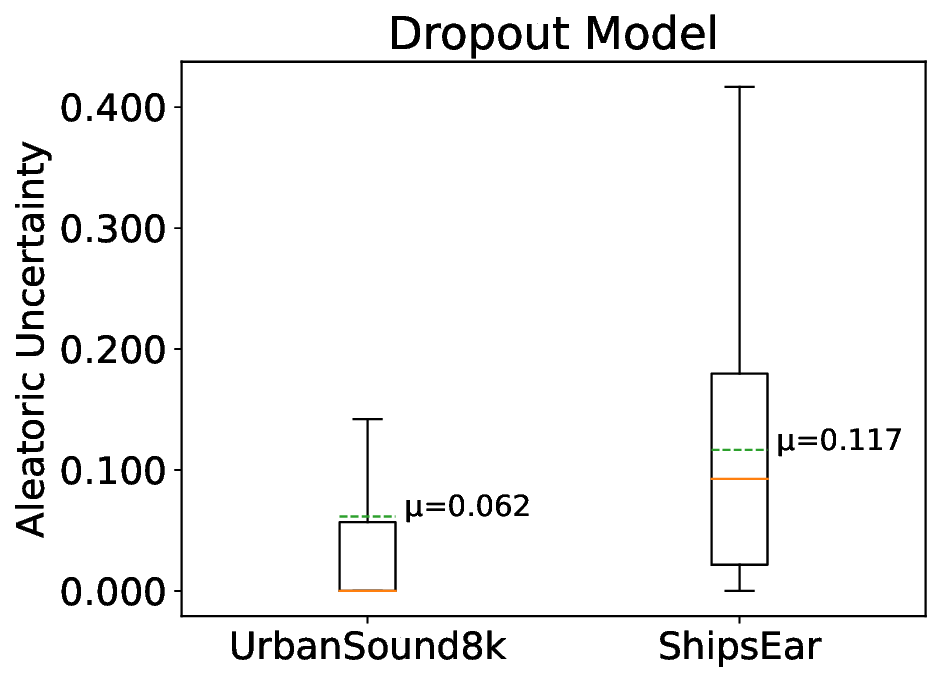}
        } \hfill
        \subfloat{
            \centering
            \includegraphics[width=.32\textwidth]{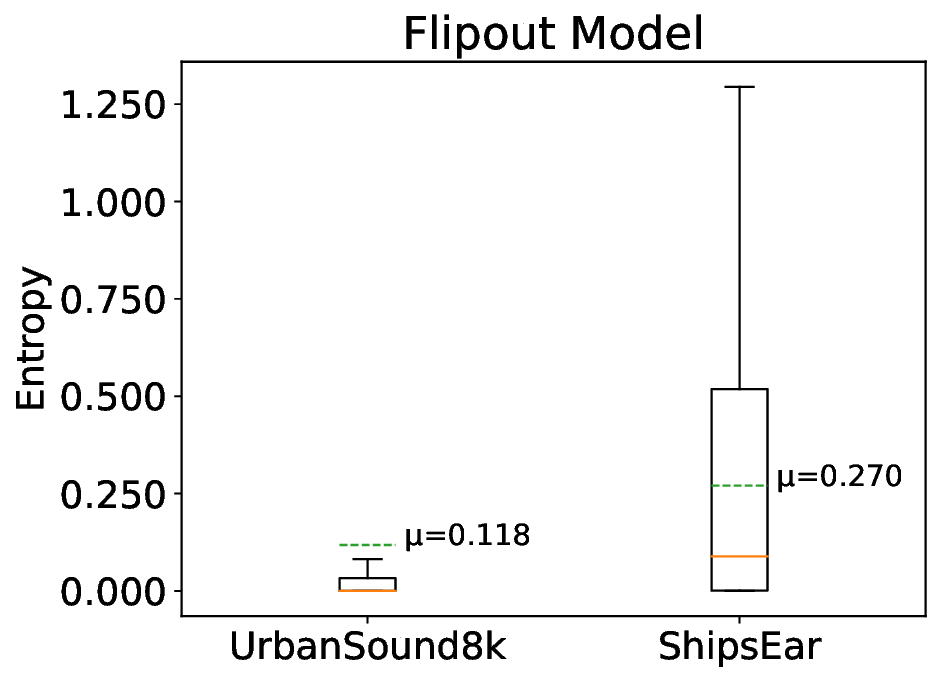}
        }
        \subfloat{
            \centering
            \includegraphics[width=.32\textwidth]{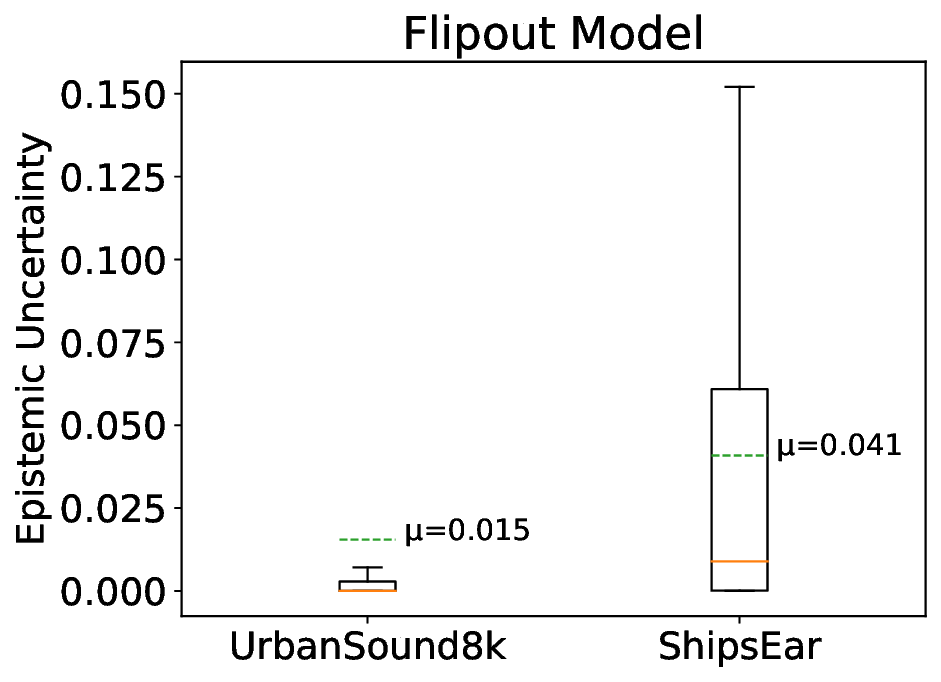}
        }
        \subfloat{
            \centering
            \includegraphics[width=.32\textwidth]{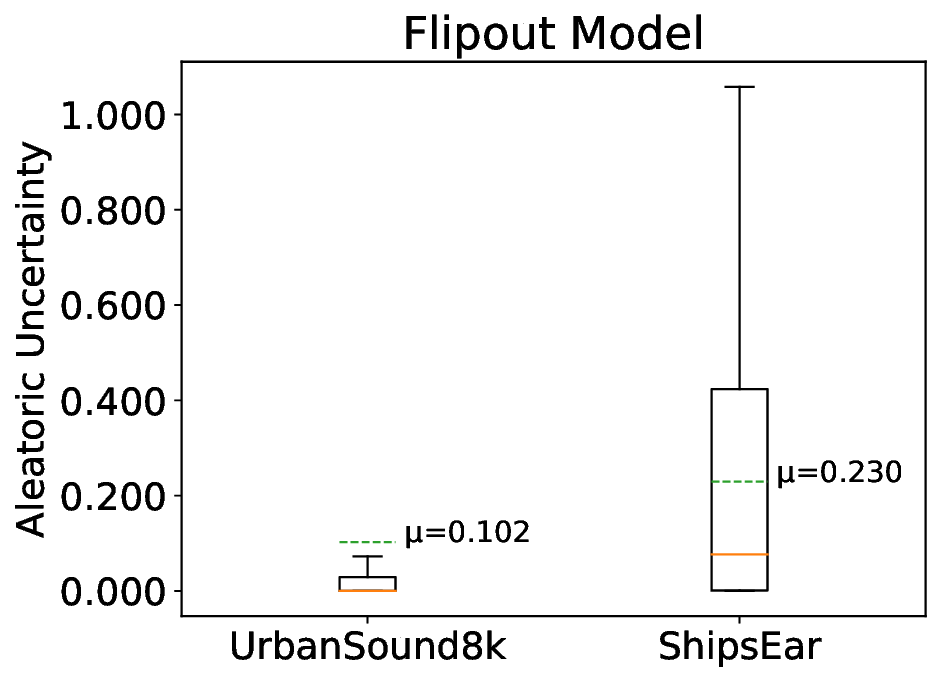}
        } \hfill
        \subfloat{
            \centering
            \includegraphics[width=.32\textwidth]{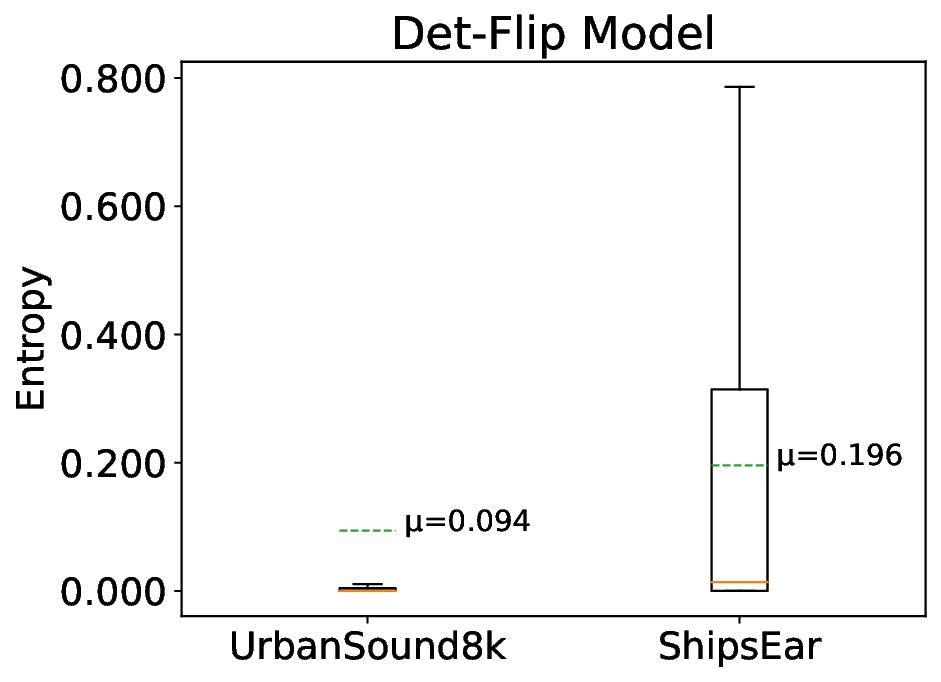}
        }
        \subfloat{
            \centering
            \includegraphics[width=.32\textwidth]{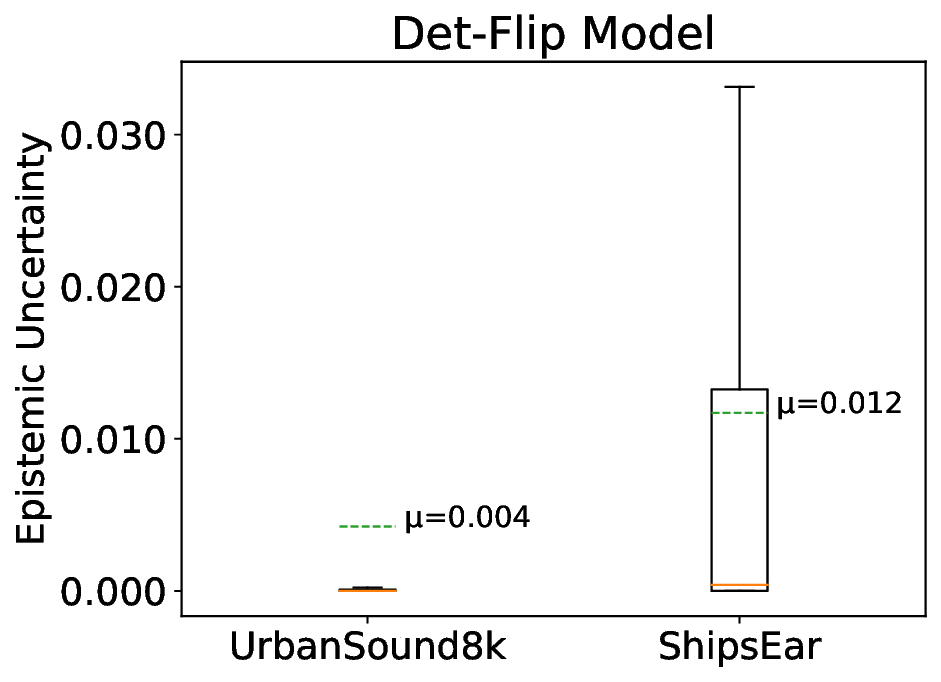}
        }
        \subfloat{
            \centering
            \includegraphics[width=.32\textwidth]{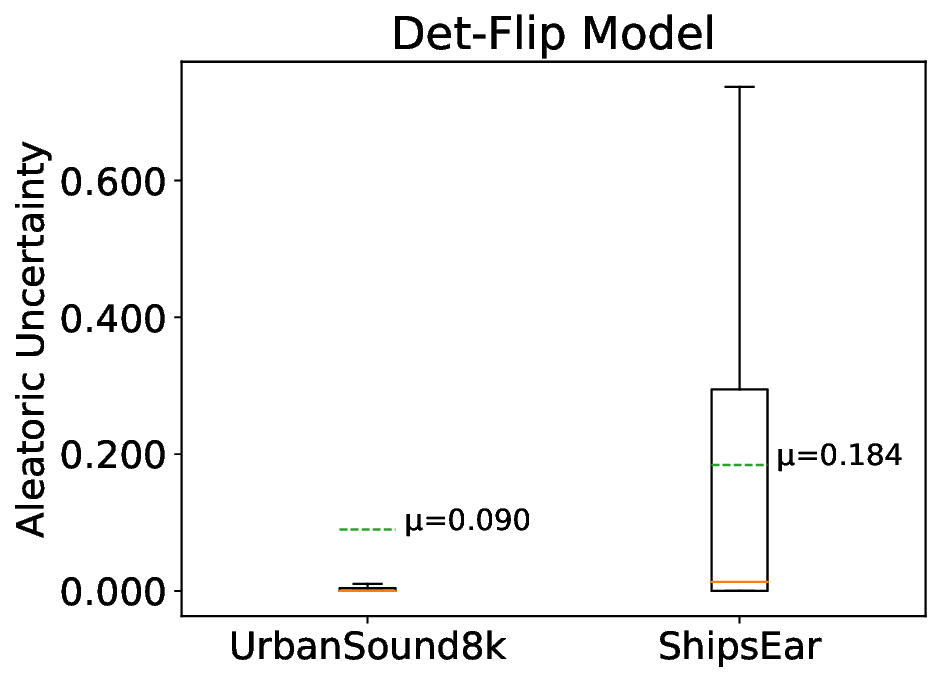}
        }

    \caption{Uncertainty box plots depicting results of \ac{MC} Dropout (top row), Flipout (middle row), and Det-Flip (bottom row) fine tuned on UrbanSound8k. The plots compare predictive entropy (left), epistemic uncertainty (middle), and aleatoric uncertainty (right) as the models are evaluated on both UrbanSound8k and the ShipsEar dataset. Both the median (orange line) and mean (dashed green line) are presented.}  
    \label{fig:UrbanSound8kBoxPlots}
\end{figure*}

\begin{figure*}[ht!]
    \centering
        \subfloat{
            \centering
            \includegraphics[width=.32\textwidth]{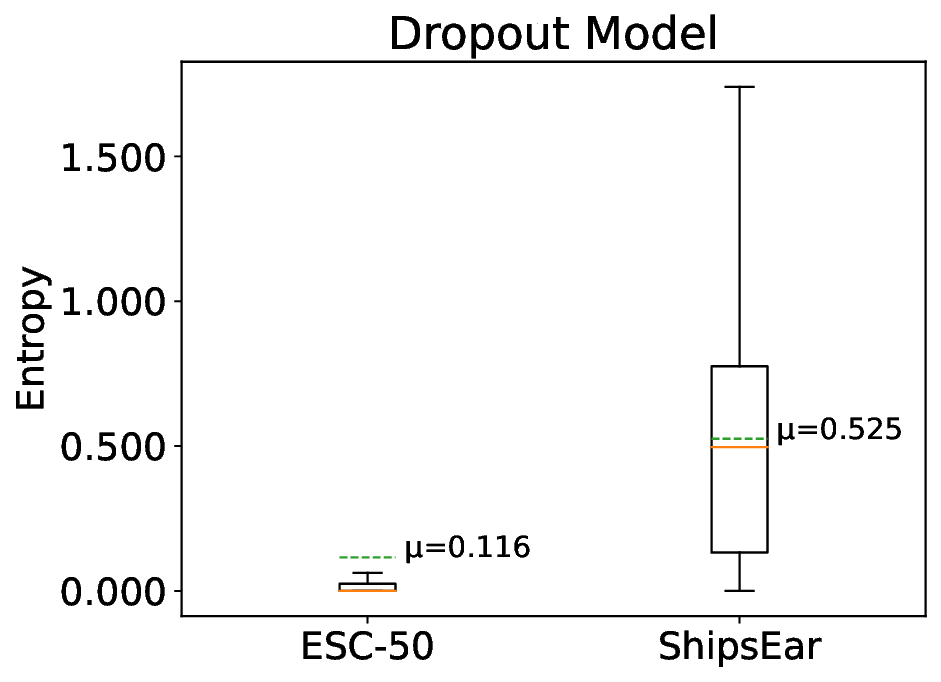}
        }
        \subfloat{
            \centering
            \includegraphics[width=.32\textwidth]{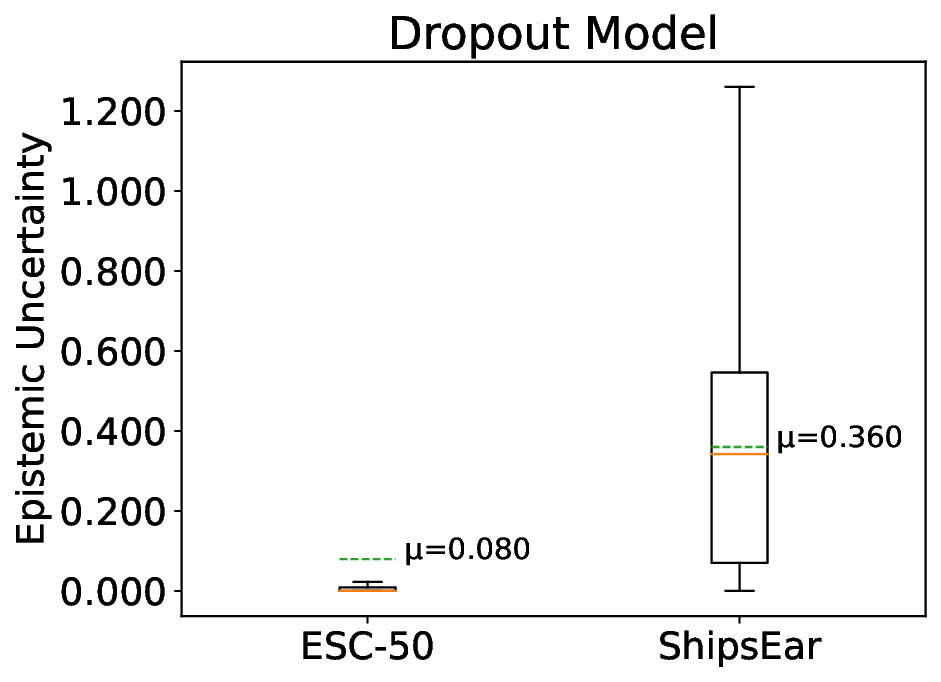}
        }
        \subfloat{
            \centering
            \includegraphics[width=.32\textwidth]{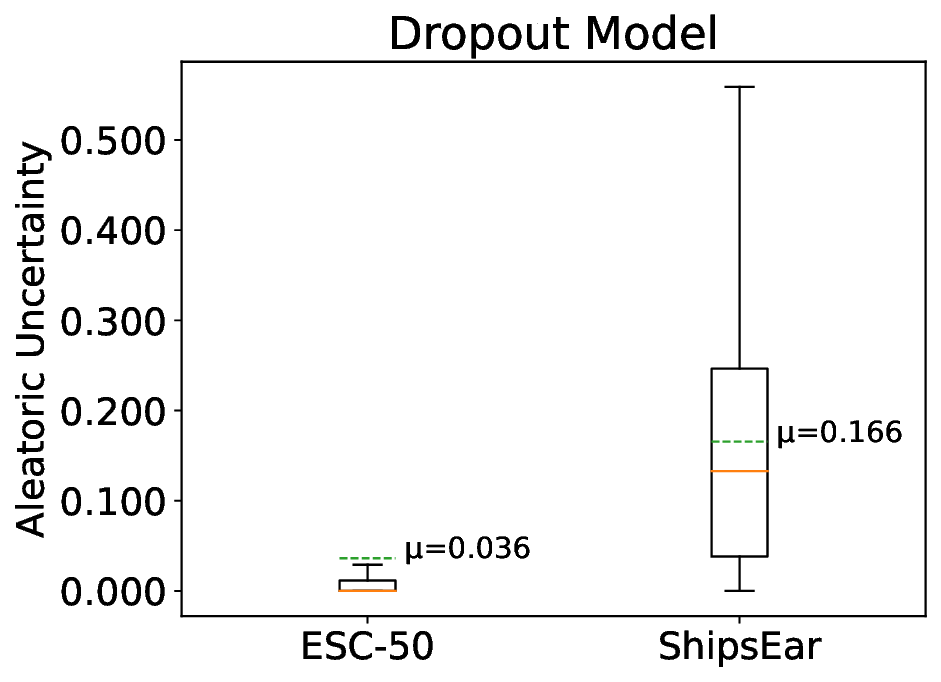}
        } \hfill
        \subfloat{
            \centering
            \includegraphics[width=.32\textwidth]{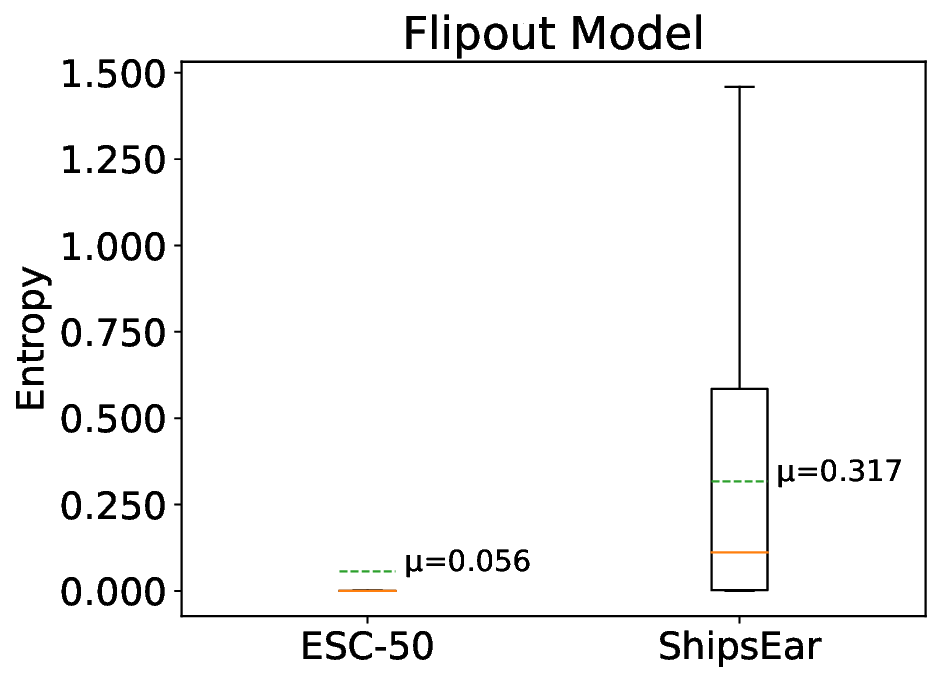}
        }
        \subfloat{
            \centering
            \includegraphics[width=.32\textwidth]{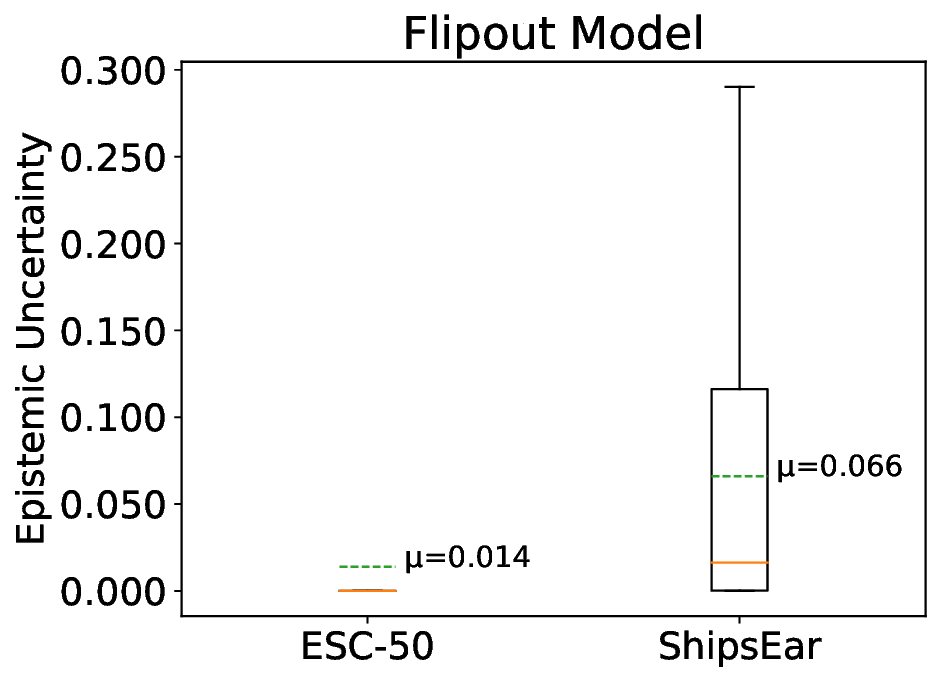}
        }
        \subfloat{
            \centering
            \includegraphics[width=.32\textwidth]{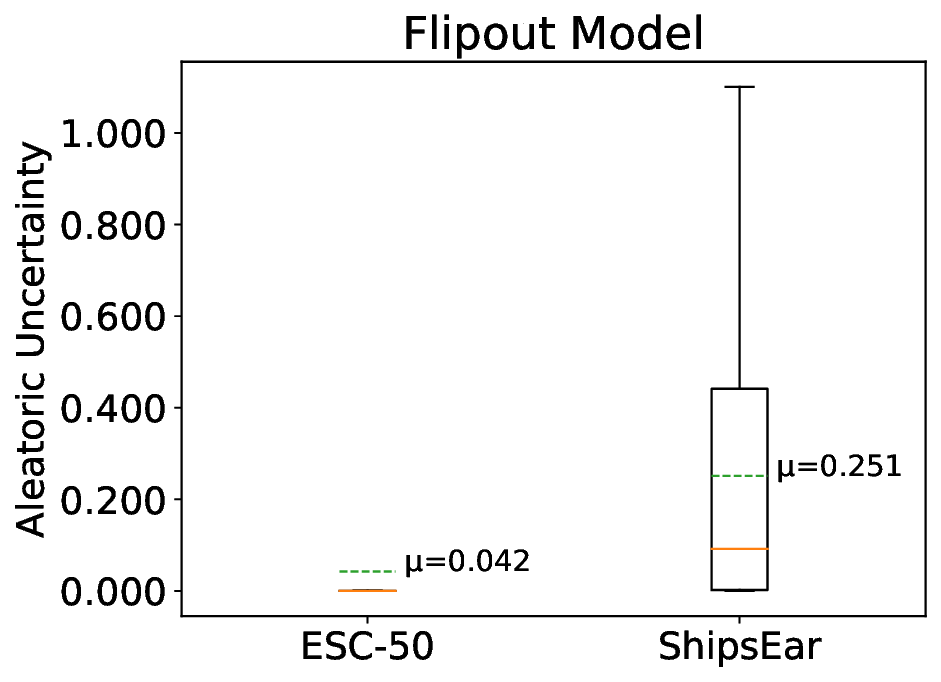}
        } \hfill
        \subfloat{
            \centering
            \includegraphics[width=.32\textwidth]{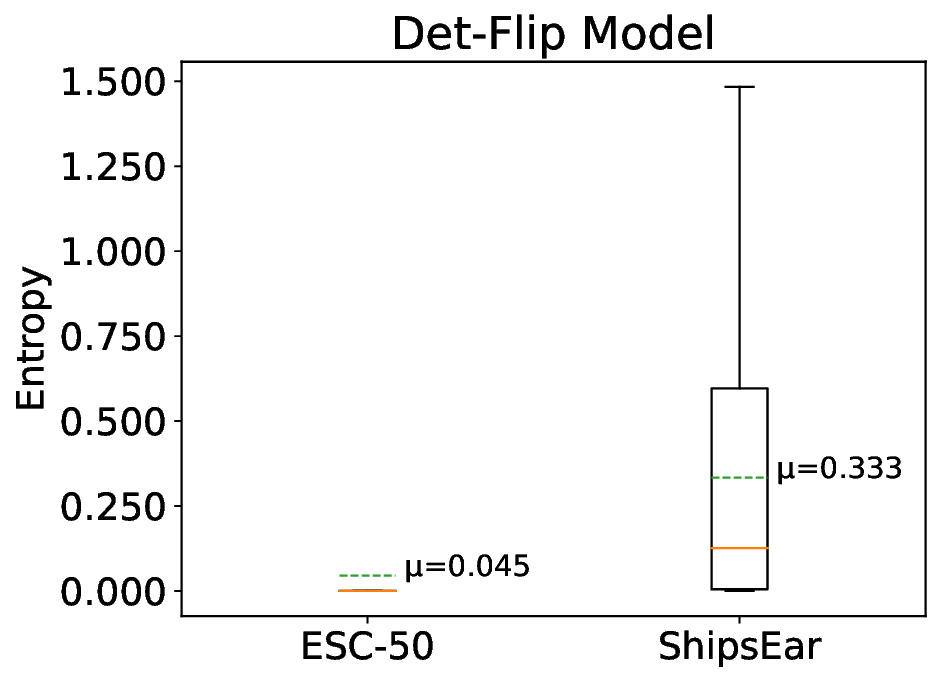}
        }
        \subfloat{
            \centering
            \includegraphics[width=.32\textwidth]{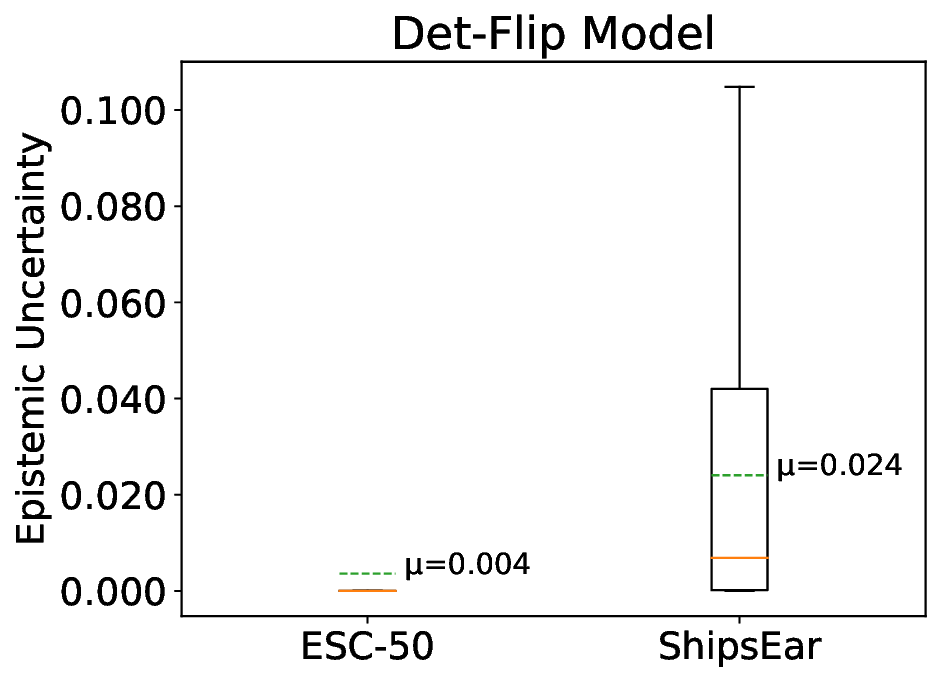}
        }
        \subfloat{
            \centering
            \includegraphics[width=.32\textwidth]{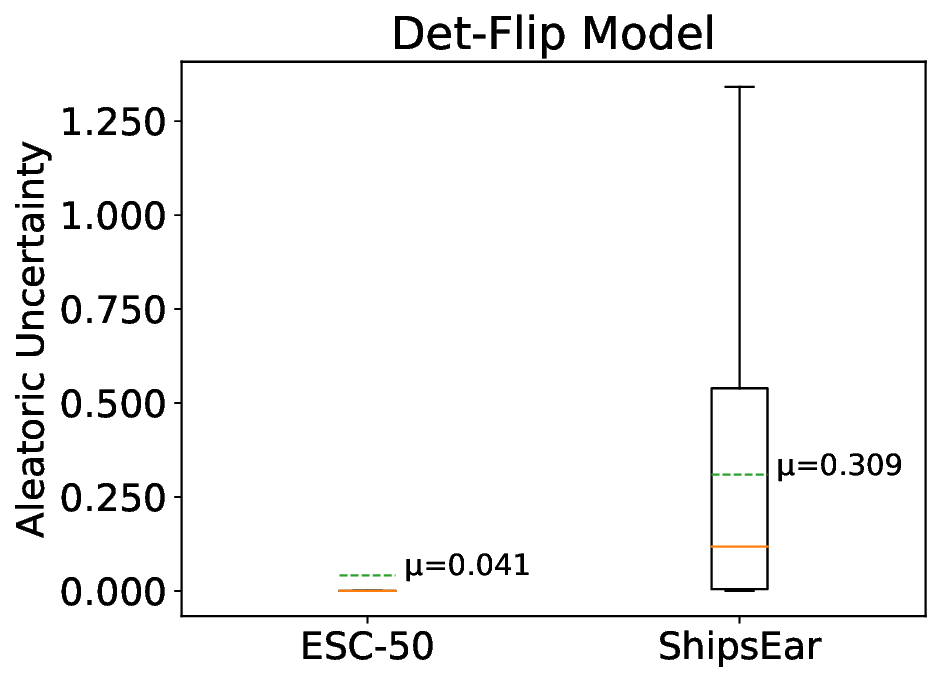}
        }

    \caption{Uncertainty box plots depicting results of \ac{MC} Dropout (top row), Flipout (middle row), and Det-Flip (bottom row) fine tuned on ESC-50. The plots compare predictive entropy (left), epistemic uncertainty (middle), and aleatoric uncertainty (right) as the models are evaluated on both the ESC-50 and the ShipsEar dataset. Both the median (orange line) and mean (dashed green line) are presented.}  
    \label{fig:ESC50BoxPlots}
\end{figure*}

\begin{figure*}[ht!]
    \centering
        \subfloat{
            \centering
            \includegraphics[width=.32\textwidth]{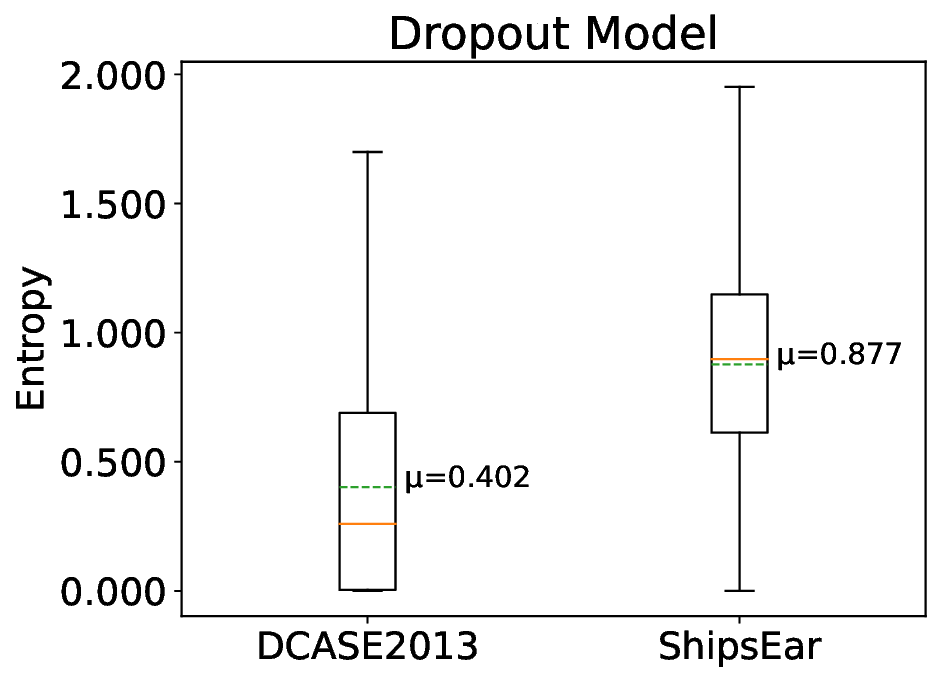}
        }
        \subfloat{
            \centering
            \includegraphics[width=.32\textwidth]{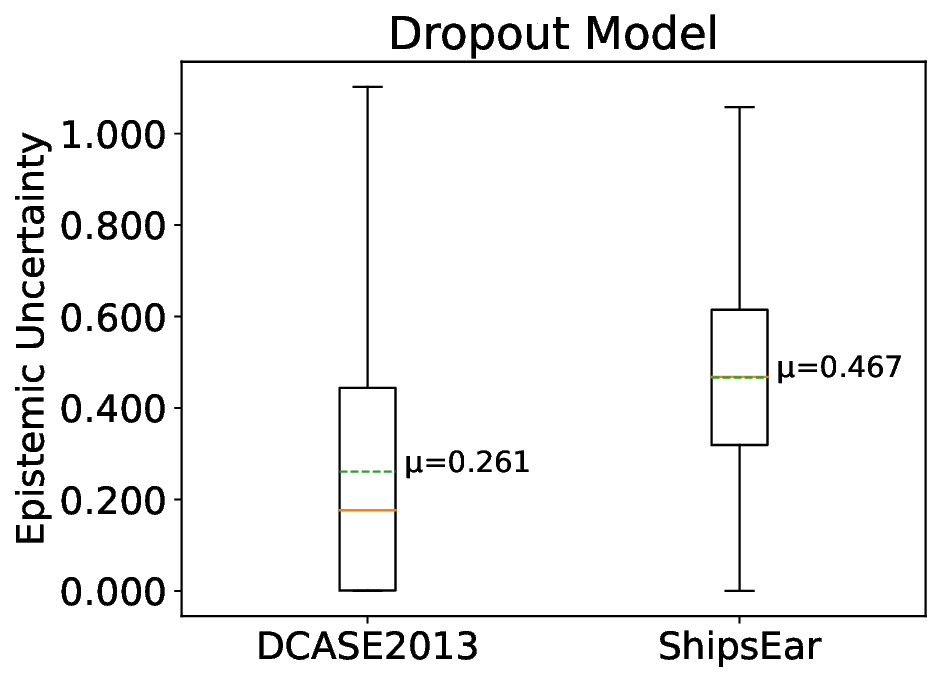}
        }
        \subfloat{
            \centering
            \includegraphics[width=.32\textwidth]{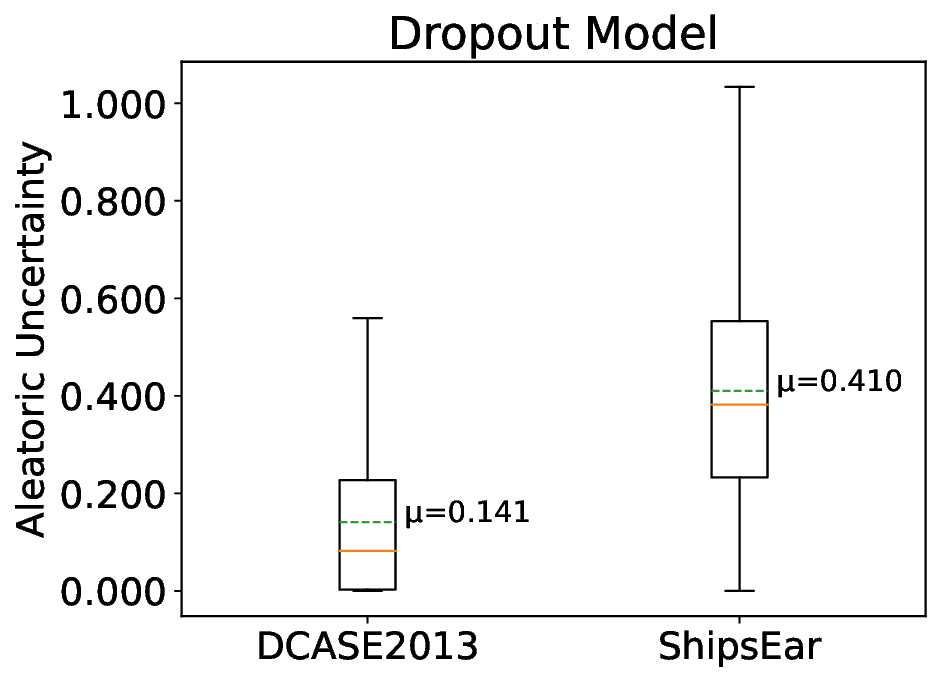}
        } \hfill
        \subfloat{
            \centering
            \includegraphics[width=.32\textwidth]{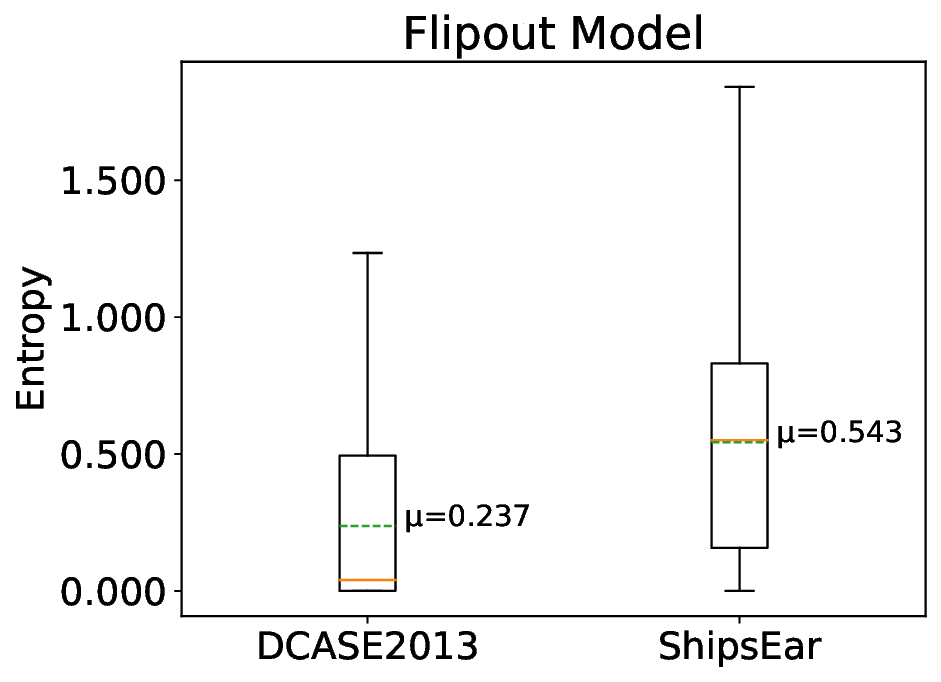}
        }
        \subfloat{
            \centering
            \includegraphics[width=.32\textwidth]{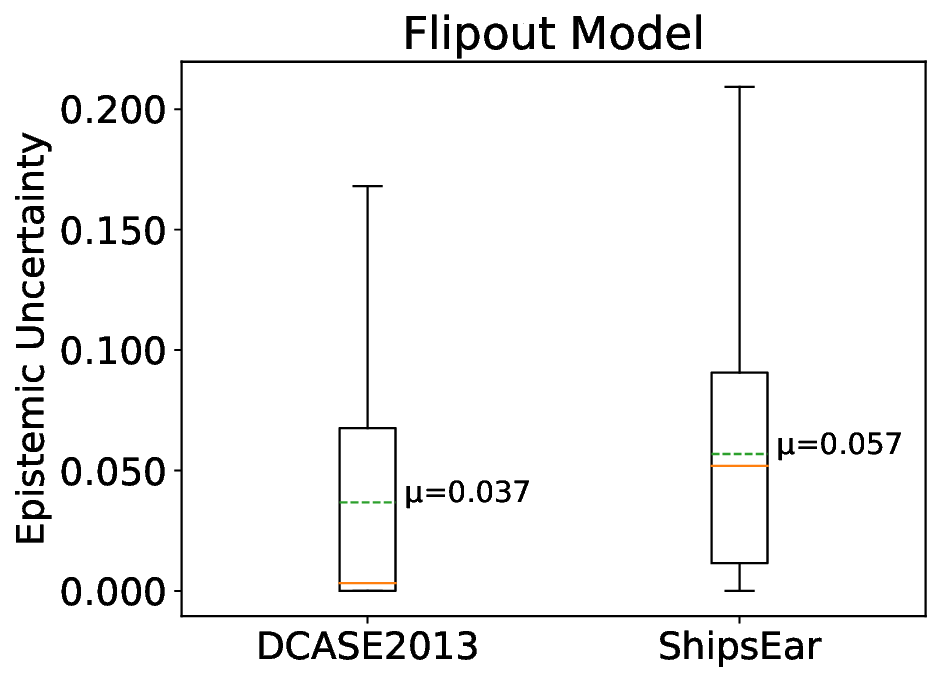}
        }
        \subfloat{
            \centering
            \includegraphics[width=.32\textwidth]{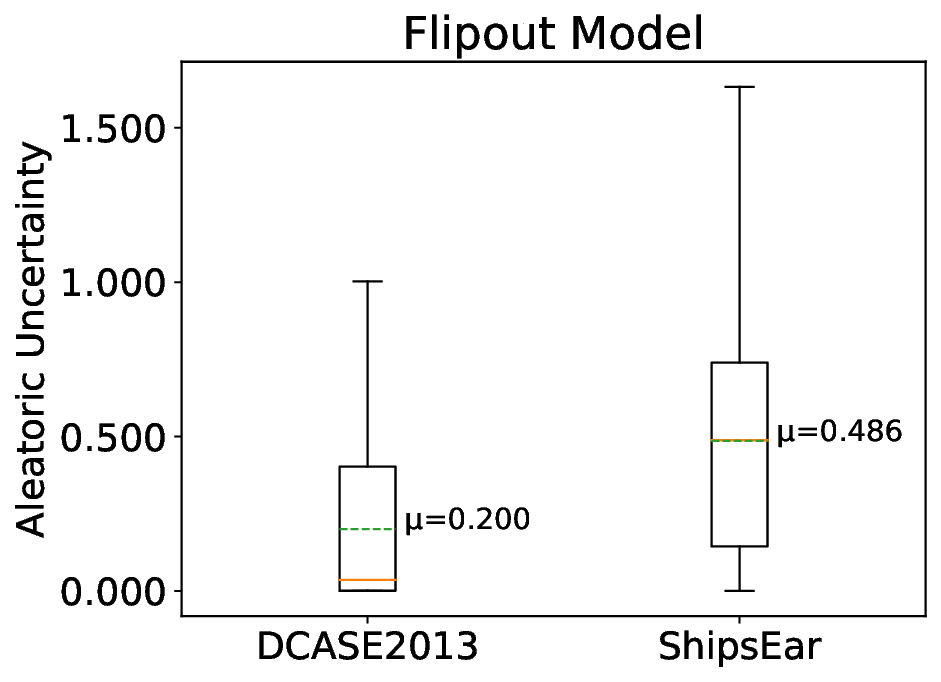}
        } \hfill
        \subfloat{
            \centering
            \includegraphics[width=.32\textwidth]{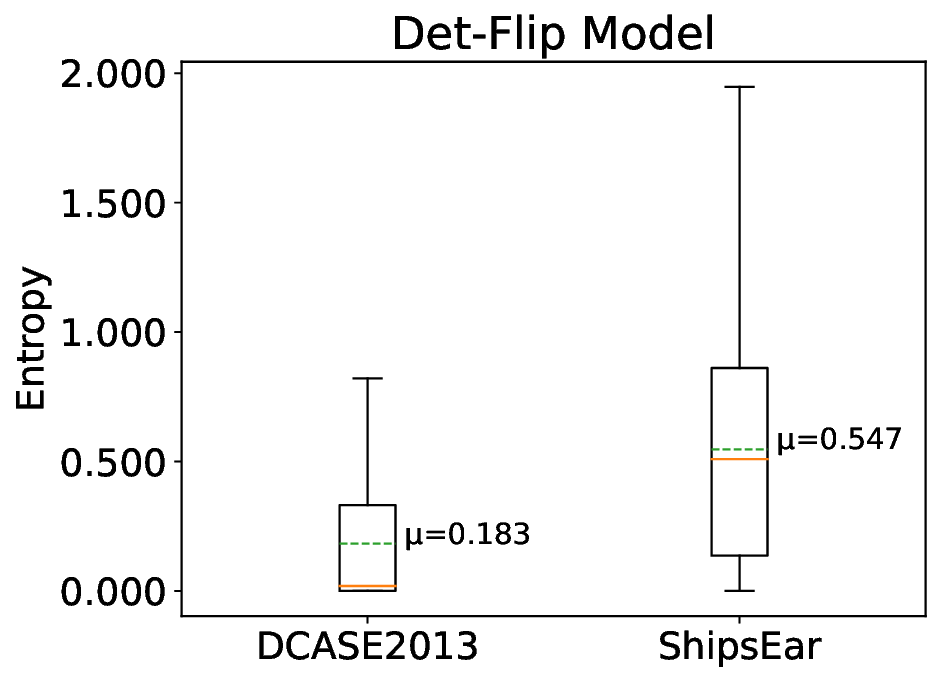}
        }
        \subfloat{
            \centering
            \includegraphics[width=.32\textwidth]{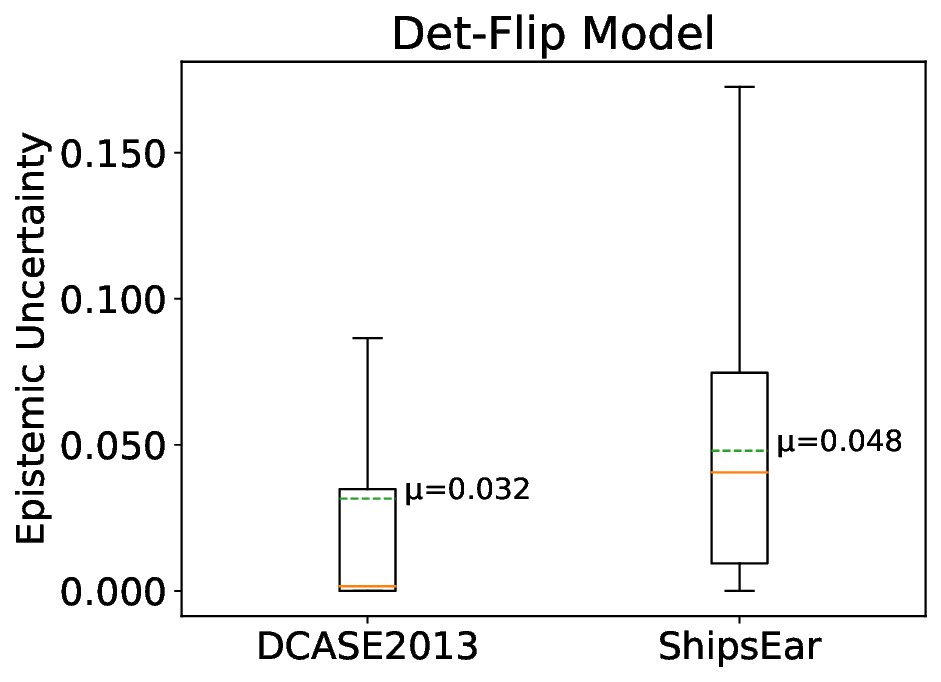}
        }
        \subfloat{
            \centering
            \includegraphics[width=.32\textwidth]{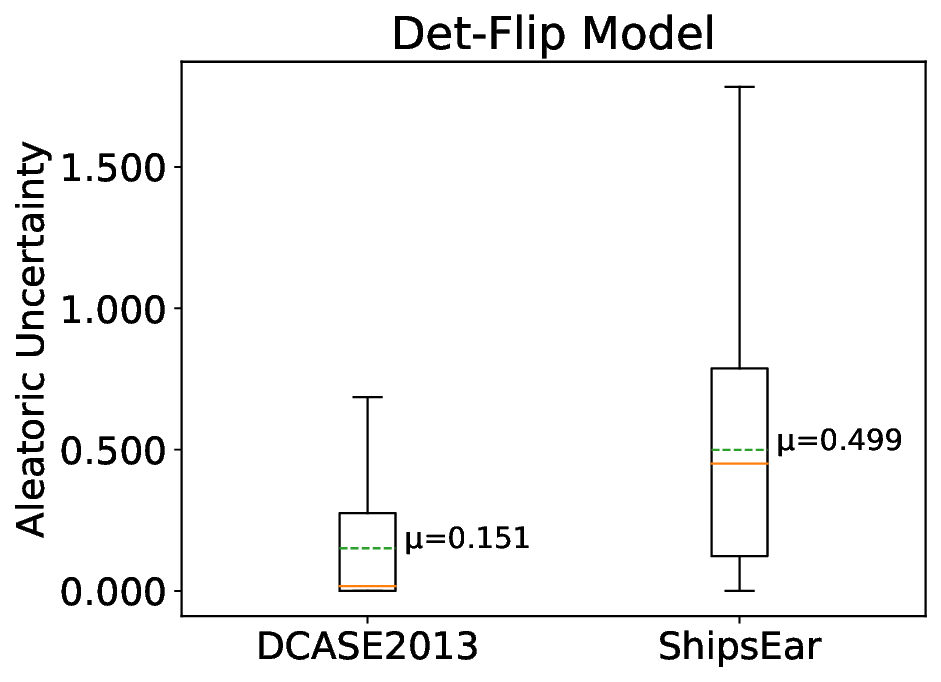}
        }

    \caption{Uncertainty box plots depicting results of \ac{MC} Dropout (top row), Flipout (middle row), and Det-Flip (bottom row) fine tuned on DCASE2013. The plots compare predictive entropy (left), epistemic uncertainty (middle), and aleatoric uncertainty (right) as the models are evaluated on both DCASE2013 and the ShipsEar dataset. Both the median (orange line) and mean (dashed green line) are presented.}  
    \label{fig:DCASE2013BoxPlots}
\end{figure*}

In Figs. \ref{fig:UrbanSound8kBoxPlots}, \ref{fig:ESC50BoxPlots}, and \ref{fig:DCASE2013BoxPlots}, we depict box plots that compare model uncertainty for each of the \ac{TL} datasets (UrbanSound8k, ESC-50, and DCASE2013) and the ShipsEar dataset. Since the fine-tuned models demonstrated superior performance compared to the fixed-feature models, we showcase the results of the fine-tuned Dropout, Flipout, and Det-Flip models. In contrast to the AudioSet results, when assessed on the ShipsEar dataset, all three model variants exhibit a notable increase in average entropy, epistemic uncertainty, and aleatoric uncertainty compared to the \ac{TL} dataset. Moreover, the plots illustrate a considerably broader distribution of uncertainty compared to the \ac{TL} datasets. This outcome is anticipated given that these models perform exceptionally well with low uncertainty on the \ac{TL} datasets.

\newpage


\section{Conclusion}\label{sec:conclusions}

In this study, we introduce \acp{VI-PANN} as a Bayesian alternative to widely adopted deterministic audio embedding methods. Trained on AudioSet, our \acp{VI-PANN} exhibit calibrated models through the use of the Flipout approach, underscoring the significance of variational audio embeddings. By adapting uncertainty decomposition techniques for multi-label classification, we enable a nuanced analysis of uncertainty estimates not only on AudioSet but also on other multi-label datasets. Notably, our work represents the first adaptation of the uncertainty decomposition from \citep{pmlr-v80-depeweg18a, Chai2018Uncertainty} for application in multi-label problems.

Our transfer learning (\ac{TL}) experiments on well-established datasets demonstrate comparable or improved performance compared to previous state-of-the-art methods, leveraging a similar model architecture. Importantly, our Det-Flip \ac{VI-PANN}, constructed with a deterministic \ac{PANN} and a Flipout classification head, achieves high performance at a relatively low cost compared to pre-training a full Flipout model. This establishes robust baselines for uncertainty-aware audio transfer learning in scenarios with limited labeled data, offering valuable insights for practitioners.

Crucially, the presented methodology for \ac{TL} with variational audio embeddings is universal and applicable to diverse audio tasks. The insights gained emphasize the intrinsic value of Bayesian neural networks in facilitating reliable and transparent transfer learning within the audio domain.

\section{Disclaimer}
The views and conclusions contained herein are those of the authors and should not be interpreted as necessarily representing the official policies or endorsements, either expressed or implied, of the U.S. Government. The U.S. Government is authorized to reproduce and distribute reprints for Government purposes notwithstanding any copyright annotations thereon.

\newpage

\bibliographystyle{IEEEbib}
\bibliography{refs}  

\begin{thebibliography}{10}

\bibitem{kong2020}
Qiuqiang Kong, Yin Cao, Turab Iqbal, Yuxuan Wang, Wenwu Wang, and Mark~D. Plumbley,
\newblock ``{PANN}s: Large-scale pretrained audio neural networks for audio pattern recognition,''
\newblock {\em IEEE/ACM Transactions on Audio, Speech, and Language Processing}, vol. 28, pp. 2880--2894, 2020.

\bibitem{tsalera2021TL}
Eleni Tsalera, Andreas Papadakis, and Maria Samarakou,
\newblock ``Comparison of pre-trained cnns for audio classification using transfer learning,''
\newblock {\em Journal of Sensor and Actuator Networks}, vol. 10, no. 4, 2021.

\bibitem{choi2017transfer}
Keunwoo Choi, George Fazekas, Mark Sandler, and Kyunghyun Cho,
\newblock ``Transfer learning for music classification and regression tasks,''
\newblock in {\em The 18th International Society of Music Information Retrieval (ISMIR) Conference 2017, Suzhou, China}. International Society of Music Information Retrieval, 2017.

\bibitem{KoikeTLHeart}
Tomoya Koike, Kun Qian, Qiuqiang Kong, Mark~D. Plumbley, Björn~W. Schuller, and Yoshiharu Yamamoto,
\newblock ``Audio for audio is better? an investigation on transfer learning models for heart sound classification,''
\newblock in {\em 2020 42nd Annual International Conference of the IEEE Engineering in Medicine \& Biology Society (EMBC)}, 2020, pp. 74--77.

\bibitem{lopez2021}
Paulo Lopez-Meyer, Juan~A. del Hoyo~Ontiveros, Hong Lu, and Georg Stemmer,
\newblock ``Efficient end-to-end audio embeddings generation for audio classification on target applications,''
\newblock in {\em ICASSP 2021 - 2021 IEEE International Conference on Acoustics, Speech and Signal Processing (ICASSP)}, 2021, pp. 601--605.

\bibitem{dimentweaklylabeledaudio}
Aleksandr Diment and Tuomas Virtanen,
\newblock ``Transfer learning of weakly labelled audio,''
\newblock in {\em 2017 IEEE Workshop on Applications of Signal Processing to Audio and Acoustics (WASPAA)}, 2017, pp. 6--10.

\bibitem{beckler_multi-label_2021}
Brandon Beckler, Andrew Pfau, Marko Orescanin, Sabrina Atchley, Nicholas Villemez, John~E. Joseph, Christopher~W. Miller, and Tetyana Margolina,
\newblock ``Multilabel classification of heterogeneous underwater soundscapes with {B}ayesian deep learning,''
\newblock {\em IEEE Journal of Oceanic Engineering}, vol. 47, no. 4, pp. 1143--1154, 2022.

\bibitem{FischerJoE}
John Fischer, Marko Orescanin, Paul Leary, and Kevin~B. Smith,
\newblock ``Active {B}ayesian deep learning with vector sensor for passive sonar sensing of the ocean,''
\newblock {\em IEEE Journal of Oceanic Engineering}, vol. 48, no. 3, pp. 837--852, 2023.

\bibitem{ortizTGRS}
Pedro Ortiz, Marko Orescanin, Veljko Petković, Scott~W. Powell, and Benjamin Marsh,
\newblock ``Decomposing satellite-based classification uncertainties in large earth science datasets,''
\newblock {\em IEEE Transactions on Geoscience and Remote Sensing}, vol. 60, pp. 1--11, 2022.

\bibitem{orescanin2021bayesian}
Marko Orescanin, Veljko Petkovi{\'c}, Scott~W Powell, Benjamin~R Marsh, and Sean~C Heslin,
\newblock ``Bayesian deep learning for passive microwave precipitation type detection,''
\newblock {\em IEEE Geoscience and Remote Sensing Letters}, 2021.

\bibitem{he2016deep}
Kaiming He, Xiangyu Zhang, Shaoqing Ren, and Jian Sun,
\newblock ``Deep residual learning for image recognition,''
\newblock in {\em Proceedings of the IEEE conference on computer vision and pattern recognition}, 2016, pp. 770--778.

\bibitem{esc50}
Karol~J. Piczak,
\newblock ``{ESC}: {Dataset} for {Environmental Sound Classification},''
\newblock in {\em Proceedings of the 23rd {Annual ACM Conference} on {Multimedia}}. 2015, pp. 1015--1018, {ACM Press}.

\bibitem{urbanSound}
Justin Salamon, Christopher Jacoby, and Juan~Pablo Bello,
\newblock ``A dataset and taxonomy for urban sound research,''
\newblock in {\em Proceedings of the 22nd ACM International Conference on Multimedia}, New York, NY, USA, 2014, MM '14, p. 1041–1044, Association for Computing Machinery.

\bibitem{dcase2013}
Dan Stowell, Dimitrios Giannoulis, Emmanouil Benetos, Mathieu Lagrange, and Mark~D. Plumbley,
\newblock ``Detection and classification of acoustic scenes and events,''
\newblock {\em IEEE Transactions on Multimedia}, vol. 17, no. 10, pp. 1733--1746, 2015.

\bibitem{Gal2016Uncertainty}
Yarin Gal,
\newblock {\em Uncertainty in Deep Learning},
\newblock Ph.D. thesis, University of Cambridge, 2016.

\bibitem{pmlr-v80-depeweg18a}
Stefan Depeweg, Jose-Miguel Hernandez-Lobato, Finale Doshi-Velez, and Steffen Udluft,
\newblock ``Decomposition of uncertainty in {B}ayesian deep learning for efficient and risk-sensitive learning,''
\newblock in {\em Proceedings of the 35th International Conference on Machine Learning}, Jennifer Dy and Andreas Krause, Eds. 10--15 Jul 2018, vol.~80 of {\em Proceedings of Machine Learning Research}, pp. 1184--1193, PMLR.

\bibitem{Chai2018Uncertainty}
Lucy~R Chai,
\newblock {\em Uncertainty estimation in {B}ayesian neural networks and links to interpretability},
\newblock M.S. thesis, University of Cambridge, 2018.

\bibitem{pfau_multi-label_2020}
Andrew~M. Pfau,
\newblock ``Multi-label classification of underwater soundscapes using deep convolutional neural networks,''
\newblock M.S. thesis, Naval Postgraduate School, Monterey, CA, 2020.

\bibitem{audioset}
Jort~F. Gemmeke, Daniel P.~W. Ellis, Dylan Freedman, Aren Jansen, Wade Lawrence, R.~Channing Moore, Manoj Plakal, and Marvin Ritter,
\newblock ``Audio set: An ontology and human-labeled dataset for audio events,''
\newblock in {\em Proc. IEEE ICASSP 2017}, New Orleans, LA, 2017.

\bibitem{gal2016dropout}
Yarin Gal and Zoubin Ghahramani,
\newblock ``Dropout as a {B}ayesian approximation: Representing model uncertainty in deep learning,''
\newblock in {\em international conference on machine learning}. PMLR, 2016, pp. 1050--1059.

\bibitem{wen2018flipout}
Yeming Wen, Paul Vicol, Jimmy Ba, Dustin Tran, and Roger~B. Grosse,
\newblock ``Flipout: Efficient pseudo-independent weight perturbations on mini-batches,''
\newblock in {\em 6th International Conference on Learning Representations, {ICLR} 2018, Vancouver, BC, Canada, April 30 - May 3, 2018, Conference Track Proceedings}. 2018, OpenReview.net.

\bibitem{kwon2020uncertainty}
Yongchan Kwon, Joong-Ho Won, Beom~Joon Kim, and Myunghee~Cho Paik,
\newblock ``Uncertainty quantification using {B}ayesian neural networks in classification: Application to biomedical image segmentation,''
\newblock {\em Computational Statistics \& Data Analysis}, vol. 142, pp. 106816, 2020.

\bibitem{McClureDWC}
Patrick McClure, Charles~Y. Zheng, Jakub Kaczmarzyk, John Rogers{-}Lee, Satrajit~S. Ghosh, Dylan Nielson, Peter~A. Bandettini, and Francisco Pereira,
\newblock ``Distributed weight consolidation: {A} brain segmentation case study,''
\newblock in {\em Advances in Neural Information Processing Systems 31: Annual Conference on Neural Information Processing Systems 2018, NeurIPS 2018, December 3-8, 2018, Montr{\'{e}}al, Canada}, Samy Bengio, Hanna~M. Wallach, Hugo Larochelle, Kristen Grauman, Nicol{\`{o}} Cesa{-}Bianchi, and Roman Garnett, Eds., 2018, pp. 4097--4107.

\bibitem{autonomousvehicles}
Fabio Arnez, Huascar Espinoza, Ansgar Radermacher, and François Terrier,
\newblock ``Towards dependable autonomous systems based on bayesian deep learning components,''
\newblock in {\em 2022 18th European Dependable Computing Conference (EDCC)}, 2022, pp. 65--72.

\bibitem{JMLR:v14:hoffman13a}
Matthew~D. Hoffman, David~M. Blei, Chong Wang, and John Paisley,
\newblock ``Stochastic variational inference,''
\newblock {\em Journal of Machine Learning Research}, vol. 14, no. 4, pp. 1303--1347, 2013.

\bibitem{bayesiantorch}
Ranganath Krishnan, Pi~Esposito, and Mahesh Subedar,
\newblock ``Bayesian-torch: Bayesian neural network layers for uncertainty estimation,'' https://github.com/IntelLabs/bayesian-torch, Jan. 2022.

\bibitem{kingma_variational_2015}
Durk~P Kingma, Tim Salimans, and Max Welling,
\newblock ``Variational dropout and the local reparameterization trick,''
\newblock in {\em Advances in Neural Information Processing Systems}, C.~Cortes, N.~Lawrence, D.~Lee, M.~Sugiyama, and R.~Garnett, Eds. 2015, vol.~28, Curran Associates, Inc.

\bibitem{Kendall_Gal_NIPS2017_2650d608}
Alex Kendall and Yarin Gal,
\newblock ``What uncertainties do we need in {B}ayesian deep learning for computer vision?,''
\newblock in {\em Advances in Neural Information Processing Systems}, I.~Guyon, U.~V. Luxburg, S.~Bengio, H.~Wallach, R.~Fergus, S.~Vishwanathan, and R.~Garnett, Eds. 2017, vol.~30, Curran Associates, Inc.

\bibitem{filos2019systematic}
Angelos Filos, Sebastian Farquhar, Aidan~N Gomez, Tim~GJ Rudner, Zachary Kenton, Lewis Smith, Milad Alizadeh, Arnoud de~Kroon, and Yarin Gal,
\newblock ``A systematic comparison of {B}ayesian deep learning robustness in diabetic retinopathy tasks,''
\newblock {\em arXiv preprint arXiv:1912.10481}, 2019.

\bibitem{moped}
Ranganath Krishnan, Mahesh Subedar, and Omesh Tickoo,
\newblock ``Specifying weight priors in {B}ayesian deep neural networks with empirical {B}ayes,''
\newblock in {\em Proceedings of the AAAI Conference on Artificial Intelligence}, 2020, vol.~34, pp. 4477--4484.

\bibitem{ortiz2023uncertainty}
Pedro Ortiz, Eleanor Casas, Marko Orescanin, Scott~W. Powell, Veljko Petkovic, and Micky Hall,
\newblock ``Uncertainty calibration of passive microwave brightness temperatures predicted by bayesian deep learning models,''
\newblock {\em Artificial Intelligence for the Earth Systems}, vol. 2, no. 4, pp. e220056, 2023.

\bibitem{santos-dominguez_shipsear_2016}
David Santos-Domínguez, Soledad Torres-Guijarro, Antonio Cardenal-López, and Antonio Pena-Gimenez,
\newblock ``{ShipsEar}: {An} underwater vessel noise database,''
\newblock {\em Applied Acoustics}, vol. 113, pp. 64--69, Dec. 2016.

\bibitem{zhang2018mixup}
Hongyi Zhang, Moustapha Cisse, Yann~N. Dauphin, and David Lopez-Paz,
\newblock ``mixup: Beyond empirical risk minimization,''
\newblock in {\em International Conference on Learning Representations}, 2018.

\bibitem{gal2017concrete}
Yarin Gal, Jiri Hron, and Alex Kendall,
\newblock ``Concrete dropout,''
\newblock in {\em Neural Information Processing Systems}, 2017.

\bibitem{srivastava2024omnivec}
Siddharth Srivastava and Gaurav Sharma,
\newblock ``Omnivec: Learning robust representations with cross modal sharing,''
\newblock in {\em Proceedings of the IEEE/CVF Winter Conference on Applications of Computer Vision}, 2024, pp. 1236--1248.

\bibitem{verbitskiy2022}
Sergey Verbitskiy, Vladimir Berikov, and Viacheslav Vyshegorodtsev,
\newblock ``Eranns: Efficient residual audio neural networks for audio pattern recognition,''
\newblock {\em Pattern Recognition Letters}, vol. 161, pp. 38--44, 2022.

\end{thebibliography}

\end{document}